\documentclass[9pt, red, preprint, secnum]{lapreprint}

\usepackage[version=4]{mhchem} 
\usepackage{siunitx}    
\usepackage{pdflscape}  
\usepackage{rotating}   
\usepackage{textgreek}  
\usepackage{gensymb}    
\usepackage[misc]{ifsym} 
\usepackage{orcidlink}  
\usepackage{listings}   
\usepackage{colortbl}   
\usepackage{tabularx}   
\usepackage{longtable}  
\usepackage{multirow}
\usepackage{snotez}     
\usepackage{csquotes}   
\usepackage{annotate-equations}

\usepackage{amsmath}
\usepackage{subfigure}
\usepackage{makecell}
\usepackage{diagbox}
\usepackage{booktabs}
\usepackage{algorithm}%
\usepackage{algorithmicx}%
\usepackage{algpseudocode}%
\usepackage{pdflscape}
\usepackage[bottom]{footmisc}
\usepackage{arydshln}

\makeatletter
\newcommand*{\defeq}{\mathrel{\rlap{%
                     \raisebox{0.3ex}{$\m@th\cdot$}}%
                     \raisebox{-0.3ex}{$\m@th\cdot$}}%
                     =}
\makeatother

\newcommand{\indep}{\perp \!\!\! \perp}
\newcommand{\tentry}[2]{#1{\scriptsize$\pm$#2}}

\DeclareSIUnit\Molar{M}

\usepackage[			
	backend=biber,      
    style=apa,   
	natbib=true,		
	hyperref=true,	    
	alldates=year,      
    uniquename=false,   
    maxbibnames=20,     
    url=true,
    doi=true,
]{biblatex}

\AtEveryBibitem{
    \clearfield{urlyear}
    \clearfield{urlmonth}
    \clearlist{language}
}

\title{Interpretable and intervenable ultrasonography-based machine learning
models for pediatric appendicitis}

\author[ \authfn{1} \orcidlink{0000-0001-8901-5062} 1 \Letter]{Ri\v{c}ards Marcinkevi\v{c}s}
\author[ \authfn{1} 2 \Letter]{Patricia Reis Wolfertstetter}
\author[ \authfn{1} 1]{Ugne Klimiene}
\author[ 1]{Kieran Chin-Cheong}
\author[ 3]{Alyssia Paschke}
\author[ 3]{Julia Zerres}
\author[ 2]{Markus Denzinger}
\author[ 1]{\\ David Niederberger}
\author[ 3, 4]{Sven Wellmann}
\author[ \authfn{2} \orcidlink{0000-0002-9889-6348} 5]{Ece Ozkan}
\author[ \authfn{2} 2]{Christian Knorr}
\author[ \authfn{2} \orcidlink{0000-0002-6004-7770} 1]{\\ Julia E. Vogt}

\affil[1]{Department of Computer Science, ETH Zurich}
\affil[2]{Department of Pediatric Surgery and Pediatric Orthopedics, Hospital St. Hedwig of the Order of St. John of God, University Children's Hospital Regensburg (KUNO)}
\affil[3]{Faculty of Medicine, University of Regensburg}
\affil[4]{Division of Neonatology, Hospital St. Hedwig of the Order of St. John of God, University Children's Hospital Regensburg (KUNO)}
\affil[5]{Department of Brain and Cognitive Sciences, Massachusetts Institute of Technology}

\metadata[]{\Letter\hspace{.5ex} Co-corresponding authors}{\href{mailto:ricards.marcinkevics@inf.ethz.ch}{ricards.marcinkevics@inf.ethz.ch}\\\hspace{-0.70cm}\href{mailto:patricia.reiswolfertstetter@barmherzige-regensburg.de}{patricia.reiswolfertstetter@barmherzige-regensburg.de}}
\metadata[\authfn{1}]{}{These authors have contributed equally to this work and share first authorship.}
\metadata[\authfn{2}]{}{These authors have contributed equally to this work and share last authorship.}
\metadata[]{Data availability}{Data are available on \href{https://doi.org/10.5281/zenodo.7711412
}{Zenodo}, the code is available on \href{https://github.com/i6092467/semi-supervised-multiview-cbm}{GitHub}.}
\metadata[]{Published version}{Marcinkevičs, R., Reis Wolfertstetter, P., Klimiene, U., Chin-Cheong, K., Paschke, A., Zerres, J., … Vogt, J. E. (2024). Interpretable and intervenable ultrasonography-based machine learning models for pediatric appendicitis. \textit{Medical Image Analysis, 91}, 103042. doi:\href{https://doi.org/10.1016/j.media.2023.103042}{10.1016/j.media.2023.103042}}

\leadauthor{Marcinkevi\v{c}s, Reis Wolfertstetter, \& Klimiene}
\shorttitle{}

\begin{document}
\maketitle
\begin{abstract}
Appendicitis is among the most frequent reasons for pediatric abdominal surgeries. 
Previous decision support systems for appendicitis have focused on clinical, laboratory, scoring, and computed tomography data and have ignored abdominal ultrasound, despite its noninvasive nature and widespread availability.  
In this work, we present interpretable machine learning models for predicting the diagnosis, management and severity of suspected appendicitis using ultrasound images. 
Our approach utilizes concept bottleneck models (CBM) that facilitate interpretation and interaction with high-level concepts understandable to clinicians. 
Furthermore, we extend CBMs to prediction problems with multiple views and incomplete concept sets.
Our models were trained on a dataset comprising 579 pediatric patients with 1709 ultrasound images accompanied by clinical and laboratory data. 
Results show that our proposed method enables clinicians to utilize a human-understandable and intervenable predictive model without compromising performance or requiring time-consuming image annotation when deployed.
For predicting the diagnosis, the extended multiview CBM attained an AUROC of 0.80 and an AUPR of 0.92, performing comparably to similar black-box neural networks trained and tested on the same dataset.
\end{abstract}
\section{Introduction} \label{sec:intro}

Appendicitis is one of the most frequent causes of abdominal pain resulting in hospital admissions of patients under 18 \citep{Wier2013}. 
The diagnosis can be challenging and relies on a combination of clinical, laboratory and imaging parameters \citep{DiSaverio2016}. 
 Despite extensive research, no specific and practically useful biomarkers for the early detection of appendicitis have been identified \citep{Acharya2016,Kiss2021}. 
 Epidemiologically and clinically, there are two forms of appendicitis: uncomplicated (subacute/exudative, phlegmonous) and complicated (gangrenous, perforated) \citep{Andersson2006,Bhangu2015,Kiss2021}. 
 Management forms include surgery as the standard method \citep{DiSaverio2016,Gorter2016} or conservative therapy \citep{Andersson2006,Svensson2012,Svensson2015,Gorter2016,CODA2020}. 

Typical imaging modalities for suspected pediatric appendicitis include ultrasonography (US), magnetic resonance imaging (MRI), and computed tomography (CT). 
US has become the primary choice due to widespread availability, lack of radiation, and improvements in resolution over the past years \citep{Park2011}.
Repeated US examinations, including B(rightness)-mode and Doppler, during the observation phase can improve diagnostic accuracy and help identify disease progression \citep{Dingemann2012,Ohba2016,Gorter2016}. 

Extensive research has been conducted on utilizing machine learning (ML) models to diagnose and manage patients with suspected appendicitis \citep{Hsieh2011,Deleger2013,Reismann2019,Aydin2020,Akmese2020,Stiel2020,Rajpurkar2020,Marcinkevics2021,RoigAparicio2021,Xia2022}. 
In brief, most models either utilize simple clinical and laboratory data \citep{Hsieh2011,Aydin2020,Akmese2020,Xia2022}, rely on hand-crafted US annotations \citep{Reismann2019,Stiel2020,Marcinkevics2021,RoigAparicio2021}, or require more expensive and invasive imaging modalities, such as CT \citep{Rajpurkar2020}. 
Despite having lower sensitivity and specificity than CT, US has been advocated as the preferred imaging modality for diagnosing acute appendicitis due to the absence of ionizing radiation and cost-effectiveness \citep{Mostbeck2016}. 
Although promising and practical, fully automated analysis of abdominal US images in this context remains an under-explored approach.

US imaging gives natural rise to multiview and multimodal data \citep{Wang2020,Qian2021}. 
For instance, the risk of breast cancer may be assessed based on multiview and multimodal US images of lesions. 
More generally, multiview learning \citep{Xu2013} concerns itself with the data comprising multiple views, essentially feature subsets, of the same source object. 
Additionally, multimodal learning \citep{Baltrusaitis2019} studies models combining, or fusing, multiple heterogeneous modalities, e.g. images and text. 
Both research directions have experienced renewed interest in the light of contrastive and self-supervised learning \citep{Tian2020,vonKuegelgen2021} and generative modeling \citep{Suzuki2022}.

Interpretable machine learning has emerged as an active research direction, \citep{DoshiVelez2017,Rudin2019}, with interpretability argued to be an essential model design principle for high-stakes application domains, such as healthcare. 
One recently re-explored approach is prediction based on high-level and human-understandable concepts \citep{Kumar2009,Lampert2009,Koh2020} or attributes. 
Most frameworks for concept-based prediction require auxiliary supervision in the form of high-level semantic features during training. 
Typically, two models are trained, as, for instance, in concept bottleneck models (CBM) \citep{Koh2020}: \mbox{(i)} one mapping from the explanatory variables to the given concepts and \mbox{(ii)} another predicting the target variable based on the previously predicted concept values. 
Such concept-based models are deemed interpretable since concepts can be inspected alongside the final model outputs and perceived as ``explanations''. 
Additionally, as opposed to classical multitask learning, a human user can intervene and interact with the model at test time by editing concept predictions and affecting downstream output. 
Beyond the restricted supervised setting mentioned earlier, there have been several efforts to learn semantically meaningful and identifiable representations when the concepts are not given explicitly \citep{Khemakhem2020,Taeb2022}.

\begin{figure*}[t]%
\centering
\includegraphics[width=0.85\textwidth]{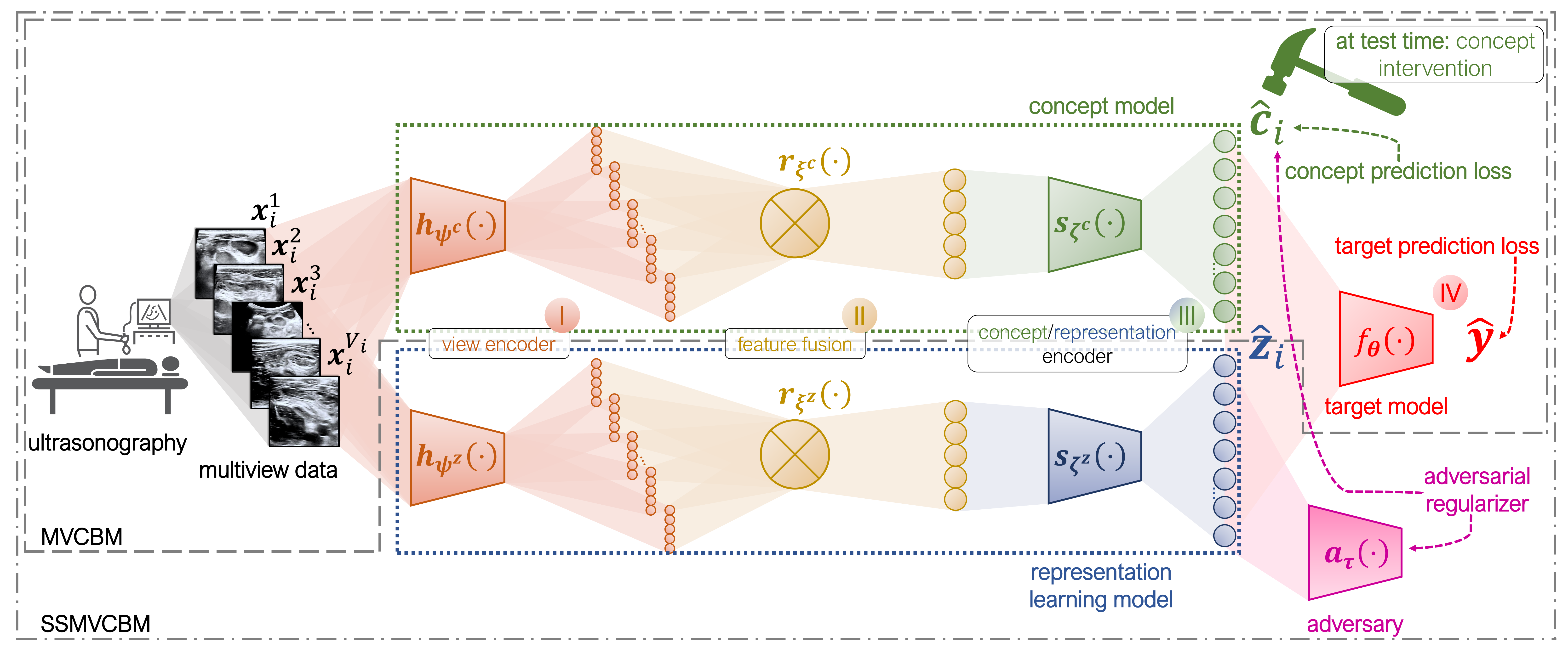}
\caption{Schematic summary of the proposed multiview concept bottleneck model (MVCBM) and its semi-supervised extension (SSMVCBM). \mbox{(I) Multiview} ultrasound images are mapped to features using a shared encoder neural network; \mbox{(II) features} are aggregated across the views; \mbox{(III) high-}level human-understandable concepts and representations are predicted based on the aggregated features; \mbox{(IV) using} concepts and representations, the target prediction is made. The MVCBM only includes view encoding, fusion, and concept prediction, whereas the SSMVCBM also performs representation learning. During training, in addition to the target prediction loss, the MVCBM is supervised by the concept prediction loss. The SSMVCBM is further penalized by an adversarial regularizer encouraging statistical independence between predicted concepts and representations.}
\label{fig:workflow}
\end{figure*}

This work presents the first effort at leveraging ML to predict diagnosis, management, and severity in pediatric patients with suspected appendicitis \emph{directly} from abdominal US images, an imaging modality frequently used in daily clinical practice. 
To this end, our models utilize interpretable concept-based classification approach due to its potential acceptance among clinicians and investigate the trade-off between interpretability and predictive performance. 
Furthermore, we propose extensions of the concept bottleneck models \citep{Koh2020} to improve their scalability to real-world medical imaging data, contributing to the recent works identifying and addressing the limitations of concept-based models \citep{Mahinpei2021,Margeloiu2021,Sawada2022,Marconato2022}. 
Specifically, we extend conventional CBMs \mbox{(i) to} the multiview classification setting and \mbox{(ii) propose} a semi-supervised representation learning approach to overcome the limitations of incomplete concept sets, i.e. when the given set of concepts does not capture the entire predictive relationship between the images and labels, making it challenging to achieve high predictive performance. 
The presented generalization of the CBMs to multiple views and incomplete concept sets is summarized in Figure~\ref{fig:workflow}. 
It is not restricted to the considered use case of pediatric appendicitis and ultrasound and can be applied to other multiview and multimodal medical imaging datasets.


\section{Materials and Methods}\label{sec:methods}

\subsection{Dataset}
In our retrospective analysis, we examined data from a cohort of 579 children and adolescents (aged 0--18 years) admitted as inpatients to the Department of Pediatric Surgery and Pediatric Orthopedics at the tertiary Children's Hospital St. Hedwig in Regensburg, Germany between January 1, 2016, and December 31, 2021, with suspected appendicitis.
Our study builds and expands upon the previous analysis of a smaller cohort of patients, published by \citet{Marcinkevics2021}.

 We utilized the hospital's database to collect retrospective data, including (potentially) multiple abdominal B-mode ultrasound images for each patient (totaling 1709 images). 
The number of views per subject ranges from 1 to 15; the images depict various regions of interest, such as the abdomen's right lower quadrant (RLQ), appendix, intestines, lymph nodes, and reproductive organs (Figure~\ref{fig:us_example}). 
Ultrasound images from admission and, if available, initial clinical course were retrieved using the software Clinic WinData/E\&L. For surgical patients, US images from the preoperative clinical course were also included. 
The images were acquired on Toshiba Xario and Aplio XG machines using Toshiba 6 MHz Convex and 12 MHz Linear transducers.
For each subject, all images relevant to the findings from Table~\ref{tab:concept_distribution} were included. Images of the organs unrelated to appendicitis, such as the liver or spleen, were excluded from the dataset. 
We also retrieved information encompassing laboratory tests, physical examination results, clinical scores, such as Alvarado (AS) and pediatric appendicitis (PAS) scores \citep{Alvarado1986,Samuel2002,RIFT2020}. AS and PAS were utilized due to the widespread use by pediatricians and pediatric surgeons for the risk stratification of children and adolescents with abdominal pain \citep{Dingemann2012}.
Last but not least, we collected expert-produced ultrasonographic findings represented by categorically-valued features. 
A subset of the latter was identified as high-level concepts relevant to decision support (Table~\ref{tab:concept_distribution}).
For patients treated operatively, surgical and histological parameters were recorded. 

The subjects were labeled w.r.t. three target variables: \mbox{(i) diag}nosis (\emph{appendicitis} vs. \emph{no appendicitis}), \mbox{(ii) ma}nagement (\emph{surgical} vs. \emph{conservative}), and \mbox{(iii) se}verity (\emph{complicated} vs. \emph{uncomplicated or no appendicitis}). 
The diagnosis was confirmed histologically in the patients who underwent appendectomy. 
Subjects treated conservatively were labeled as having appendicitis if their appendix diameter was at least \mbox{6 mm} and either AS or PAS were at least 4. 
Note that the labeling criterion above is only a proxy for the ground-truth disease status. AS and PAS help exclude children with no appendicitis \citep{RIFT2020}, whereas the addition of the US information on the enlarged appendix has been shown to increase the positive predictive value \citep{Gendel2011,Dingemann2012}.
This labeling criterion has already been utilized in the previous analyses of the data from an overlapping patient cohort \citep{Marcinkevics2021,RoigAparicio2021}. \citet{Marcinkevics2021} present a more detailed exploration to justify it. 
The management label reflects the decision made by a senior pediatric surgeon based on clinical, laboratory and US data. 
For the severity, complicated appendicitis includes cases with abscess formation, gangrene, or perforation. 

Note that the analysis below utilizes only ultrasound images and findings extracted from them. 
Our goal was to explore US image analysis and its benefits for predictive models for pediatric appendicitis. 
Nevertheless, we publicize the entire dataset, including modalities other than imaging.
Tables \ref{tab:contingency} and \ref{fig:data_overview} provide an overview of the dataset used in the final analysis. Appendix~\ref{app:appendicitis_data} contains a more comprehensive description of the dataset and its acquisition. 

\begin{table}[h]
\footnotesize
\caption{The contingency table of the pediatric appendicitis dataset of the management (M) by severity (S)
stratified by the diagnosis (D).}
\label{tab:contingency} 
\centering
\begin{tabular}{lccc}
        \toprule%
       & \multicolumn{3}{@{}c@{}}{\textbf{D: \textit{appendicitis}}} \\ \cmidrule{2-4}
        \diagbox{{\textbf{M}}}{{\textbf{S}}} & \textit{complicated} & \makecell{\textit{uncomplicated}} & \textbf{Total} \\
        \toprule
        \textit{surgical} & 97 & 135 & 232 \\
        \textit{conservative} & 0 & 151 & 151 \\
        \cline{1-4}
        \textbf{Total} & 97 & 286 & 383 \\
        \bottomrule
        & \multicolumn{3}{@{}c@{}}{\textbf{D: \textit{no appendicitis}}} \\\cmidrule{2-4}
        \diagbox{{\textbf{M}}}{{\textbf{S}}} & \textit{complicated} & \makecell{\textit{uncomplicated}} & \textbf{Total} \\
        \toprule
        \textit{surgical} & 0 & 2 & 2 \\
        \textit{conservative} & 0 & 194 & 194 \\
        \cline{1-4}
        \textbf{Total} & 0 & 196 & 196 \\
        \bottomrule
\end{tabular}
\end{table}

\subsubsection{Data Preprocessing}\label{subsec:data-preprocessing}
\begin{figure}[h]%
\centering
\includegraphics[width=0.4\columnwidth]{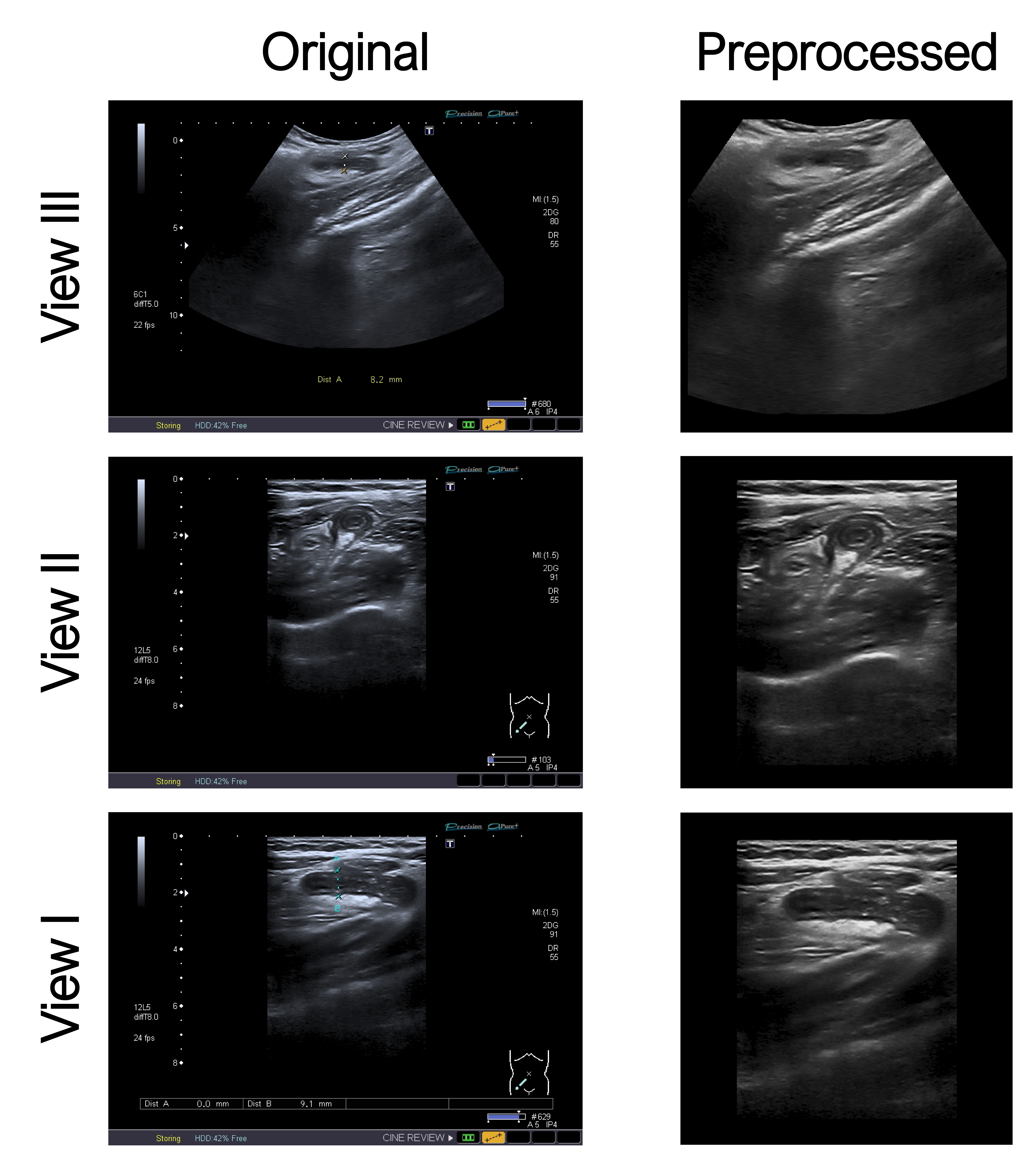}
\caption{An example of multiple US images acquired from a single patient from the pediatric appendicitis dataset. For this patient, views I and \mbox{II} correspond to longitudinal and transverse sections of the appendix, respectively; view \mbox{III} depicts the reaction in tissue surrounding the appendix. Original images \mbox{(\emph{left})} contain graphical interface elements and expert-made markers, whereas preprocessed images \mbox{(\emph{right})} have been inpainted, cropped, and padded.}\label{fig:us_example}
\end{figure}

Prior to model development and evaluation, pre-processing was performed on B-mode ultrasound images to eliminate undesired variability. 
The study being retrospective, ultrasonograms were collected as per clinical routine, and therefore, original images contained graphical user interface elements, markers, distance measurements, and other annotations.
We employed a generative inpainting model DeepFill \citep{Yu2018}, to mask and fill such objects. 
Subsequently, images were resized to 400$\times$400 px\textsuperscript{2} dimensions using zero padding when needed. 
Finally, contrast-limited histogram equalization (CLAHE) was applied, and pixel intensities were normalized to the range of $0$ and $1$. 
Figure~\ref{fig:us_example} shows an example of the multiple US views acquired from a single subject from our cohort before and after preprocessing. 

\subsection{Problem Setting and Notation}
Throughout the remaining sections, we will assume the following setting and notation. 
Consider a dataset comprising $N$ triples $\left(\left\{\boldsymbol{x}_i^{v}\right\}_{v=1}^{V_i},\, \boldsymbol{c}_i,\,y_i\right)$, for $1\leq i\leq N$, with view sequences $\left\{\boldsymbol{x}_i^{v}\right\}_{v=1}^{V_i}$, concept vectors $\boldsymbol{c}_i\in\mathbb{R}^K$ provided at training time, and labels $y_i$. 
Note that the number of views $V_i\geq1$ may vary across data points $1\leq i\leq N$. 
We will concentrate on the scenario where all views can be preprocessed and rescaled into the same dimensionality.
Nevertheless, our approach can be extended to heterogeneous data types.

Motivated by medical imaging applications, we focus on the data exhibiting characteristics described informally below. 
\mbox{(i) \emph{Partial}} \emph{observability}: not all concepts are identifiable from all views.
\mbox{(ii) \emph{View}} \emph{homogeneity}: most views contain a considerable amount of shared information and are visually similar. 
\mbox{(iii) \emph{View}} \emph{ordering}: views belonging to the same data point may be loosely ordered, e.g. spatially, temporally, or based on their importance for predicting the label. 
These properties are inspired by the multiview ultrasound dataset explored in our experiments and support some design choices described below. 

\subsection{Multiview Concept Bottleneck Models} \label{subsec:MVCBM}

Below, we present a novel approach that extends the concept bottleneck models \citep{Koh2020} to the multiview classification scenario. 
We refer to this extension as the multiview concept bottleneck model (MVCBM) hereon. 
A schematic overview of the MVCBM architecture is shown in Figure~\ref{fig:workflow}, while the model's forward pass is specified by Eqs.~(\ref{eqn:mvcbm_interim})--(\ref{eqn:mvcbm_target}). 
In brief, MVCBM comprises four modules: \mbox{(i) per-view} feature extraction; \mbox{(ii) feature} fusion; \mbox{(iii) concept} prediction, and \mbox{(iv) label} prediction. 

To address scenarios where the set of concepts provided is incomplete, aka insufficient, either due to the lack of domain knowledge or the cost of acquiring additional annotation, we have also developed a semi-supervised variant of the MVCBM, referred to as semi-supervised MVCBM (SSMVCBM). 
This approach not only utilizes the given concepts but also learns an independent representation predictive of the label.
Note that this extension will be described in the later sections. 

For data point $1\leq i\leq N$, a forward pass of the multiview concept bottleneck is given by the following equations:
\begin{subequations}
    \begin{align}
        \textrm{(i) Feature}&\textrm{{ }extraction: }\notag \\
        \boldsymbol{h}^{v}_{i}&=\boldsymbol{h}_{\boldsymbol{\psi}}\left(\boldsymbol{x}_i^{v}\right),\, 1\leq v\leq V_i,\label{eqn:mvcbm_interim}\\
        \textrm{(ii) Feature}&\textrm{{ }fusion: } \notag \\ 
        \boldsymbol{\bar{h}}_i&=\boldsymbol{r}_{\boldsymbol{\xi}}\left(\left\{\boldsymbol{h}^{v}_{i}\right\}_{v=1}^{V_i}\right),\label{eqn:mvcbm_fusion}\\
        \textrm{(iii) Concep}&\textrm{t{ }prediction: } \notag \\ 
        \boldsymbol{\hat{c}}_i&=\boldsymbol{s}_{\boldsymbol{\zeta}}\left(\boldsymbol{\bar{h}}_i\right),\label{eqn:mvcbm_concepts}\\
        \textrm{(iv) Label p}&\textrm{rediction: } \notag \\ 
        \hat{y}_i&=f_{\boldsymbol{\theta}}\left(\boldsymbol{\hat{c}}_i\right),\label{eqn:mvcbm_target}
    \end{align}
    \label{eqn:mvcbm}
\end{subequations}
where Latin letters correspond to functions and variables and Greek letters denote learnable parameters. 
Observe that parameters $\boldsymbol{\phi}=\left\{\boldsymbol{\psi}, \boldsymbol{\xi}, \boldsymbol{\zeta}\right\}$ define the concept model $\boldsymbol{g}_{\boldsymbol{\phi}}(\cdot)$ mapping a multiview feature sequence to the predicted concept values; whereas $f_{\boldsymbol{\theta}}(\cdot)$ is the target model, linking concepts and labels. 
Thus, similar to the vanilla concept bottleneck, MVCBM's forward pass can be rewritten as $\hat{y}_i=f_{\boldsymbol{\theta}}\left(\boldsymbol{g}_{\boldsymbol{\phi}}\left(\left\{\boldsymbol{h}^{v}_{i}\right\}_{v=1}^{V_i}\right)\right)$. 
In the following paragraphs, we detail each of the steps in Eq.~(\ref{eqn:mvcbm}).

\paragraph{Feature extraction}
Given an ordered view sequence $\left\{\boldsymbol{x}_i^{v}\right\}_{v=1}^{V_i}$, we first encode each view into a lower-dimensional representation, as in Eq.~(\ref{eqn:mvcbm_interim}). 
We employ a \emph{shared} encoder neural network, denoted by $\boldsymbol{h}_{\boldsymbol{\psi}}(\cdot)$. 
Weight sharing is justified by the view homogeneity and could be helpful in smaller datasets with high missingness of views. 
On the other hand, in multimodal datasets, the dissimilarities between images acquired from the same subject are significant and consistent. In this scenario, it may be preferable to train a dedicated encoder for each modality to learn modality-specific features. 
In practice, it may be prudent to use a pretrained model to initialize $\boldsymbol{h}_{\boldsymbol{\psi}}(\cdot)$, e.g. the use of ResNet and VGG architectures pretrained on natural images is standard for medical imaging applications \citep{Cheplygina2019}. 
As a result, we obtain a sequence of view-specific features.

\paragraph{Feature fusion}
To accommodate multiple views, we need to fuse, i.e. aggregate, the view-specific features within the model, as in Eq.~(\ref{eqn:mvcbm_fusion}). 
MVCBM follows a \emph{hybrid} fusion approach \citep{Baltrusaitis2019}: rather than concatenating views at the input level (\emph{early fusion}) or training an ensemble of view-specific models (\emph{late fusion}); we aggregate intermediate view-specific features $\boldsymbol{h}^{v}_{i}$ from the previous step within a single neural network. 
Although there are many viable fusion functions, in our context, the fusion must handle varying numbers of views per data point. 
As a naive approach, we consider arithmetic mean across the views $\boldsymbol{\bar{h}}_i=\frac{1}{V_i}\sum_{v=1}^{V_i}\boldsymbol{h}^{v}_{i}$ \citep{Havaei2016}. 

More generally, in Eq.~(\ref{eqn:mvcbm_fusion}) $\boldsymbol{\bar{h}}_i$ denotes the fused feature vector and $\boldsymbol{r}_{\boldsymbol{\xi}}(\cdot)$ is the fusion function with parameters $\boldsymbol{\xi}$. 
Considering partial observability of the concepts and ordering of the views, we, in addition, investigate aggregation via a \emph{learnable} function. 
Similar to \citet{Ma2019}, who utilize this trick in multiview 3D shape recognition, we combine view-specific representations via a long short-term memory (LSTM) network. 
In particular, we set the aggregated representation $\boldsymbol{\bar{h}}_i$ to the last hidden state of the view sequence, i.e. at step $V_i$. 
Note that both averaging and LSTM can handle varying numbers of views. 
Nevertheless, there are other options for $\boldsymbol{r}_{\boldsymbol{\xi}}(\cdot)$, e.g. Hadamard product or weighted average, the investigation of which we leave for future work.

\paragraph{Concept and label prediction}
The last two steps in Eqs.~(\ref{eqn:mvcbm_concepts})--(\ref{eqn:mvcbm_target}) are similar to the vanilla concept bottleneck. 
First, we predict concepts $\boldsymbol{\hat{c}}_i$ based on the fused representation $\boldsymbol{\bar{h}}_i$, using a concept encoder network $\boldsymbol{s}_{\boldsymbol{\zeta}}(\cdot)$ parameterized by $\boldsymbol{\zeta}$. 
Note that the choice of activation functions at the output of $\boldsymbol{s}_{\boldsymbol{\zeta}}(\cdot)$ depends on the type of concepts and should be adapted to whether an individual concept is categorically or continuously valued. 
The vector $\hat{\boldsymbol{c}}_i$ is then used as an input to the target model $f_{\boldsymbol{\theta}}(\cdot)$, predicting the label $\hat{y}$. 
The output activation should be chosen based on the downstream task, which can be, for example, classification or regression.

\paragraph{Loss function and optimization}
The parameters of vanilla CBMs can be optimized using independent, sequential and joint procedures \citep{Koh2020}. 
In this work, we focus on the sequential and joint approaches since they offer a more balanced trade-off between predictive performance and intervenability, as shown experimentally by \citet{Koh2020}. 

In the \emph{sequential} training, we first optimize the concept model parameters:
\begin{equation}
   \boldsymbol{\hat{\phi}}=\arg\min_{\boldsymbol{\phi}}\sum_{i=1}^N\sum_{k=1}^{K}w_i^t w_i^{c_k}\mathcal{L}^{c_k}(\hat{c}_{i,k}, c_{i,k}),
   \label{eqn:opt_seq_1}
\end{equation}
where $\mathcal{L}^{c_k}(\cdot,\,\cdot)$ is the loss function for the $k$-th concept, e.g. one could use the cross-entropy for categorically valued and squared error for a continuously valued concept, and $c_{i,k}$ refers to the value of the $k$-th concept for the $i$-th data point. 

Additionally, to address potential imbalances in the concept distributions and sparsity of specific concept-target combinations, we have introduced weights $w_i^{c_k}$ for the $k$-th concept and $w_i^t$ for the target variable of the $i$-th point, s.t. $\sum_{i=1}^N\sum_{k=1}^Kw_{i}^{c_k}=1$ and $\sum_{i=1}^Nw_i^t=1$. 
In practice, these weights can be set to the normalized inverse counts of samples in the corresponding variable classes, i.e. $w^t_i\propto1/\sum_{j=1}^N\boldsymbol{1}_{\left\{y_j=y_i\right\}}$ and $w^{c_k}_i\propto1/\sum_{j=1}^N\boldsymbol{1}_{\left\{c_{j,k}=c_{i,k}\right\}}$, where $\boldsymbol{1}_{\{\cdot\}}$ is the indicator function. 
However, other sample weighting schemes are viable.

Next, parameters $\boldsymbol{\hat{\phi}}$ are frozen, and the parameters of the target model $f_{\boldsymbol{\theta}}$ are optimized: 
\begin{equation}
    \boldsymbol{\hat{\theta}}=\arg\min_{\boldsymbol{\theta}}\sum_{i=1}^Nw_i^t\mathcal{L}^t\left(f_{\boldsymbol{\theta}}\left(\boldsymbol{\hat{c}}_i\right), y_i\right),
    \label{eqn:opt_seq_2}
\end{equation}
where $\mathcal{L}^t(\cdot,\,\cdot)$ is the loss function for the target task, and $\boldsymbol{\hat{c}}_i$ are predictions made by the frozen concept model $\boldsymbol{g}_{\boldsymbol{\hat{\phi}}}(\cdot)$.

For the \emph{joint} training, we combine the loss functions from Eqs. (\ref{eqn:opt_seq_1}) and (\ref{eqn:opt_seq_2}) into a single objective: 
\begin{equation}
    \begin{aligned}
    \boldsymbol{\hat{\phi}},\,\boldsymbol{\hat{\theta}}=&\arg\min_{\boldsymbol{\phi},\,\boldsymbol{\theta}}\Bigg\{\sum_{i=1}^N w^t_i\mathcal{L}^t(\hat{y}_i,y_i)+ \\
    &\;\;\;\;\;\;\;\;\;\;\;\;\;\;\;\;\alpha\sum_{i=1}^N\sum_{k=1}^{K}w^t_iw_i^{c_k}\mathcal{L}^{c_k}(\hat{c}_{i,k}, c_{i,k})\bigg\}, 
    \end{aligned}
    \label{eqn:opt_joint}
\end{equation}
where $\alpha>0$ controls the trade-off between target and concept predictive performance. Observe that parameters $\boldsymbol{\phi}$ and $\boldsymbol{\theta}$ are optimized simultaneously.

\paragraph{Intervenability}
A salient difference between CBMs and multitask models is that a practitioner utilizing a CBM model can interact with it by intervening on concept predictions, e.g. ``correcting'' the model by setting the predicted values to the ground truth $\hat{c}_{i,k}\defeq c_{i,k}$. 
In particular, for a data point $1\leq i\leq N$, the updated prediction after the intervention on the concepts from a subset $\mathcal{S}\subseteq\left\{1,...,K\right\}$ is given by
\begin{equation}
    \hat{y}_i^{\mathcal{S}}=f_{\boldsymbol{\hat{\theta}}}\left(\boldsymbol{\hat{c}}_{\left\{1,...,K\right\}\setminus\mathcal{S}},\,\boldsymbol{c}_{\mathcal{S}}\right),
    \label{eqn:intervention}
\end{equation}
where $\boldsymbol{\hat{c}}$ and $\boldsymbol{c}$ refer to the predicted and ground truth concept vectors, respectively. Note the notation abuse in the order of the arguments in $f_{\boldsymbol{\hat{\theta}}}(\cdot)$.

\subsection{Semi-supervised Multiview Concept Bottleneck Models}

\begin{figure}[h]
\centering

\subfigure[]{
    \centering
    \tikzset{every picture/.style={line width=0.75pt}} 
    \begin{tikzpicture}[x=0.75pt,y=0.75pt,yscale=-1,xscale=1]
    
    \draw    (93.47,86.4) -- (123,129.62) ;
    \draw [shift={(124.13,131.27)}, rotate = 235.65] [fill={rgb, 255:red, 0; green, 0; blue, 0 }  ][line width=0.08]  [draw opacity=0] (12,-3) -- (0,0) -- (12,3) -- cycle    ;
    \draw    (135.8,33.67) -- (135.8,70.93) ;
    \draw [shift={(135.8,72.93)}, rotate = 270] [fill={rgb, 255:red, 0; green, 0; blue, 0 }  ][line width=0.08]  [draw opacity=0] (12,-3) -- (0,0) -- (12,3) -- cycle    ;
    \draw    (135.47,72.93) -- (135.47,123.6) ;
    \draw [shift={(135.47,125.6)}, rotate = 270] [fill={rgb, 255:red, 0; green, 0; blue, 0 }  ][line width=0.08]  [draw opacity=0] (12,-3) -- (0,0) -- (12,3) -- cycle    ;
    \draw    (134.8,33.67) -- (102.74,72.46) ;
    \draw [shift={(101.47,74)}, rotate = 309.57] [fill={rgb, 255:red, 0; green, 0; blue, 0 }  ][line width=0.08]  [draw opacity=0] (12,-3) -- (0,0) -- (12,3) -- cycle    ;
    
    \draw  [fill={rgb, 255:red, 220; green, 220; blue, 220 }  ,fill opacity=1 ]  (135.38, 33.3) circle [x radius= 13.89, y radius= 13.89]   ;
    \draw (135.38,33.3) node   [align=left] {$\displaystyle \boldsymbol{x}$};
    \draw  [fill={rgb, 255:red, 220; green, 220; blue, 220 }  ,fill opacity=1 ]  (135.38, 139.3) circle [x radius= 13.89, y radius= 13.89]   ;
    \draw (135.38,139.3) node   [align=left] {$\displaystyle y$};
    \draw  [fill={rgb, 255:red, 220; green, 220; blue, 220 }  ,fill opacity=1 ]  (135.38, 86.3) circle [x radius= 13.89, y radius= 13.89]   ;
    \draw (135.38,86.3) node   [align=left] {$\displaystyle \boldsymbol{c}$};
    \draw  [fill={rgb, 255:red, 255; green, 255; blue, 255 }  ,fill opacity=1 ]  (93.38, 86.3) circle [x radius= 14.71, y radius= 14.71]   ;
    \draw (93.38,86.3) node   [align=left] {$\displaystyle \boldsymbol{c} '$};

    \end{tikzpicture}
}\quad\quad\quad
\subfigure[]{
    \centering
    \tikzset{every picture/.style={line width=0.75pt}} 

    \begin{tikzpicture}[x=0.75pt,y=0.75pt,yscale=-1,xscale=1]
    
    \draw    (93.47,86.4) -- (123,129.62) ;
    \draw [shift={(124.13,131.27)}, rotate = 235.65] [fill={rgb, 255:red, 0; green, 0; blue, 0 }  ][line width=0.08]  [draw opacity=0] (12,-3) -- (0,0) -- (12,3) -- cycle    ;
    \draw    (135.47,72.93) -- (135.47,123.6) ;
    \draw [shift={(135.47,125.6)}, rotate = 270] [fill={rgb, 255:red, 0; green, 0; blue, 0 }  ][line width=0.08]  [draw opacity=0] (12,-3) -- (0,0) -- (12,3) -- cycle    ;
    \draw    (136.13,86.93) -- (105.61,130.63) ;
    \draw [shift={(104.47,132.27)}, rotate = 304.94] [fill={rgb, 255:red, 0; green, 0; blue, 0 }  ][line width=0.08]  [draw opacity=0] (12,-3) -- (0,0) -- (12,3) -- cycle    ;
    \draw    (92.8,72.93) -- (92.8,123.6) ;
    \draw [shift={(92.8,125.6)}, rotate = 270] [fill={rgb, 255:red, 0; green, 0; blue, 0 }  ][line width=0.08]  [draw opacity=0] (12,-3) -- (0,0) -- (12,3) -- cycle    ;
    
    \draw  [fill={rgb, 255:red, 220; green, 220; blue, 220 }  ,fill opacity=1 ]  (93.38, 139.97) circle [x radius= 13.89, y radius= 13.89]   ;
    \draw (93.38,139.97) node   [align=left] {$\displaystyle \boldsymbol{x}$};
    \draw  [fill={rgb, 255:red, 220; green, 220; blue, 220 }  ,fill opacity=1 ]  (135.38, 139.3) circle [x radius= 13.89, y radius= 13.89]   ;
    \draw (135.38,139.3) node   [align=left] {$\displaystyle y$};
    \draw  [fill={rgb, 255:red, 220; green, 220; blue, 220 }  ,fill opacity=1 ]  (135.38, 86.3) circle [x radius= 13.89, y radius= 13.89]   ;
    \draw (135.38,86.3) node   [align=left] {$\displaystyle \boldsymbol{c}$};
    \draw  [fill={rgb, 255:red, 255; green, 255; blue, 255 }  ,fill opacity=1 ]  (93.38, 86.3) circle [x radius= 14.71, y radius= 14.71]   ;
    \draw (93.38,86.3) node   [align=left] {$\displaystyle \boldsymbol{c} '$};

    \end{tikzpicture}
}
\caption{Generative models that result in incomplete concept sets summarized as directed graphical models. Shaded and unshaded nodes correspond to observed and unobserved variables, respectively. For both (a) and (b), in general, $\boldsymbol{x}\not{\indep} y\,\vert\,\boldsymbol{c}$ since there exists an active path \citep{Geiger1990} between $\boldsymbol{x}$ and $y$ through unobserved concepts $\boldsymbol{c}'$.}
\label{fig:incomplete_concepts}
\end{figure}
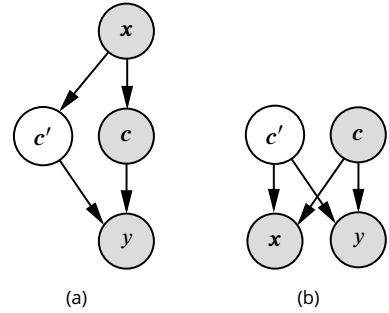

As previously stated, the set of $K$ concepts given at the training may prove incomplete, owing to factors such as the high cost of annotation, the lack of knowledge, or ethical concerns regarding the measurement of certain variables.
More formally, concept bottlenecks implicitly assume that concepts are a sufficient statistic for the target variable \citep{Yeh2020}; in other words, $\boldsymbol{x}\indep y\,\vert\,\boldsymbol{c}$.
A situation where $\boldsymbol{x}\not{\indep} y\,\vert\,\boldsymbol{c}$ may occur when some ground-truth concept variables are systematically missing in the acquired dataset, i.e. unobserved for all data points. Figure~\ref{fig:incomplete_concepts} depicts two data-generating mechanisms that may lead to the scenario described above.
When this is the case, the predictive performance of the CBM is limited since the model solely relies on the predefined set of concepts which is insufficient. 
To address this limitation, we propose a semi-supervised variant of the MVCBM (Figure~\ref{fig:workflow}) that additionally learns representations complementary to the concepts and relevant to the downstream prediction task.

Next to the feature extraction and concept prediction, \mbox{SSMVCBM} includes an unsupervised module mapping views $\left\{\boldsymbol{x}_i^v\right\}^{V_i}_{v=1}$ to the representation $\boldsymbol{\hat{z}}_i\in\mathbb{R}^J$.
To predict the label, $\boldsymbol{\hat{c}}_i$ and $\boldsymbol{\hat{z}}_i$ are concatenated and fed into the target model. 
This variant of the model is semi-supervised in that the label is predicted based on both $\boldsymbol{\hat{c}}_i$ and $\boldsymbol{\hat{z}}_i$, where $\boldsymbol{\hat{c}}_i$ are supervised by the concept prediction loss (Equation~\ref{eqn:opt_ssmvcbm_minimax}), while $\boldsymbol{\hat{z}}_i$ are complementary representations learnt without concept labels. Representations $\boldsymbol{\hat{z}}_i$ are meant to capture the residual relationship between $\boldsymbol{x}$ and $\boldsymbol{y}$ not represented among the observed concepts $\boldsymbol{c}$. 
A forward pass of the SSMVCBM is given by 
\begin{subequations}
    \begin{align}
        \textrm{(i) Feature}&\textrm{{ }extraction: }\notag \\
        \boldsymbol{h}^{\boldsymbol{c},v}_{i}&=\boldsymbol{h}_{\boldsymbol{\psi}^{\boldsymbol{c}}}\left(\boldsymbol{x}_i^{v}\right),\,1\leq v\leq V_i,\label{eqn:ssmvcbm_interim}\\
        \boldsymbol{h}^{\boldsymbol{z},v}_{i}&=\boldsymbol{h}_{\boldsymbol{\psi}^{\boldsymbol{z}}}\left(\boldsymbol{x}_i^{v}\right),\, 1\leq v\leq V_i,\notag \\
        \textrm{(ii) Feature}&\textrm{{ }fusion: } \notag \\ 
        \boldsymbol{\bar{h}}^{\boldsymbol{c}}_i&=\boldsymbol{r}_{\boldsymbol{\xi}^{\boldsymbol{c}}}\left(\left\{\boldsymbol{h}^{\boldsymbol{c},v}_{i}\right\}_{v=1}^{V_i}\right),\label{eqn:ssmvcbm_fusion}\\
        \boldsymbol{\bar{h}}^{\boldsymbol{z}}_i&=\boldsymbol{r}_{\boldsymbol{\xi}^{\boldsymbol{z}}}\left(\left\{\boldsymbol{h}^{\boldsymbol{z},v}_{i}\right\}_{v=1}^{V_i}\right),\notag \\
        \textrm{(iii) Concep}&\textrm{t{ }and representation prediction: } \notag \\ 
        \boldsymbol{\hat{c}}_i&=\boldsymbol{s}_{\boldsymbol{\zeta}^{\boldsymbol{c}}}\left(\boldsymbol{\bar{h}}^{\boldsymbol{c}}_i\right),\label{eqn:ssmvcbm_concepts}\\
        \boldsymbol{\hat{z}}_i&=\boldsymbol{s}_{\boldsymbol{\zeta}^{\boldsymbol{z}}}\left(\boldsymbol{\bar{h}}^{\boldsymbol{z}}_i\right),\notag \\
        \textrm{(iv) Label p}&\textrm{rediction: } \notag \\ 
        \hat{y}_i&=f_{\boldsymbol{\theta}}\left(\left[\boldsymbol{\hat{c}}_i, \boldsymbol{\hat{z}}_i\right]\right),\label{eqn:ssmvcbm_target}
    \end{align}
    \label{eqn:ssmvcbm}
\end{subequations}
where variables and parameters superscripted by $\boldsymbol{c}$ and $\boldsymbol{z}$ correspond to the concept and representation learning modules, respectively.

To avoid learning a representation redundant to the concepts, it is desirable that $\boldsymbol{\hat{c}}\indep\boldsymbol{\hat{z}}\,\rvert\, y$, i.e. the predicted concepts and unsupervised representations should be statistically independent conditional on the label. 
Concretely, we use another neural network $\boldsymbol{a}_{\boldsymbol{\tau}}:\:\mathbb{R}^J\rightarrow \mathbb{R}^K$, parameterized by weights $\boldsymbol{\tau}$, to quantify the degree of statistical dependence as $\max_{\boldsymbol{\tau}}\mathrm{corr}\left(\boldsymbol{a}_{\boldsymbol{\tau}}\left(\boldsymbol{\hat{z}}\right),\,\boldsymbol{\hat{c}}\right)$ \citep{Adeli2021}. 
Thus, network $\boldsymbol{a}_{\tau}$ is used to adversarially regularize representation $\boldsymbol{\hat{z}}$. 
Empirically, we observed that this regularization scheme helps de-correlate $\boldsymbol{\hat{z}}$ from concept predictions and improves the model's intervenability (Appendix~\ref{app:ssmvcbm_abl}). Additionally, note that, for the data-generating mechanisms shown in Figure~\ref{fig:incomplete_concepts}, $\boldsymbol{\hat{z}}$ does not need to identify unobserved concepts $\boldsymbol{c}'$ but rather represents the residual relationship between $\boldsymbol{x}$ and $y$.

The procedure to train SSMVCBMs is outlined in Algorithm~\ref{alg:opt_ssmvcbm}. 
Similar to the sequential optimization for (MV)CBMs as in Eqs.~(\ref{eqn:opt_seq_1}) and (\ref{eqn:opt_seq_2}), it consists of multiple steps. First, parameters $\boldsymbol{\phi}^{\boldsymbol{c}}=\left\{\boldsymbol{\psi}^{\boldsymbol{c}},\,\boldsymbol{\xi}^{\boldsymbol{c}},\,\boldsymbol{\zeta}^{\boldsymbol{c}}\right\}$ involved in concept prediction are optimized using the loss function analogous to Eq.~(\ref{eqn:opt_seq_1}). 
Then, we fix $\boldsymbol{\hat{\phi}}^{\boldsymbol{c}}$ and optimize parameters $\boldsymbol{\phi}^{\boldsymbol{z}}=\left\{\boldsymbol{\psi}^{\boldsymbol{z}},\,\boldsymbol{\xi}^{\boldsymbol{z}},\,\boldsymbol{\zeta}^{\boldsymbol{z}}\right\}$ by solving the following problem:
\begin{equation}
    \begin{aligned}
     \boldsymbol{\hat{\phi}}^{\boldsymbol{z}},\boldsymbol{\tilde{\theta}}=&\arg\min_{\boldsymbol{\phi}^{\boldsymbol{z}},\boldsymbol{\theta}}\max_{\boldsymbol{\tau}}\sum_{i=1}^Nw_i^t\mathcal{L}^t\left(\hat{y}_i, y_i\right) -\\ 
    & \;\;\;\;\;\;\;\;\;\;\;\;\;\;\;\;\;\;\;\;\lambda\sum_{i=1}^N\sum_{k=1}^Kw_i^{c_k}\mathcal{L}^{c_k}\left(\left[a_{\boldsymbol{\tau}}\left(\boldsymbol{\hat{z}}_i\right)\right]_k,\hat{c}_{i,k}\right),
    \end{aligned}
    \label{eqn:opt_ssmvcbm_minimax}
\end{equation}
where $\lambda>0$ is a tuning parameter corresponding to the weight of the adversarial regularizer. The loss function above can be extended with further regularization terms, e.g. to de-correlate individual dimensions of $\boldsymbol{\hat{z}}$ \citep{Cogswell2016}, facilitating a more straightforward interpretation. In practice, the minimax objective is optimized using adversarial training similarly to the generative adversarial networks \citep{Goodfellow2020}. 
Last but not least, parameters of the target model are re-optimized, cf. Eq.~(\ref{eqn:opt_seq_2}), treating $\boldsymbol{\hat{\phi}}^{\boldsymbol{c}}$ and $\boldsymbol{\hat{\phi}}^{\boldsymbol{z}}$ as fixed: $\boldsymbol{\hat{\theta}}=\arg\min_{\boldsymbol{\theta}}\sum_{i=1}^Nw_i^t\mathcal{L}^t\left(f_{\boldsymbol{\theta}}\left(\left[\boldsymbol{\hat{c}_i},\boldsymbol{\hat{z}_i}\right]\right),y_i\right)$.
\section{Experiments and Results}\label{sec:results}

The purpose of our experiments was twofold: \mbox{(i) to} present a proof of concept for the introduced extensions of the CBMs on simple benchmarks and \mbox{(ii) to} apply our techniques to a real-world medical imaging dataset. 
In the subsequent sections, we provide a more detailed overview of the experimental setup.

\subsection{Experimental Setup}
\paragraph{Datasets and validation scheme}
To test the feasibility of the proposed concept-based multiview classification approaches, we conducted an initial experiment using a synthetic tabular nonlinear classification problem. 
The generative process of this dataset was defined directly based on the classical concept bottleneck model, involving \mbox{(i) the} sampling of a design matrix, \mbox{(ii) the} mapping of features to concepts, and \mbox{(iii) the} use of these concepts to construct labels. 
In addition, we constructed multiple ``views'', each comprising a subset of the original feature set. 
This dataset is particularly suited to multiview approaches due to its inherent structure.
Its essential advantage over the conventional benchmarks from the literature, such as the UCSD Birds, is the presence of reliable per-data-point concept labels.
Additional details can be found in Appendix~\ref{app:synthetic}.
This problem features binary concepts that are identifiable from the given multiview observations. 
Although, herein, concept and target prediction are classification problems, all methods present are easily extendable to regression. 
In our experiments, we assessed the models' performance at \mbox{(i) target}, \mbox{(ii) concept} prediction, and \mbox{(iii) the} effectiveness of interventions on the predicted concepts. 
Additionally, to explore the scenario where the set of concepts is incomplete, we purposefully trained the models on concept subsets of varying sizes. 
We compared the performance of our approach with that of single- and multiview black-box classifiers and the vanilla concept bottlenecks \citep{Koh2020}. 
In addition to the tabular data, we constructed a semi-synthetic attribute-based natural image dataset based on the \emph{Animals with Attributes 2} \citep{Lampert2009,Xian2019} (Appendix~\ref{app:mvawa}). The experimental results for this benchmark are reported in Appendix~\ref{app:mvawa_results}.

Last but not least, to demonstrate the effectiveness of our proposed methods on real-world data, we employed ultrasound imaging and tabular clinical, laboratory, and scoring data from pediatric patients with suspected appendicitis. 
We explored three different target variables encompassing the diagnosis, treatment assignment, and complications. 
A comprehensive overview of this dataset is available in the previous sections and in Appendix~\ref{app:appendicitis_data}.
For model validation and comparison, we divided the data according to the \mbox{90\%-10\%} train-test split. 
Hyperparameter tuning was performed only on the training set using five-fold cross-validation. 
The final hyperparameter values are reported in Tables~\ref{tab:parameters_synthetic}--\ref{tab:parameters_app_severity}.
The list of high-level concepts relevant to decision support for pediatric appendicitis can be found in Table \ref{tab:concept_distribution}. 
The selection criteria for these variables were the following: \mbox{(\emph{i}) the} concept had to be detectable from ultrasound images, as confirmed by a qualified physician, and \mbox{(\emph{ii}) the} variable had to had been collected preoperatively. 

\begin{table}[h]
\centering
\caption{Explanation and descriptive statistics for the concept variables chosen for the pediatric appendicitis dataset. All concept variables are binary. The right-most column reports the percentage of the positive outcome values.}
\label{tab:concept_distribution}
\scriptsize
    \begin{tabular}{cp{3cm}p{8cm}r} 
        \toprule
        & \textbf{Name} & \textbf{Description} & \textbf{Pos., \%} \\[0.5ex] 
        \toprule
        $c_1$ & Visibility of the appendix & visibility of the vermiform appendix during the examination & 76 \\
        \midrule
        $c_2$ & Free intraperitoneal fluid & free fluids in the abdomen & 43 \\
        \midrule
        $c_3$ & Appendix layer structure & characterization of the appendix layers, e.g. irregular in case of an increasing inflammation & 14 \\
        \midrule
        $c_4$ & Target sign & axial image of the appendix with the fluid-filled center surrounded by echogenic mucosa and submucosa and hypoechoic muscularis & 13 \\
        \midrule
        $c_5$ & Surrounding tissue reaction & inflammation signs in tissue surrounding the appendix  & 33 \\ 
        \midrule
        $c_6$ & Pathological lymph nodes & enlarged and inflamed intra-abdominal lymph nodes   & 21 \\
        \midrule
        $c_7$ & Thickening of the bowel wall & edema of the intestinal wall, $>$ 2--3 mm & 8 \\
        \midrule
        $c_8$ & Coprostasis & fecal impaction in the colon & 6 \\
        \midrule
        $c_9$ & Meteorism & accumulation of gas in the intestine & 15 \\
        \bottomrule
    \end{tabular}
\end{table}

\paragraph{Ablations}
We compared several variations of the proposed multiview concept bottlenecks to better understand the role of the design choices made. 
Specifically, we trained models using sequential \linebreak \mbox{(MVCBM-seq)} and joint \mbox{(MVCBM-joint)} optimization procedures given by Eqs.~(\ref{eqn:opt_seq_1})--(\ref{eqn:opt_joint}). 
We also compared the semi-supervised extension \mbox{(SSMVCBM)} defined in Eq.~(\ref{eqn:ssmvcbm}) to the basic MVCBM. 
To facilitate meaningful comparison, we purposefully trained models under insufficient concept sets to observe if the SSMVCBM could achieve any performance improvement over the MVCBM. 
Furthermore, we investigated the impact of two fusion functions, namely, the arithmetic mean \linebreak \mbox{((SS)MVCBM-avg)} and LSTM \mbox{((SS)MVCBM-LSTM)}. 
Lastly, similar to \citet{Koh2020}, we explored interventions on the concept bottlenecks by replacing the predicted concept values with the ground truth at test time. 
The goal was to investigate whether a practitioner utilizing a concept-based model could improve its predictions interactively.

\paragraph{Baselines}
We benchmarked the performance of the (SS)MVCBMs against several baselines. 
Across all datasets, we applied single-view neural-network-based classifiers. 
Specifically, we trained MLPs for tabular data and fine-tuned ResNet-18 \citep{He2016} on images. 
As an interpretable single-view baseline, we employed vanilla CBMs.
To ensure a fair comparison between CBMs and (SS)MVCBMs, we utilized identical architectures for individual modules. 
As a black-box multiview baseline, we employed a neural network with the same architecture as for the MVCBM but trained without concept supervision in the bottleneck layer, which we refer to as multiview bottleneck (MVBM).
Similarly, as for its interpretable counterpart, we compared two ways of aggregating per-view representations: averaging and LSTM. 
Lastly, specific to the pediatric appendicitis dataset, in addition to deep-learning- and concept-based approaches, we also investigated an alternative baseline predictive model: a random forest (RF) \citep{Breiman2001} fitted on radiomic features \citep{vanGriethuysen2017}. The features were extracted from every image and averaged across the views for each subject.

\paragraph{Evaluation}
Since the intended use case of our models in healthcare applications is decision support rather than decision-making, we mainly focused on evaluating the performance of concept and label predictions using areas under receiver operating characteristic (AUROC) and precision-recall (AUPR) curves. 
Notably, for pediatric appendicitis, different metrics may be relevant depending on the target variable, e.g. a low false negative rate may be critical for diagnosis and severity, while a low false positive rate may be desirable for management to avert negative appendectomies \citep{Kryzauskas2016}. Furthermore, for appendicitis, we also assessed the predictions' calibration using the Brier score.

\paragraph{Implementation details}
We implemented MVCBM and \mbox{SSMVCBM} in PyTorch (v 1.11.0) \citep{Paszke2019}. 
Across all experiments and models, when applicable, we fine-tuned pretrained \mbox{ResNet-18 \citep{He2016}} as the shared view encoder. 
For the concept encoder and target model, we utilized MLPs with ReLU hidden activations. 
Detailed architecture specifications are provided in Appendix~\ref{app:imp}. 

We used the \emph{PyRadiomics} package \citep{vanGriethuysen2017} for radiomic feature extraction. 
Features were extracted from the whole images without prior segmentation of the region of interest since segmentation is beyond the scope of the current work. 
We computed first-order statistics, gray level size zone and gray level run length matrix features from the original and square-filtered images. 
Random forests were trained with a cost-sensitive loss function to account for class imbalance. 
ANOVA $F$-value-based feature selection was performed using nested cross-validation to improve the performance of this baseline further. 
The remainder of the implementation details can be found in Appendix~\ref{app:imp} and within the publicly available code and documentation.

\subsection{Proof of Concept on Synthetic Data}
\begin{figure*}[t]
\centering

\subfigure[]{
    \includegraphics[width=0.32\linewidth]{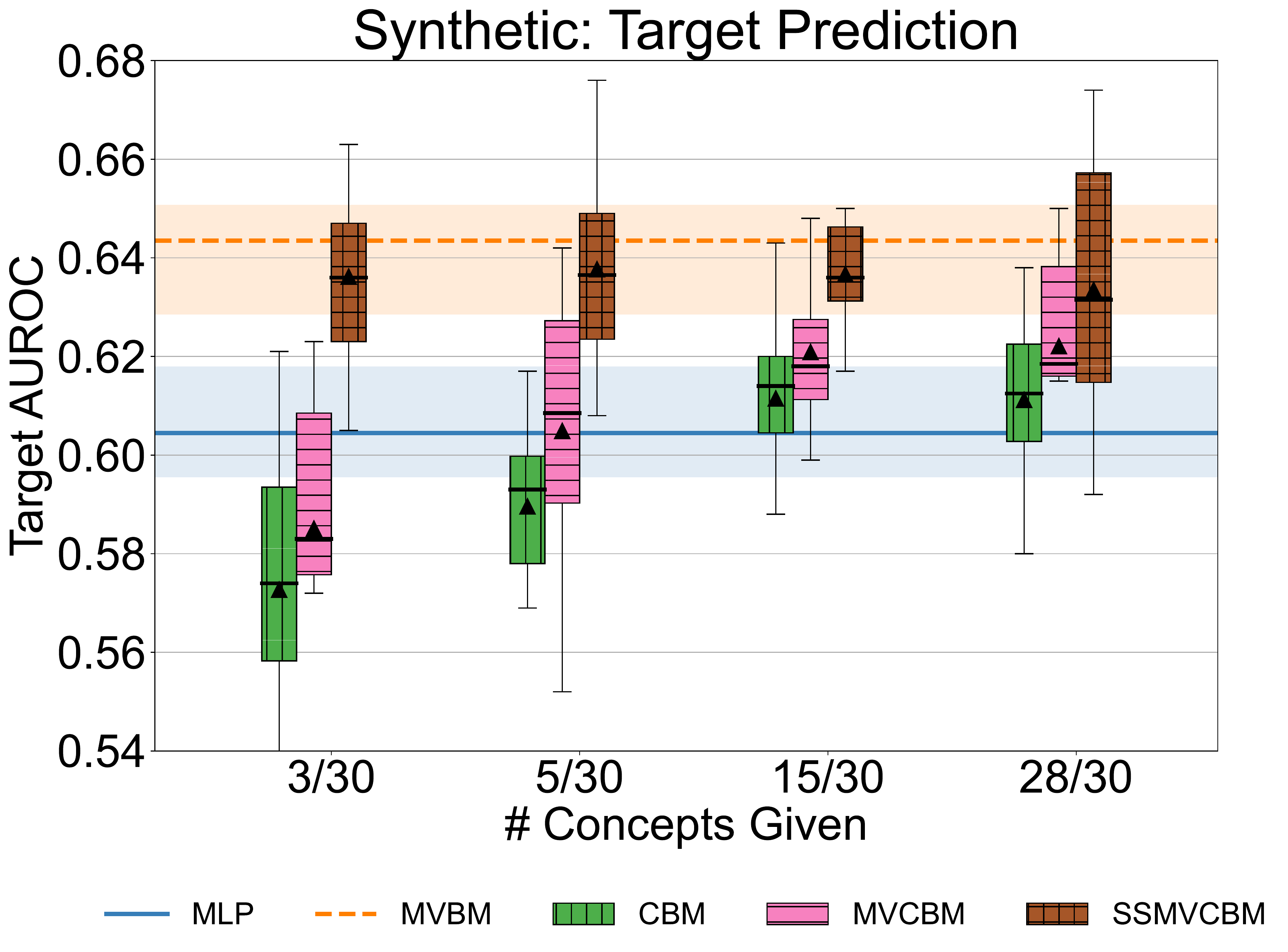}
}
\subfigure[]{
    \includegraphics[width=0.315\linewidth]{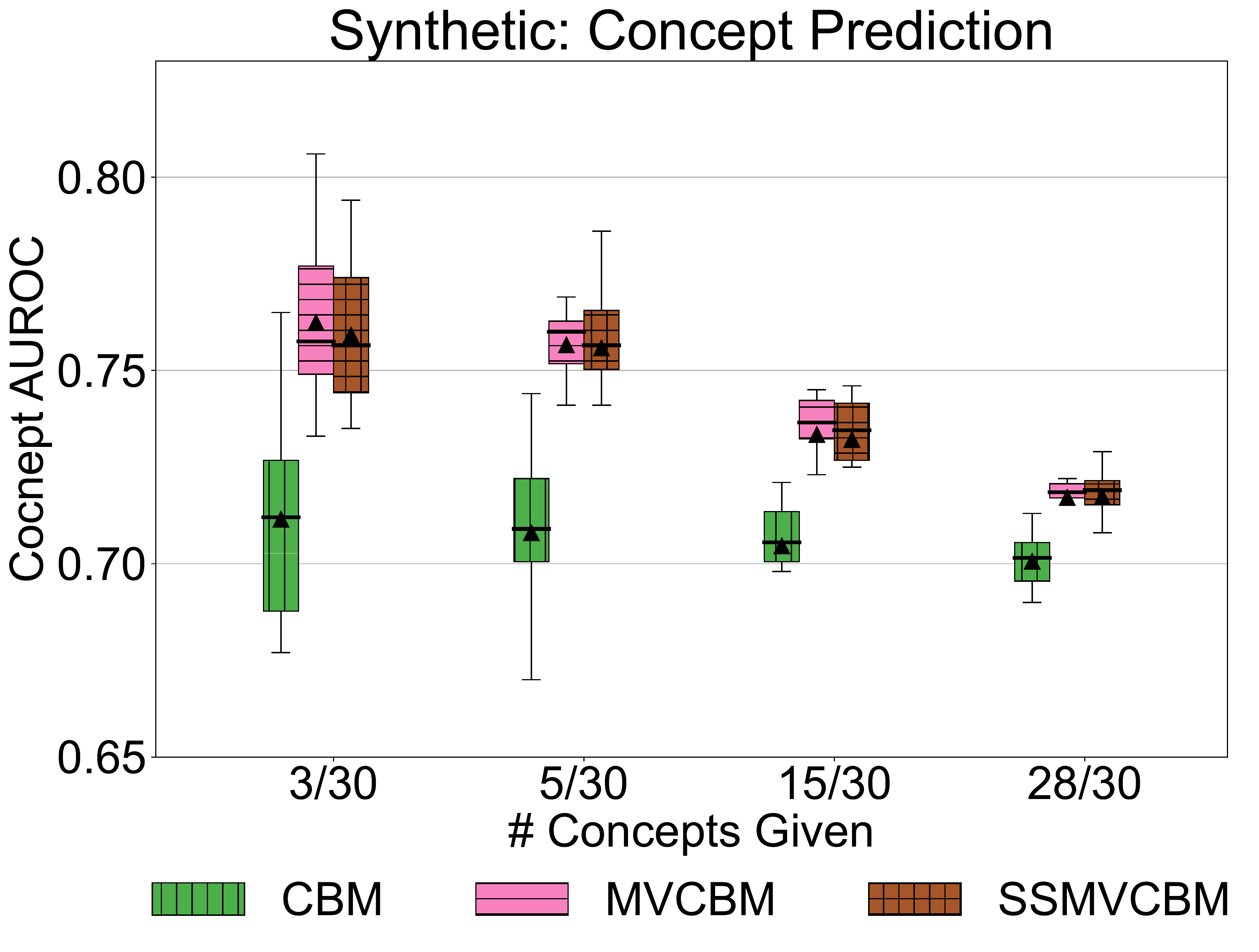}
}
\subfigure[]{
    \includegraphics[width=0.32\linewidth]{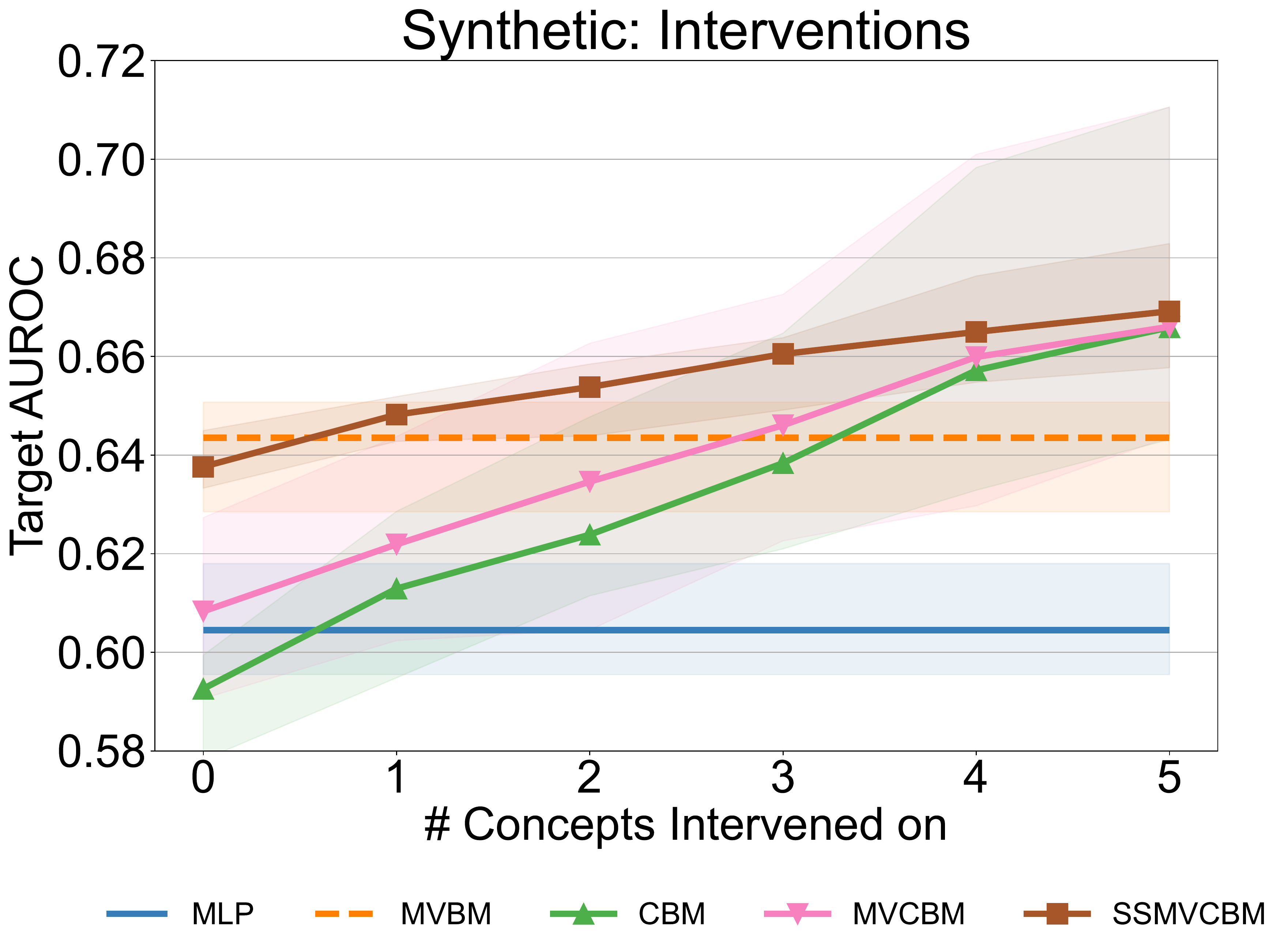}
}
\caption{Target and concept prediction results on synthetic data for the proposed multiview concept bottleneck (MVCBM) and semi-supervised multiview concept bottleneck (SSMVCBM) models alongside several baselines. All plots were produced across ten independent simulations. \mbox{(a) One-}vs-all AUROCs for predicting the target on the test data under the varying number of observed concepts. MLP and MVBM do not rely on concepts; their AUROCs are shown as horizontal lines for reference. \mbox{(b) AUROCs} for predicting concepts on the test data under the varying number of observed concepts. AUROCs were averaged across the observed concepts. \mbox{(c) AUROCs} for predicting the target on the test data after intervening on the varying number of concepts. The intervention experiment was performed for 5/30 observed concepts, i.e. under an incomplete concept set. The performances of non-intervenable MLP and MVBM baselines are shown as horizontal lines. Confidence bands correspond to interquartile ranges across independent simulations and several randomly sampled concept subsets.}
\label{fig:synthetic_res}
\end{figure*}

The first benchmark we considered was tabular synthetic nonlinear data. 
Figure~\ref{fig:synthetic_res} contains the summary of the results.
As expected, black-box and concept-based multiview approaches are consistently more accurate than their single-view counterparts at target (Figure~\ref{fig:synthetic_res}(a)) and concept prediction (Figure~\ref{fig:synthetic_res}(b)). 
Namely, a multiview bottleneck model without concept supervision (MVBM) performs considerably better than a multilayer perceptron trained on a single view (MLP) (paired $t$-test $p$-value $<0.0001$ for target AUROC); similarly, a multiview concept bottleneck (MVCBM) outperforms a simple CBM (for all numbers of concepts given, $p$-value $<0.05$ for target and concept AUROC). 
Notably, the target prediction accuracy for CBM and MVCBM increases with the number of concepts given, as shown in Figure~\ref{fig:synthetic_res}(a). 
When almost a complete concept set is provided, the performance of the multiview CBM becomes closer to that of the multiview black-box classifier. 
The semi-supervised MVCBM (SSMVCBM) performs well even when very few concepts are known and is close to the black-box baseline in most settings (for at least 5/30 concepts given, $p$-value $>0.05$ for target AUROC). 

For the concept prediction, MVCBM and SSMVCBM attain comparable performance with higher AUROCs than the single-view model (Figure~\ref{fig:synthetic_res}(b)). As expected, the semi-supervised model predicts the concepts equally well compared to the MVCBM (for all numbers of concepts given, $p$-value $>0.05$ for concept AUROC); thus, representation learning has no effect on the concept prediction. 
Lastly, we observe from Figure~\ref{fig:synthetic_res}(c) that similarly to the classical CBM, both multiview variants are intervenable, i.e. their predictive performance improves when replacing predicted concepts with the ground truth at test time. 

In addition to the results above, Appendix~\ref{app:mvawa_results} describes experiments on a semi-synthetic attribute-based natural image dataset. 
In brief, we observed similar results to the ones reported in Figure~\ref{fig:synthetic_res}. In Appendix~\ref{app:ssmvcbm_abl}, we explore the SSMVCBM in more detail, performing an ablation study on the effect of adversarial regularization.

\subsection{Application to Pediatric Appendicitis\label{sec:results_app}}
Our multiview concept bottleneck models are readily applicable to medical imaging datasets, which, in practice, often include multiple views and heterogeneous data types. 
In the following, we explore the application of the multiview CBMs to the pediatric appendicitis dataset. 

\paragraph{Predicting high-level ultrasound features}
We first evaluated the ability of all concept-based models to predict high-level appendix ultrasound features (Table~\ref{tab:concept_distribution}) from (multiple) abdominal US images. 
Table~\ref{tab:appendicitis_concept_res} contains test-set AUROCs and AUPRs achieved by the different variants of the concept bottleneck. 
In addition to comparing vanilla CBMs to their multiview and semi-supervised extensions, we investigated the effect of the optimization procedure, sequential vs. joint, and view-specific feature fusion, averaging vs. long short-term memory (LSTM). 
The models included in Table~\ref{tab:appendicitis_concept_res} were trained to predict the diagnosis (\emph{appendicitis} vs. \emph{no appendicitis}); however, we observed similar results for the management and severity, as shown in Tables~\ref{tab:appendicitis_concept_res_management}--\ref{tab:appendicitis_concept_res_severity}. 
Minor discrepancies across the three classification problems are attributable to the differences in the weights assigned to data points in the cost-sensitive loss function (Eqs.~(\ref{eqn:opt_seq_1})--(\ref{eqn:opt_joint}) and (\ref{eqn:opt_ssmvcbm_minimax})) and the choice of hyperparameter values (Tables~\ref{tab:parameters_synthetic}--\ref{tab:parameters_app_severity}).

\begin{table*}[h]
    \scriptsize
    \begin{center}
    \caption{Models' test-set performance at concept prediction on the pediatric appendicitis dataset with the diagnosis as the target variable. Test-set AUROCs and AUPRs are reported as averages and standard deviations across ten independent initializations. Herein, ``seq'' and ``joint'' denote sequential and joint optimization, respectively, whereas ``avg'' and ``LSTM'' stand for the averaging- and LSTM-based fusion. AUROCs and AUPRs that are significantly greater than the expected performance of a fair coin flip (random) are marked by ``\textsuperscript{\textbf{*}}''. Bold indicates the best result; italics indicates the second best. The meaning of the concept variables: $c_1$, visibility of the appendix; $c_2$, free intraperitoneal fluid; $c_3$, appendix layer structure; $c_4$, target sign; $c_5$, surrounding tissue reaction; $c_6$, pathological lymph nodes; $c_7$, thickening of the bowel wall; $c_8$, coprostasis; $c_9$, meteorism.}
    \label{tab:appendicitis_concept_res} 
    \begin{tabular}{p{0.8cm}lp{0.8cm}p{0.8cm}p{0.8cm}p{0.8cm}p{0.8cm}p{0.8cm}p{0.8cm}p{0.8cm}p{0.8cm}}
        \toprule
        \multirow{2}{*}{\raisebox{-\heavyrulewidth}{\textbf{Metric}}} & \multirow{2}{*}{\raisebox{-\heavyrulewidth}{\textbf{Model}}} & \multicolumn{9}{c}{\textbf{Concept}}\\
        \cmidrule{3-11} & & $c_1$ & $c_2$ & $c_3$ & $c_4$ & $c_5$ & $c_6$ & $c_7$ & $c_8$ & $c_9$ \\
        \toprule
        \multirow{11}{20mm}{\textbf{AUROC}}
        & \multicolumn{1}{l}{Random} & 0.50 & 0.50 & 0.50 & 0.50 & 0.50 & 0.50 & 0.50 & 0.50 & 0.50 \\
        \cmidrule{2-11} & \multicolumn{1}{l}{CBM-seq} & \tentry{0.52}{0.04} & \tentry{0.47}{0.04} & \tentry{0.60}{0.07}\textsuperscript{\textbf{*}} & \tentry{0.56}{0.08} & \tentry{0.63}{0.05}\textsuperscript{\textbf{*}} & \tentry{0.57}{0.05}\textsuperscript{\textbf{*}} & \tentry{0.45}{0.08} & \tentry{0.48}{0.08} & \tentry{0.39}{0.07} \\
        & \multicolumn{1}{l}{CBM-joint} & \tentry{0.50}{0.05} & \tentry{0.47}{0.03} & \tentry{0.57}{0.05}\textsuperscript{\textbf{*}} & \tentry{0.54}{0.06} & \tentry{0.64}{0.04}\textsuperscript{\textbf{*}} & \tentry{0.59}{0.05}\textsuperscript{\textbf{*}} & \tentry{0.39}{0.06} & \tentry{0.57}{0.12} & \tentry{0.38}{0.09} \\
        \cmidrule{2-11} & \multicolumn{1}{l}{MVCBM-seq-avg} & \tentry{0.61}{0.05}\textsuperscript{\textbf{*}} & \tentry{0.49}{0.05} & \tentry{0.66}{0.08}\textsuperscript{\textbf{*}} & \tentry{0.60}{0.08}\textsuperscript{\textbf{*}} & \tentry{0.51}{0.08} & \tentry{0.66}{0.08}\textsuperscript{\textbf{*}} & \textit{\tentry{0.50}{0.04}} & \tentry{0.47}{0.12} & \tentry{0.55}{0.07} \\
        & \multicolumn{1}{l}{MVCBM-seq-LSTM}   & \textit{\tentry{0.83}{0.03}\textsuperscript{\textbf{*}}} & \textit{\tentry{0.59}{0.03}\textsuperscript{\textbf{*}}} & \tentry{0.62}{0.04}\textsuperscript{\textbf{*}} & \textbf{\tentry{0.71}{0.04}\textsuperscript{\textbf{*}}} & \tentry{0.65}{0.04}\textsuperscript{\textbf{*}} & \tentry{0.67}{0.07}\textsuperscript{\textbf{*}} & \tentry{0.49}{0.07} & \textbf{\tentry{0.68}{0.10}\textsuperscript{\textbf{*}}} & \textit{\tentry{0.73}{0.06}\textsuperscript{\textbf{*}}} \\
        & \multicolumn{1}{l}{MVCBM-joint-avg}   & \tentry{0.55}{0.10} & \tentry{0.47}{0.07} & \textbf{\tentry{0.73}{0.07}\textsuperscript{\textbf{*}}} & \tentry{0.63}{0.07}\textsuperscript{\textbf{*}} & \tentry{0.61}{0.06}\textsuperscript{\textbf{*}} & \tentry{0.63}{0.07}\textsuperscript{\textbf{*}} & \tentry{0.48}{0.06} & \tentry{0.45}{0.13} & \tentry{0.54}{0.11} \\
        & \multicolumn{1}{l}{MVCBM-joint-LSTM} & \textbf{\tentry{0.85}{0.03}\textsuperscript{\textbf{*}}} & \tentry{0.55}{0.04}\textsuperscript{\textbf{*}} & \tentry{0.58}{0.04}\textsuperscript{\textbf{*}} & \textit{\tentry{0.70}{0.03}\textsuperscript{\textbf{*}}} & \textbf{\tentry{0.75}{0.02}\textsuperscript{\textbf{*}}} & \tentry{0.55}{0.09} & \tentry{0.45}{0.12} & \tentry{0.68}{0.17} & \textbf{\tentry{0.77}{0.03}\textsuperscript{\textbf{*}}} \\
        \cmidrule{2-11} & \multicolumn{1}{l}{SSMVCBM-avg} & \tentry{0.62}{0.05}\textsuperscript{\textbf{*}} & \textbf{\tentry{0.60}{0.05}\textsuperscript{\textbf{*}}} & \textit{\tentry{0.72}{0.05}\textsuperscript{\textbf{*}}} & \tentry{0.67}{0.05}\textsuperscript{\textbf{*}} & \tentry{0.54}{0.05} & \textit{\tentry{0.68}{0.08}\textsuperscript{\textbf{*}}} & \textbf{\tentry{0.53}{0.11}} & \tentry{0.43}{0.08} & \tentry{0.47}{0.07} \\
        & \multicolumn{1}{l}{SSMVCBM-LSTM} & \textbf{\tentry{0.85}{0.04}\textsuperscript{\textbf{*}}} & \tentry{0.58}{0.06}\textsuperscript{\textbf{*}} & \tentry{0.66}{0.05}\textsuperscript{\textbf{*}} & \textbf{\tentry{0.71}{0.06}\textsuperscript{\textbf{*}}} & \textit{\tentry{0.67}{0.04}\textsuperscript{\textbf{*}}} & \textbf{\tentry{0.69}{0.06}\textsuperscript{\textbf{*}}} & \tentry{0.45}{0.09} & \textit{\tentry{0.66}{0.11}\textsuperscript{\textbf{*}}} & \textit{\tentry{0.73}{0.05}\textsuperscript{\textbf{*}}} \\
        \midrule
        \multirow{11}{20mm}{\textbf{AUPR}}
        & \multicolumn{1}{l}{Random} & 0.72 & 0.49 & 0.19 & 0.23 & 0.51 & 0.26 & 0.16 & 0.13 & 0.14 \\
        \cmidrule{2-11} & \multicolumn{1}{l}{CBM-seq} & \tentry{0.71}{0.03} & \tentry{0.53}{0.03}\textsuperscript{\textbf{*}} & \tentry{0.29}{0.06}\textsuperscript{\textbf{*}} & \tentry{0.26}{0.05} & \tentry{0.64}{0.05}\textsuperscript{\textbf{*}} & \tentry{0.38}{0.06}\textsuperscript{\textbf{*}} & \tentry{0.15}{0.03} & \tentry{0.12}{0.02} & \tentry{0.11}{0.02} \\
        & \multicolumn{1}{l}{CBM-joint} & \tentry{0.73}{0.05} & \tentry{0.49}{0.04} & \tentry{0.30}{0.06}\textsuperscript{\textbf{*}} & \tentry{0.30}{0.08} & \tentry{0.64}{0.05}\textsuperscript{\textbf{*}} & \tentry{0.38}{0.09}\textsuperscript{\textbf{*}} & \tentry{0.15}{0.05} & \tentry{0.19}{0.08} & \tentry{0.11}{0.02} \\
        \cmidrule{2-11} & \multicolumn{1}{l}{MVCBM-seq-avg} & \tentry{0.79}{0.04}\textsuperscript{\textbf{*}} & \tentry{0.53}{0.06} & \textit{\tentry{0.34}{0.10}\textsuperscript{\textbf{*}}} & \tentry{0.35}{0.10}\textsuperscript{\textbf{*}} & \tentry{0.53}{0.07} & \textit{\tentry{0.41}{0.07}\textsuperscript{\textbf{*}}} & \tentry{0.17}{0.04} & \tentry{0.14}{0.04} & \tentry{0.25}{0.12} \\
        & \multicolumn{1}{l}{MVCBM-seq-LSTM}   & \tentry{0.92}{0.02}\textsuperscript{\textbf{*}} & \textit{\tentry{0.59}{0.04}\textsuperscript{\textbf{*}}} & \tentry{0.32}{0.05} & \textbf{\tentry{0.38}{0.04}\textsuperscript{\textbf{*}}} & \textit{\tentry{0.67}{0.04}\textsuperscript{\textbf{*}}} & \textbf{\tentry{0.42}{0.10}\textsuperscript{\textbf{*}}} & \tentry{0.15}{0.02} & \textit{\tentry{0.21}{0.08}} & \textbf{\tentry{0.40}{0.11}\textsuperscript{\textbf{*}}} \\
        & \multicolumn{1}{l}{MVCBM-joint-avg} & \tentry{0.75}{0.08} & \tentry{0.48}{0.06} & \textbf{\tentry{0.38}{0.09}\textsuperscript{\textbf{*}}} & \tentry{0.30}{0.06} & \tentry{0.58}{0.05}\textsuperscript{\textbf{*}} & \tentry{0.39}{0.08}\textsuperscript{\textbf{*}} & \textbf{\tentry{0.21}{0.08}} & \tentry{0.15}{0.08} & \tentry{0.16}{0.05} \\
        & \multicolumn{1}{l}{MVCBM-joint-LSTM} & \textbf{\tentry{0.94}{0.01}\textsuperscript{\textbf{*}}} & \tentry{0.50}{0.05} & \tentry{0.26}{0.08} & \textit{\tentry{0.37}{0.07}\textsuperscript{\textbf{*}}} & \textbf{\tentry{0.74}{0.04}\textsuperscript{\textbf{*}}} & \tentry{0.32}{0.09} & \tentry{0.16}{0.08} & \textbf{\tentry{0.31}{0.20}} & \tentry{0.28}{0.07}\textsuperscript{\textbf{*}} \\
        \cmidrule{2-11} & \multicolumn{1}{l}{SSMVCBM-avg} & \tentry{0.79}{0.04}\textsuperscript{\textbf{*}} & \tentry{0.58}{0.03}\textsuperscript{\textbf{*}} & \textbf{\tentry{0.38}{0.05}\textsuperscript{\textbf{*}}} & \tentry{0.34}{0.04}\textsuperscript{\textbf{*}} & \tentry{0.54}{0.06} & \textbf{\tentry{0.42}{0.08}\textsuperscript{\textbf{*}}} & \textit{\tentry{0.20}{0.06}} & \tentry{0.12}{0.04} & \tentry{0.17}{0.07} \\
        & \multicolumn{1}{l}{SSMVCBM-LSTM} & \textit{\tentry{0.93}{0.03}\textsuperscript{\textbf{*}}} & \textbf{\tentry{0.60}{0.06}\textsuperscript{\textbf{*}}} & \tentry{0.31}{0.06}\textsuperscript{\textbf{*}} & \textbf{\tentry{0.38}{0.06}\textsuperscript{\textbf{*}}} & \textit{\tentry{0.67}{0.04}\textsuperscript{\textbf{*}}} & \tentry{0.39}{0.06}\textsuperscript{\textbf{*}} & \tentry{0.19}{0.06} & \tentry{0.19}{0.07} & \textit{\tentry{0.30}{0.09}\textsuperscript{\textbf{*}}} \\
        \bottomrule
    \end{tabular}
    \vspace{1cm}
    \end{center}
\end{table*}
\begin{table*}[h!]
    \scriptsize
    \begin{center}
    \caption{Models' test-set performance at concept prediction on the appendicitis dataset with the management as the target variable. Test-set AUROCs and AUPRs are reported as averages and standard deviations across ten independent initializations. Herein, ``seq'' and ``joint'' denote sequential and joint optimization, respectively, whereas ``avg'' and ``LSTM'' stand for the averaging- and LSTM-based fusion. AUROCs and AUPRs that are significantly greater than the expected performance of a fair coin flip (random) are marked by ``\textsuperscript{\textbf{*}}''. Bold indicates the best result; italics indicates the second best. The meaning of the concept variables: $c_1$, visibility of the appendix; $c_2$, free intraperitoneal fluid; $c_3$, appendix layer structure; $c_4$, target sign; $c_5$, surrounding tissue reaction; $c_6$, pathological lymph nodes; $c_7$, thickening of the bowel wall; $c_8$, coprostasis; $c_9$, meteorism.}
    \label{tab:appendicitis_concept_res_management} 
    \begin{tabular}{p{0.8cm}lp{0.8cm}p{0.8cm}p{0.8cm}p{0.8cm}p{0.8cm}p{0.8cm}p{0.8cm}p{0.8cm}p{0.8cm}}
        \toprule
        \multirow{2}{*}{\raisebox{-\heavyrulewidth}{\textbf{Metric}}} & \multirow{2}{*}{\raisebox{-\heavyrulewidth}{\textbf{Model}}} & \multicolumn{9}{c}{\textbf{Concept}}\\
        \cmidrule{3-11} & & $c_1$ & $c_2$ & $c_3$ & $c_4$ & $c_5$ & $c_6$ & $c_7$ & $c_8$ & $c_9$ \\
        \toprule
        \multirow{11}{20mm}{\textbf{AUROC}}
        & \multicolumn{1}{l}{Random} & 0.50 & 0.50 & 0.50 & 0.50 & 0.50 & 0.50 & 0.50 & 0.50 & 0.50 \\
        \cmidrule{2-11} & \multicolumn{1}{l}{CBM-seq} & \tentry{0.51}{0.05} & \tentry{0.54}{0.07} & \tentry{0.63}{0.05}\textsuperscript{\textbf{*}} & \tentry{0.49}{0.07} & \tentry{0.65}{0.07}\textsuperscript{\textbf{*}} & \tentry{0.56}{0.06} & \tentry{0.47}{0.10} & \tentry{0.60}{0.10} & \tentry{0.54}{0.07} \\
        & \multicolumn{1}{l}{CBM-joint} & \tentry{0.54}{0.08} & \tentry{0.51}{0.08} & \tentry{0.64}{0.06}\textsuperscript{\textbf{*}} & \tentry{0.49}{0.06} & \textit{\tentry{0.67}{0.03}\textsuperscript{\textbf{*}}} & \tentry{0.54}{0.07} & \tentry{0.49}{0.07} & \tentry{0.56}{0.10} & \tentry{0.47}{0.09} \\
        \cmidrule{2-11} & \multicolumn{1}{l}{MVCBM-seq-avg} & \tentry{0.62}{0.06}\textsuperscript{\textbf{*}} & \tentry{0.48}{0.07} & \tentry{0.69}{0.03}\textsuperscript{\textbf{*}} & \tentry{0.54}{0.12} & \tentry{0.49}{0.08} & \tentry{0.60}{0.07}\textsuperscript{\textbf{*}} & \tentry{0.48}{0.09} & \tentry{0.47}{0.13} & \tentry{0.57}{0.09} \\
        & \multicolumn{1}{l}{MVCBM-seq-LSTM}   & \textbf{\tentry{0.86}{0.05}\textsuperscript{\textbf{*}}} & \textit{\tentry{0.55}{0.05}} & \tentry{0.62}{0.05}\textsuperscript{\textbf{*}} & \textit{\tentry{0.69}{0.03}\textsuperscript{\textbf{*}}} & \tentry{0.66}{0.04}\textsuperscript{\textbf{*}} & \textbf{\tentry{0.65}{0.06}\textsuperscript{\textbf{*}}} & \textit{\tentry{0.50}{0.07}} & \textbf{\tentry{0.75}{0.09}\textsuperscript{\textbf{*}}} & \textbf{\tentry{0.74}{0.06}\textsuperscript{\textbf{*}}} \\
        & \multicolumn{1}{l}{MVCBM-joint-avg}   & \tentry{0.52}{0.07} & \tentry{0.53}{0.06} & \textit{\tentry{0.71}{0.07}\textsuperscript{\textbf{*}}} & \tentry{0.59}{0.05}\textsuperscript{\textbf{*}} & \tentry{0.64}{0.07}\textsuperscript{\textbf{*}} & \textbf{\tentry{0.65}{0.04}\textsuperscript{\textbf{*}}} & \tentry{0.48}{0.10} & \tentry{0.54}{0.07} & \tentry{0.52}{0.15} \\
        & \multicolumn{1}{l}{MVCBM-joint-LSTM} & \tentry{0.80}{0.05}\textsuperscript{\textbf{*}} & \tentry{0.41}{0.08} & \tentry{0.66}{0.07}\textsuperscript{\textbf{*}} & \tentry{0.61}{0.04}\textsuperscript{\textbf{*}} & \tentry{0.66}{0.03}\textsuperscript{\textbf{*}} & \textit{\tentry{0.62}{0.07}\textsuperscript{\textbf{*}}} & \textbf{\tentry{0.51}{0.07}} & \tentry{0.62}{0.11} & \tentry{0.63}{0.08}\textsuperscript{\textbf{*}} \\
        \cmidrule{2-11} & \multicolumn{1}{l}{SSMVCBM-avg} & \tentry{0.62}{0.07}\textsuperscript{\textbf{*}} & \textbf{\tentry{0.57}{0.08}} & \textbf{\tentry{0.73}{0.04}\textsuperscript{\textbf{*}}} & \tentry{0.63}{0.05}\textsuperscript{\textbf{*}} & \tentry{0.55}{0.04} & \textbf{\tentry{0.65}{0.07}\textsuperscript{\textbf{*}}} & \textit{\tentry{0.50}{0.08}} & \tentry{0.49}{0.08} & \tentry{0.52}{0.05} \\
        & \multicolumn{1}{l}{SSMVCBM-LSTM} & \textit{\tentry{0.84}{0.02}\textsuperscript{\textbf{*}}} & \tentry{0.54}{0.05} & \tentry{0.70}{0.05}\textsuperscript{\textbf{*}} & \textbf{\tentry{0.70}{0.03}\textsuperscript{\textbf{*}}} & \textbf{\tentry{0.68}{0.05}\textsuperscript{\textbf{*}}} & \textit{\tentry{0.62}{0.07}\textsuperscript{\textbf{*}}} & \textit{\tentry{0.50}{0.10}} & \textit{\tentry{0.72}{0.05}\textsuperscript{\textbf{*}}} & \textit{\tentry{0.72}{0.10}\textsuperscript{\textbf{*}}} \\
        \midrule
        \multirow{11}{20mm}{\textbf{AUPR}}
        & \multicolumn{1}{l}{Random} & 0.72 & 0.49 & 0.19 & 0.23 & 0.51 & 0.26 & 0.16 & 0.13 & 0.14 \\
        \cmidrule{2-11} & \multicolumn{1}{l}{CBM-seq} & \tentry{0.76}{0.03} & \textit{\tentry{0.55}{0.07}} & \tentry{0.37}{0.09}\textsuperscript{\textbf{*}} & \tentry{0.23}{0.03} & \textit{\tentry{0.66}{0.07}\textsuperscript{\textbf{*}}} & \tentry{0.35}{0.10} & \textit{\tentry{0.19}{0.06}} & \tentry{0.20}{0.13} & \tentry{0.17}{0.03} \\
        & \multicolumn{1}{l}{CBM-joint} & \tentry{0.77}{0.04}\textsuperscript{\textbf{*}} & \tentry{0.51}{0.06} & \textbf{\tentry{0.45}{0.08}\textsuperscript{\textbf{*}}} & \tentry{0.24}{0.07} & \tentry{0.64}{0.04}\textsuperscript{\textbf{*}} & \tentry{0.29}{0.04} & \textit{\tentry{0.19}{0.05}} & \tentry{0.17}{0.09} & \tentry{0.15}{0.06} \\
        \cmidrule{2-11} & \multicolumn{1}{l}{MVCBM-seq-avg} & \tentry{0.79}{0.04}\textsuperscript{\textbf{*}} & \tentry{0.52}{0.08} & \tentry{0.35}{0.04}\textsuperscript{\textbf{*}} & \tentry{0.31}{0.14} & \tentry{0.51}{0.06} & \tentry{0.37}{0.08}\textsuperscript{\textbf{*}} & \tentry{0.17}{0.04} & \tentry{0.12}{0.04} & \tentry{0.18}{0.05} \\
        & \multicolumn{1}{l}{MVCBM-seq-LSTM}   & \textbf{\tentry{0.95}{0.02}\textsuperscript{\textbf{*}}} & \textit{\tentry{0.55}{0.03}\textsuperscript{\textbf{*}}} & \tentry{0.32}{0.08}\textsuperscript{\textbf{*}} & \textbf{\tentry{0.38}{0.04}\textsuperscript{\textbf{*}}} & \textit{\tentry{0.66}{0.03}\textsuperscript{\textbf{*}}} & \textit{\tentry{0.38}{0.09}\textsuperscript{\textbf{*}}} & \tentry{0.16}{0.02} & \textbf{\tentry{0.30}{0.16}} & \textbf{\tentry{0.30}{0.06}\textsuperscript{\textbf{*}}} \\
        & \multicolumn{1}{l}{MVCBM-joint-avg} & \tentry{0.71}{0.04} & \tentry{0.53}{0.05} & \tentry{0.36}{0.10}\textsuperscript{\textbf{*}} & \tentry{0.28}{0.03}\textsuperscript{\textbf{*}} & \tentry{0.60}{0.07}\textsuperscript{\textbf{*}} & \textbf{\tentry{0.39}{0.06}\textsuperscript{\textbf{*}}} & \tentry{0.17}{0.05} & \tentry{0.20}{0.07} & \tentry{0.21}{0.10} \\
        & \multicolumn{1}{l}{MVCBM-joint-LSTM} & \tentry{0.91}{0.03}\textsuperscript{\textbf{*}} & \tentry{0.44}{0.05} & \tentry{0.31}{0.06}\textsuperscript{\textbf{*}} & \tentry{0.33}{0.06}\textsuperscript{\textbf{*}} & \tentry{0.64}{0.03}\textsuperscript{\textbf{*}} & \textit{\tentry{0.38}{0.06}\textsuperscript{\textbf{*}}} & \textit{\tentry{0.19}{0.04}} & \tentry{0.19}{0.11} & \textit{\tentry{0.28}{0.14}} \\
        \cmidrule{2-11} & \multicolumn{1}{l}{SSMVCBM-avg} & \tentry{0.78}{0.06} & \textbf{\tentry{0.60}{0.07}\textsuperscript{\textbf{*}}} & \textit{\tentry{0.41}{0.08}\textsuperscript{\textbf{*}}} & \tentry{0.33}{0.08}\textsuperscript{\textbf{*}} & \tentry{0.55}{0.05} & \textbf{\tentry{0.39}{0.07}\textsuperscript{\textbf{*}}} & \textbf{\tentry{0.22}{0.06}} & \tentry{0.12}{0.02} & \tentry{0.23}{0.08} \\
        & \multicolumn{1}{l}{SSMVCBM-LSTM} & \textit{\tentry{0.93}{0.01}\textsuperscript{\textbf{*}}} & \textit{\tentry{0.55}{0.06}} & \tentry{0.38}{0.09}\textsuperscript{\textbf{*}} & \textit{\tentry{0.37}{0.06}\textsuperscript{\textbf{*}}} & \textbf{\tentry{0.67}{0.06}\textsuperscript{\textbf{*}}} & \tentry{0.35}{0.06}\textsuperscript{\textbf{*}} & \tentry{0.17}{0.05} & \textit{\tentry{0.24}{0.05}\textsuperscript{\textbf{*}}} & \tentry{0.27}{0.08}\textsuperscript{\textbf{*}} \\
        \bottomrule
    \end{tabular}
    \end{center}
\end{table*}
\begin{table*}[h!]
    \scriptsize
    \begin{center}
    \caption{Models' test-set performance at concept prediction on the appendicitis dataset with the severity as the target variable. Test-set AUROCs and AUPRs are reported as averages and standard deviations across ten independent initializations. Herein, ``seq'' and ``joint'' denote sequential and joint optimization, respectively, whereas ``avg'' and ``LSTM'' stand for the averaging- and LSTM-based fusion. AUROCs and AUPRs that are significantly greater than the expected performance of a fair coin flip (random) are marked by ``\textsuperscript{\textbf{*}}''. Bold indicates the best result; italics indicates the second best. The meaning of the concept variables: $c_1$, visibility of the appendix; $c_2$, free intraperitoneal fluid; $c_3$, appendix layer structure; $c_4$, target sign; $c_5$, surrounding tissue reaction; $c_6$, pathological lymph nodes; $c_7$, thickening of the bowel wall; $c_8$, coprostasis; $c_9$, meteorism.}
    \label{tab:appendicitis_concept_res_severity} 
    \begin{tabular}{p{0.8cm}lp{0.8cm}p{0.8cm}p{0.8cm}p{0.8cm}p{0.8cm}p{0.8cm}p{0.8cm}p{0.8cm}p{0.8cm}}
        \toprule
        \multirow{2}{*}{\raisebox{-\heavyrulewidth}{\textbf{Metric}}} & \multirow{2}{*}{\raisebox{-\heavyrulewidth}{\textbf{Model}}} & \multicolumn{9}{c}{\textbf{Concept}}\\
        \cmidrule{3-11} & & $c_1$ & $c_2$ & $c_3$ & $c_4$ & $c_5$ & $c_6$ & $c_7$ & $c_8$ & $c_9$ \\
        \toprule
        \multirow{11}{20mm}{\textbf{AUROC}}
        & \multicolumn{1}{l}{Random} & 0.50 & 0.50 & 0.50 & 0.50 & 0.50 & 0.50 & 0.50 & 0.50 & 0.50 \\
        \cmidrule{2-11} & \multicolumn{1}{l}{CBM-seq} & \tentry{0.51}{0.04} & \textit{\tentry{0.58}{0.06}\textsuperscript{\textbf{*}}} & \tentry{0.61}{0.08}\textsuperscript{\textbf{*}} & \tentry{0.52}{0.09} & \tentry{0.62}{0.04}\textsuperscript{\textbf{*}} & \tentry{0.62}{0.05}\textsuperscript{\textbf{*}} & \tentry{0.47}{0.09} & \tentry{0.57}{0.11} & \tentry{0.50}{0.08} \\
        & \multicolumn{1}{l}{CBM-joint} & \tentry{0.55}{0.06} & \tentry{0.46}{0.06} & \tentry{0.66}{0.06}\textsuperscript{\textbf{*}} & \tentry{0.47}{0.06} & \textit{\tentry{0.64}{0.04}\textsuperscript{\textbf{*}}} & \tentry{0.53}{0.07} & \tentry{0.50}{0.07} & \tentry{0.58}{0.10}\textsuperscript{\textbf{*}} & \tentry{0.49}{0.04} \\
        \cmidrule{2-11} & \multicolumn{1}{l}{MVCBM-seq-avg} & \tentry{0.54}{0.08} & \tentry{0.55}{0.04} & \textbf{\tentry{0.72}{0.07}\textsuperscript{\textbf{*}}} & \tentry{0.62}{0.04}\textsuperscript{\textbf{*}} & \tentry{0.50}{0.05} & \tentry{0.64}{0.06}\textsuperscript{\textbf{*}} & \tentry{0.51}{0.10} & \tentry{0.47}{0.11} & \tentry{0.54}{0.10} \\
        & \multicolumn{1}{l}{MVCBM-seq-LSTM}   & \textbf{\tentry{0.82}{0.04}\textsuperscript{\textbf{*}}} & \tentry{0.53}{0.04} & \tentry{0.62}{0.04}\textsuperscript{\textbf{*}} & \textbf{\tentry{0.69}{0.04}\textsuperscript{\textbf{*}}} & \tentry{0.62}{0.05}\textsuperscript{\textbf{*}} & \textbf{\tentry{0.72}{0.05}\textsuperscript{\textbf{*}}} & \textbf{\tentry{0.64}{0.06}\textsuperscript{\textbf{*}}} & \textbf{\tentry{0.78}{0.03}\textsuperscript{\textbf{*}}} & \textbf{\tentry{0.70}{0.06}\textsuperscript{\textbf{*}}} \\
        & \multicolumn{1}{l}{MVCBM-joint-avg}   & \tentry{0.54}{0.09} & \tentry{0.51}{0.06} & \tentry{0.70}{0.06}\textsuperscript{\textbf{*}} & \tentry{0.59}{0.08}\textsuperscript{\textbf{*}} & \tentry{0.61}{0.06}\textsuperscript{\textbf{*}} & \tentry{0.62}{0.05}\textsuperscript{\textbf{*}} & \tentry{0.54}{0.15} & \tentry{0.48}{0.14} & \tentry{0.55}{0.12} \\
        & \multicolumn{1}{l}{MVCBM-joint-LSTM} & \textbf{\tentry{0.82}{0.03}\textsuperscript{\textbf{*}}} & \tentry{0.48}{0.06} & \tentry{0.66}{0.07}\textsuperscript{\textbf{*}} & \tentry{0.64}{0.06}\textsuperscript{\textbf{*}} & \textbf{\tentry{0.65}{0.05}\textsuperscript{\textbf{*}}} & \tentry{0.64}{0.09}\textsuperscript{\textbf{*}} & \tentry{0.47}{0.09} & \tentry{0.61}{0.14} & \textit{\tentry{0.65}{0.05}\textsuperscript{\textbf{*}}} \\
        \cmidrule{2-11} & \multicolumn{1}{l}{SSMVCBM-avg} & \tentry{0.53}{0.06}\textsuperscript{\textbf{*}} & \tentry{0.56}{0.08}\textsuperscript{\textbf{*}} & \textit{\tentry{0.71}{0.05}\textsuperscript{\textbf{*}}} & \tentry{0.60}{0.06}\textsuperscript{\textbf{*}} & \tentry{0.51}{0.05} & \tentry{0.64}{0.09}\textsuperscript{\textbf{*}} & \tentry{0.46}{0.08} & \tentry{0.48}{0.09} & \tentry{0.53}{0.03} \\
        & \multicolumn{1}{l}{SSMVCBM-LSTM} & \textit{\tentry{0.77}{0.10}\textsuperscript{\textbf{*}}} & \textbf{\tentry{0.59}{0.08}} & \tentry{0.70}{0.06}\textsuperscript{\textbf{*}} & \textit{\tentry{0.67}{0.07}\textsuperscript{\textbf{*}}} & \textbf{\tentry{0.65}{0.07}\textsuperscript{\textbf{*}}} & \textit{\tentry{0.67}{0.05}\textsuperscript{\textbf{*}}} & \textit{\tentry{0.62}{0.08}\textsuperscript{\textbf{*}}} & \textit{\tentry{0.74}{0.15}\textsuperscript{\textbf{*}}} & \tentry{0.64}{0.11}\textsuperscript{\textbf{*}} \\
        \midrule
        \multirow{11}{20mm}{\textbf{AUPR}}
        & \multicolumn{1}{l}{Random} & 0.72 & 0.49 & 0.19 & 0.23 & 0.51 & 0.26 & 0.16 & 0.13 & 0.14 \\
        \cmidrule{2-11} & \multicolumn{1}{l}{CBM-seq} & \tentry{0.75}{0.03} & \textit{\tentry{0.58}{0.05}\textsuperscript{\textbf{*}}} & \tentry{0.34}{0.09}\textsuperscript{\textbf{*}} & \tentry{0.24}{0.05} & \tentry{0.64}{0.04}\textsuperscript{\textbf{*}} & \tentry{0.35}{0.06}\textsuperscript{\textbf{*}} & \tentry{0.18}{0.05} & \tentry{0.19}{0.07} & \tentry{0.15}{0.03} \\
        & \multicolumn{1}{l}{CBM-joint} & \tentry{0.77}{0.05} & \tentry{0.47}{0.04} & \textit{\tentry{0.37}{0.09}\textsuperscript{\textbf{*}}} & \tentry{0.25}{0.06} & \tentry{0.64}{0.05}\textsuperscript{\textbf{*}} & \tentry{0.30}{0.07} & \tentry{0.17}{0.04} & \tentry{0.18}{0.06} & \tentry{0.18}{0.08} \\
        \cmidrule{2-11} & \multicolumn{1}{l}{MVCBM-seq-avg} & \tentry{0.75}{0.05} & \textit{\tentry{0.58}{0.06}\textsuperscript{\textbf{*}}} & \textbf{\tentry{0.42}{0.07}\textsuperscript{\textbf{*}}} & \tentry{0.33}{0.06}\textsuperscript{\textbf{*}} & \tentry{0.53}{0.05} & \tentry{0.41}{0.08}\textsuperscript{\textbf{*}} & \tentry{0.21}{0.05} & \tentry{0.13}{0.05} & \textit{\tentry{0.24}{0.12}} \\
        & \multicolumn{1}{l}{MVCBM-seq-LSTM}   & \textit{\tentry{0.91}{0.04}\textsuperscript{\textbf{*}}} & \tentry{0.55}{0.04}\textsuperscript{\textbf{*}} & \tentry{0.33}{0.08}\textsuperscript{\textbf{*}} & \textbf{\tentry{0.40}{0.06}\textsuperscript{\textbf{*}}} & \textit{\tentry{0.65}{0.03}\textsuperscript{\textbf{*}}} & \textbf{\tentry{0.50}{0.11}\textsuperscript{\textbf{*}}} & \textit{\tentry{0.23}{0.05}\textsuperscript{\textbf{*}}} & \textit{\tentry{0.27}{0.05}\textsuperscript{\textbf{*}}} & \textbf{\tentry{0.26}{0.07}\textsuperscript{\textbf{*}}} \\
        & \multicolumn{1}{l}{MVCBM-joint-avg} & \tentry{0.74}{0.06} & \tentry{0.51}{0.07} & \textbf{\tentry{0.42}{0.09}\textsuperscript{\textbf{*}}} & \tentry{0.28}{0.07} & \tentry{0.59}{0.06}\textsuperscript{\textbf{*}} & \tentry{0.35}{0.05}\textsuperscript{\textbf{*}} & \tentry{0.22}{0.06} & \tentry{0.22}{0.13} & \tentry{0.21}{0.08} \\
        & \multicolumn{1}{l}{MVCBM-joint-LSTM} & \textbf{\tentry{0.92}{0.02}\textsuperscript{\textbf{*}}} & \tentry{0.49}{0.05} & \textit{\tentry{0.37}{0.11}\textsuperscript{\textbf{*}}} & \tentry{0.32}{0.07}\textsuperscript{\textbf{*}} & \textit{\tentry{0.65}{0.06}\textsuperscript{\textbf{*}}} & \tentry{0.39}{0.07}\textsuperscript{\textbf{*}} & \tentry{0.20}{0.06} & \tentry{0.17}{0.07} & \tentry{0.21}{0.06}\textsuperscript{\textbf{*}} \\
        \cmidrule{2-11} & \multicolumn{1}{l}{SSMVCBM-avg} & \tentry{0.73}{0.05} & \textit{\tentry{0.58}{0.07}\textsuperscript{\textbf{*}}} & \tentry{0.36}{0.05}\textsuperscript{\textbf{*}} & \tentry{0.28}{0.04}\textsuperscript{\textbf{*}} & \tentry{0.53}{0.05} & \tentry{0.37}{0.09}\textsuperscript{\textbf{*}} & \tentry{0.20}{0.06} & \tentry{0.13}{0.02} & \textit{\tentry{0.24}{0.06}\textsuperscript{\textbf{*}}} \\
        & \multicolumn{1}{l}{SSMVCBM-LSTM} & \tentry{0.88}{0.06}\textsuperscript{\textbf{*}} & \textbf{\tentry{0.60}{0.06}\textsuperscript{\textbf{*}}} & \textbf{\tentry{0.42}{0.06}\textsuperscript{\textbf{*}}} & \textit{\tentry{0.39}{0.09}\textsuperscript{\textbf{*}}} & \textbf{\tentry{0.67}{0.07}\textsuperscript{\textbf{*}}} & \textit{\tentry{0.43}{0.10}\textsuperscript{\textbf{*}}} & \textbf{\tentry{0.24}{0.08}} & \textbf{\tentry{0.30}{0.13}\textsuperscript{\textbf{*}}} & \tentry{0.20}{0.05}\textsuperscript{\textbf{*}} \\
        \bottomrule
    \end{tabular}
    \end{center}
\end{table*}

Across all target variables, most concepts could be predicted by at least one of the models significantly better than by a fair coin flip (one-sample two-sided $t$-test $p$-value $<0.05$, adjusted using the Benjamini--Yekutieli procedure with the FDR of $q=0.05$). 
Surprisingly, some of the variables with relatively few cases present in the dataset could be captured by some models, e.g. \emph{coprostasis} ($c_8$) and \emph{meteorism} ($c_9$) by the LSTM-based variants of MVCBM and SSMVCBM. 
On the other hand, the \emph{thickening of the bowel wall} ($c_7$) was particularly challenging to model, likely due to its low prevalence and the lack of predictive power in the downstream classification task: some models trained with the severity as the target were able to perform significantly better than random, as shown in Table~\ref{tab:appendicitis_concept_res_severity}.

Note that, in a few cases, some models achieved average \mbox{AUROCs} below the expected performance of a fair coin flip (Table~\ref{tab:appendicitis_concept_res}), e.g. both sequentially and jointly optimized CBMs attained an AUROC close to 0.40 for predicting \emph{meteorism}. Such performance is attributable to the sparsity of some concept variables; for instance, only 15\% of subjects had a positive label for meteorism (Table~\ref{tab:concept_distribution}). Another factor is the use of weighted loss functions for the concept and target prediction (Eqs.~(\ref{eqn:opt_seq_1})--(\ref{eqn:opt_joint}) and (\ref{eqn:opt_ssmvcbm_minimax})). Consequently, the models may over-predict the minority class and perform worse than a fair coin flip. 

Predictably, sequentially optimized models (seq) were more performant at the concept prediction than the ones optimized jointly (joint), in agreement with the findings reported in the literature \citep{Koh2020}. 
Similar to the experiments on the synthetic data shown in Figure~\ref{fig:synthetic_res}(b), the models aggregating multiple views tended to have higher AUROCs and AUPRs. 
However, by contrast, LSTM-based aggregation consistently and noticeably outperformed simple averaging (avg), especially for predicting the visibility of the appendix---one of the most important diagnostic concepts \citep{Marcinkevics2021}. 
This could be associated with the loose spatiotemporal ordering among the US images acquired for each subject. 
Last but not least, semi-supervised bottlenecks were comparable to the sequentially optimized MVCBMs. 
Thus, learning complementary representations disentangled from the concepts did not hurt the model's performance at concept prediction.

In addition to the discriminative power, we assessed the calibration of the concept predictions. The test-set Brier scores across the three targets are reported in Appendix~\ref{app:brier_scores}, Table~\ref{tab:appendicitis_concept_res_brier_scores}. Overall, similar to the findings above, multiview models attained lower Brier scores for most concept variables than the single-view CBMs. The cases wherein single-view CBMs performed better than their multiview counterparts may be attributed to the imbalances in concept distributions and the fact that the Brier score does not adjust for such situations. For instance, for very sparse response variables, a classifier trivially predicting the most frequent category would achieve a relatively low Brier score. Although many models predicted several concepts significantly better than the constant prediction of 0.5, their Brier scores were mainly in the range of 0.18-0.23, which is not considerably below the baseline of 0.25.

\paragraph{Predicting diagnosis, management, and severity}
As mentioned, the end goal of the developed models was the prediction of the \mbox{(i) diagnosis}, \linebreak \mbox{(ii) management,} and \mbox{(iii) severity} among suspected appendicitis patients based on the multiview US images.
Test-set performance for these three target variables is reported in Table~\ref{tab:appendicitis_target_res}. 

\begin{table*}[h!]
    \scriptsize
    \centering
    \caption{Models' test-set performance at predicting diagnosis, management, and severity. Test-set AUROCs, AUPRs, and Brier scores are reported as averages and standard deviations across ten independent initializations. Bold indicates the best result; italics indicates the second best.}
    \label{tab:appendicitis_target_res} 
    \begin{tabular}{lm{0.8cm}m{0.8cm}m{0.8cm}cm{0.8cm}m{0.8cm}m{0.8cm}cm{0.8cm}m{0.8cm}m{0.8cm}}
        \toprule
        \multirow{2}{*}{\textbf{Model}} & \multicolumn{3}{c}{\textbf{Diagnosis}} &  & \multicolumn{3}{c}{\textbf{Management}} &  & \multicolumn{3}{c}{\textbf{Severity}}\\
        \cmidrule{2-4}\cmidrule{6-8}\cmidrule{10-12} & \multicolumn{1}{c}{\textbf{AUROC}} & \multicolumn{1}{c}{\textbf{AUPR}} & \multicolumn{1}{c}{\textbf{Brier}} &  & \multicolumn{1}{c}{\textbf{AUROC}} & \multicolumn{1}{c}{\textbf{AUPR}} & \multicolumn{1}{c}{\textbf{Brier}} &  & \multicolumn{1}{c}{\textbf{AUROC}} & \multicolumn{1}{c}{\textbf{AUPR}} & \multicolumn{1}{c}{\textbf{Brier}} \\
        \toprule
        Random & 0.50 & 0.75 & 0.25 &  & 0.50 & 0.47 & 0.25 &  & 0.50 & 0.23 & 0.25 \\
        \midrule
        Radiomics + RF & \tentry{0.64}{0.02} & \tentry{0.82}{0.01} & \tentry{0.22}{0.00} &  & \tentry{0.65}{0.01} & \tentry{0.60}{0.02} & \textit{\tentry{0.24}{0.00}} &  & \textit{\tentry{0.77}{0.02}} & \textit{\tentry{0.58}{0.04}} & \textbf{\tentry{0.15}{0.00}} \\
        \midrule
        ResNet-18 & \tentry{0.70}{0.07} & \tentry{0.88}{0.04} & \tentry{0.25}{0.08} &  & \tentry{0.69}{0.07} & \textit{\tentry{0.71}{0.08}} & \tentry{0.27}{0.05} &  & \tentry{0.73}{0.10} & \tentry{0.52}{0.10} & \tentry{0.18}{0.04} \\
        \midrule
        CBM-seq & \tentry{0.64}{0.06} & \tentry{0.84}{0.04} & \tentry{0.22}{0.02} &  & \tentry{0.68}{0.05} & \tentry{0.68}{0.05} & \textbf{\tentry{0.23}{0.02}} &  & \tentry{0.66}{0.06} & \tentry{0.41}{0.08} & \tentry{0.23}{0.04} \\
        CBM-joint & \tentry{0.62}{0.04} & \tentry{0.83}{0.04} & \tentry{0.24}{0.02} &  & \tentry{0.66}{0.06} & \tentry{0.68}{0.04} & \textbf{\tentry{0.23}{0.02}} &  & \tentry{0.68}{0.06} & \tentry{0.44}{0.08} & \tentry{0.23}{0.02} \\
        \midrule
        MVBM-avg & \textit{\tentry{0.76}{0.05}} & \tentry{0.89}{0.04} & \tentry{0.22}{0.03} &  & \tentry{0.71}{0.04} & \tentry{0.69}{0.04} & \tentry{0.24}{0.02} &  & \tentry{0.71}{0.12} & \tentry{0.59}{0.11} & \tentry{0.20}{0.05} \\
        MVBM-LSTM & \textit{\tentry{0.76}{0.04}} & \textit{\tentry{0.91}{0.02}} & \tentry{0.23}{0.02} &  & \tentry{0.67}{0.04} & \tentry{0.61}{0.04} & \textbf{\tentry{0.23}{0.02}} &  & \tentry{0.74}{0.13} & \textit{\tentry{0.58}{0.12}} & \tentry{0.22}{0.07} \\
        \midrule
        MVCBM-seq-avg & \tentry{0.67}{0.05} & \tentry{0.85}{0.05} & \tentry{0.23}{0.02} &  & \tentry{0.58}{0.05} & \tentry{0.62}{0.06} & \tentry{0.26}{0.02} &  & \tentry{0.75}{0.07} & \tentry{0.56}{0.12} & \tentry{0.23}{0.04} \\
        MVCBM-seq-LSTM & \tentry{0.73}{0.03} & \tentry{0.89}{0.01} & \tentry{0.24}{0.04} &  & \tentry{0.57}{0.03} & \tentry{0.53}{0.04} & \tentry{0.26}{0.01} &  & \tentry{0.70}{0.11} & \tentry{0.48}{0.16} & \tentry{0.21}{0.03} \\
        MVCBM-joint-avg & \tentry{0.66}{0.09} & \tentry{0.84}{0.06} & \tentry{0.24}{0.06} &  & \tentry{0.69}{0.06} & \tentry{0.66}{0.11} & \textbf{\tentry{0.23}{0.02}} &  & \tentry{0.70}{0.06} & \tentry{0.53}{0.11} & \tentry{0.24}{0.02} \\
        MVCBM-joint-LSTM & \tentry{0.72}{0.02} & \tentry{0.88}{0.02} & \tentry{0.22}{0.01} &  & \tentry{0.57}{0.05} & \tentry{0.50}{0.04} & \tentry{0.26}{0.01} &  & \tentry{0.65}{0.07} & \tentry{0.37}{0.10} & \tentry{0.24}{0.02} \\
        \midrule
        SSMVCBM-avg & \textbf{\tentry{0.80}{0.03}} & \textbf{\tentry{0.92}{0.02}} & \textit{\tentry{0.20}{0.03}} &  & \textbf{\tentry{0.72}{0.05}} & \textbf{\tentry{0.72}{0.04}} & \tentry{0.27}{0.05} &  & \tentry{0.73}{0.07} & \tentry{0.57}{0.09} & \textit{\tentry{0.17}{0.02}} \\
        SSMVCBM-LSTM & \textbf{\tentry{0.80}{0.06}} & \textbf{\tentry{0.92}{0.04}} & \textbf{\tentry{0.19}{0.04}} &  & \tentry{0.70}{0.03} & \tentry{0.67}{0.06} & \tentry{0.27}{0.04} &  & \textbf{\tentry{0.78}{0.05}} & \textit{\tentry{0.58}{0.10}} & \tentry{0.21}{0.10} \\
        \bottomrule
    \end{tabular}
\end{table*}

With respect to AUROC and AUPR, all models were able to predict all target variables better than the naive baseline. 
Among the concept-based approaches, multiview models offered a consistent improvement over the vanilla CBM for diagnosis and severity. 
Moreover, the best-performing concept-based classifiers often achieved AUROCs and AUPRs comparable to those of the black-box MVBM. 
For the diagnosis, on average, multiview concept bottlenecks with the LSTM-based fusion outperformed averaging-based approaches. 
However, for management, the opposite was true.
Expectedly, while the LSTM-based fusion was helpful in the pediatric appendicitis dataset where US images are chronologically ordered, at test time, the target prediction performance of the LSTM-based CBMs was sensitive to the order of input images, as observed in the supplementary experimental results in Appendix~\ref{app:lstm}.
For the diagnosis and management prediction, we also observed that neural-network-based methods, overall, outperformed RFs fitted on radiomics features. 
The latter result is not surprising, given that we did not utilize manually segmented regions of interest for radiomics feature extraction. 
Lastly, across all targets, the semi-supervised extension of the MVCBM achieved higher AUROCs and AUPRs or was comparable to the approaches that purely relied on the concepts.

Brier score results partially agree with AUROCs and AUPRs; however, they feature less variability across model classes. For all target variables, most scores are $\geq0.20$. Combined with the reported AUROCs and AUPRs, the latter finding indicates that the probabilistic predictions of the models considered could benefit from calibration, which could help produce more interpretable probabilistic outputs.

Along with the model comparison w.r.t. AUROCs, AUPRs, and Brier scores, we investigated the tradeoff between true positive (TPR) and false positive (FPR) rates in more detail for predicting the diagnosis. Full results are reported in Table~\ref{tab:fpr_at_k} (Appendix~\ref{app:fprs}). In particular, we assessed the models' FPRs for a few fixed satisfactory levels of the TPR. As expected, we observed that, for all approaches, attaining high TPRs led to relatively high FPRs of $>30\%$.

In summary, concept-based classification on multiview US data is encouragingly effective at predicting the diagnosis. 
For management, aggregating multiple US images offers no improvement over simple single-view classification. 
We attribute this to the \emph{diagnostic} nature of the chosen concepts and their limited predictive power for the treatment assignment. 
Likewise, accurately predicting appendicitis severity is challenging, likely, due to the low prevalence of complicated appendicitis cases in the current dataset. 
Last but not least, in all tasks, the proposed SSMVCBM mitigated the poorer discriminative performance of concept-based approaches by learning representations complementary to the probably incomplete concept set.

\paragraph{Interacting with the model}

\begin{figure*}[h!]%
\centering

\subfigure[]{
    \includegraphics[width=0.3\linewidth]{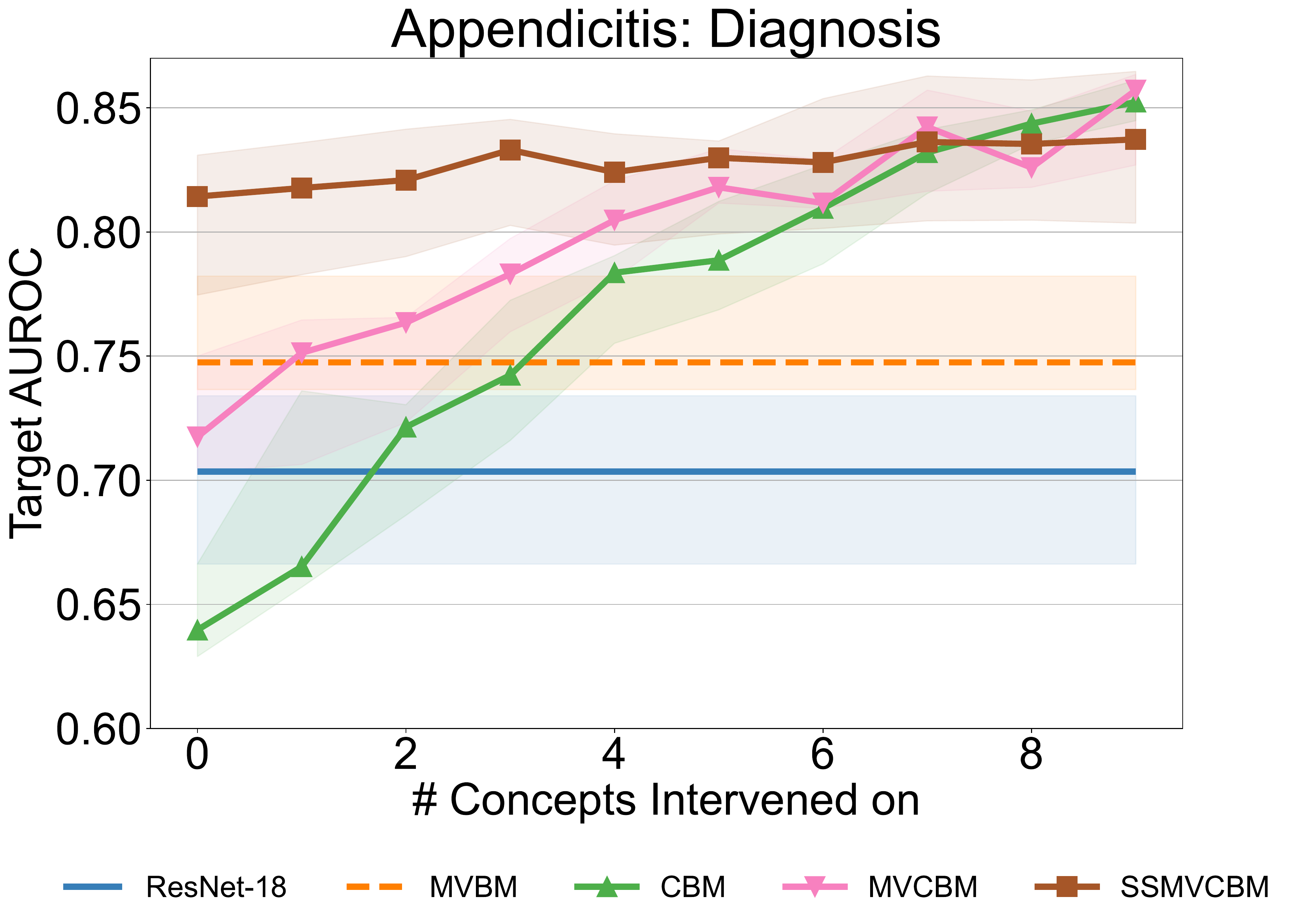}
}
\subfigure[]{
    \includegraphics[width=0.3\linewidth]{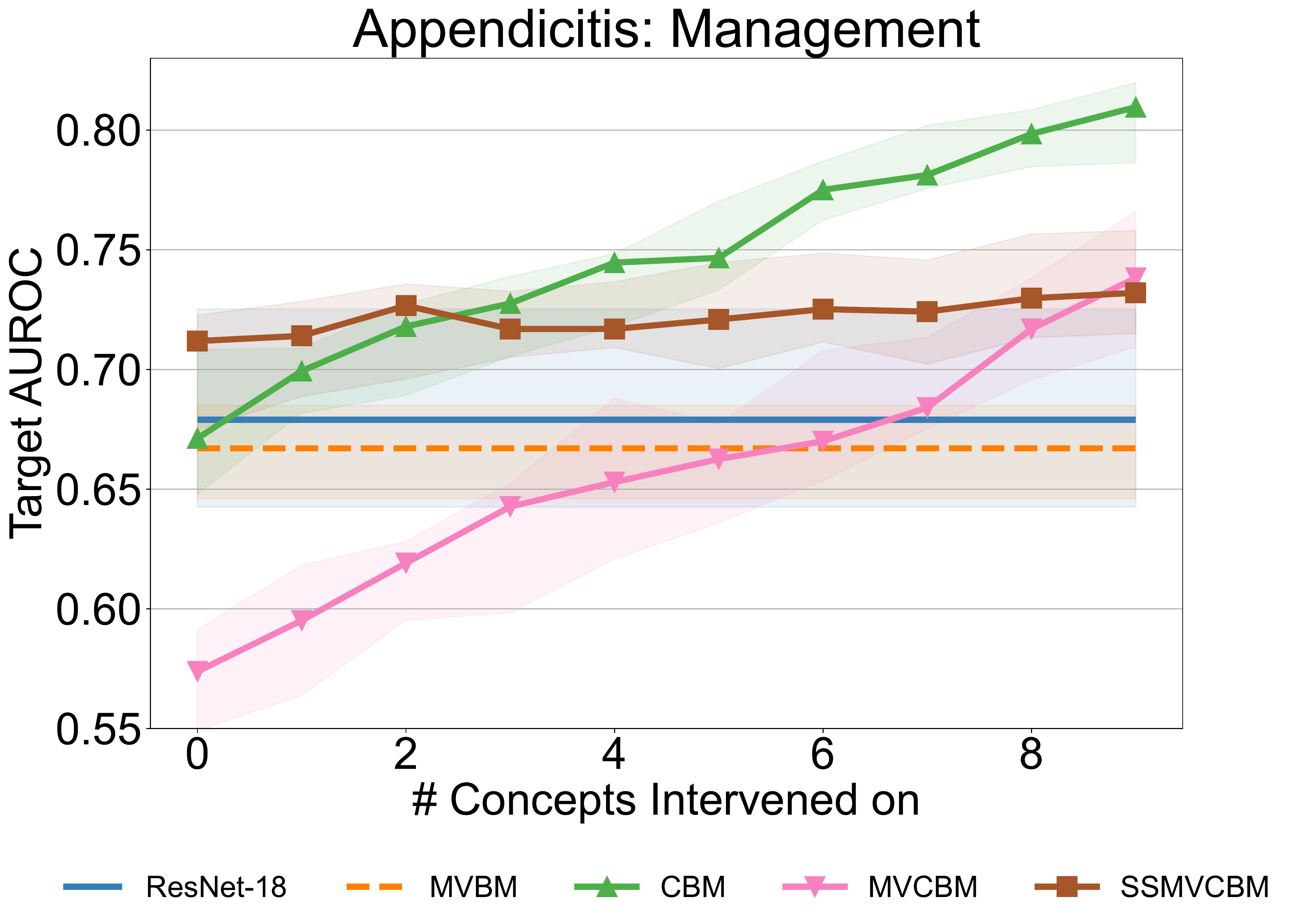}
}
\subfigure[]{
    \includegraphics[width=0.3\linewidth]{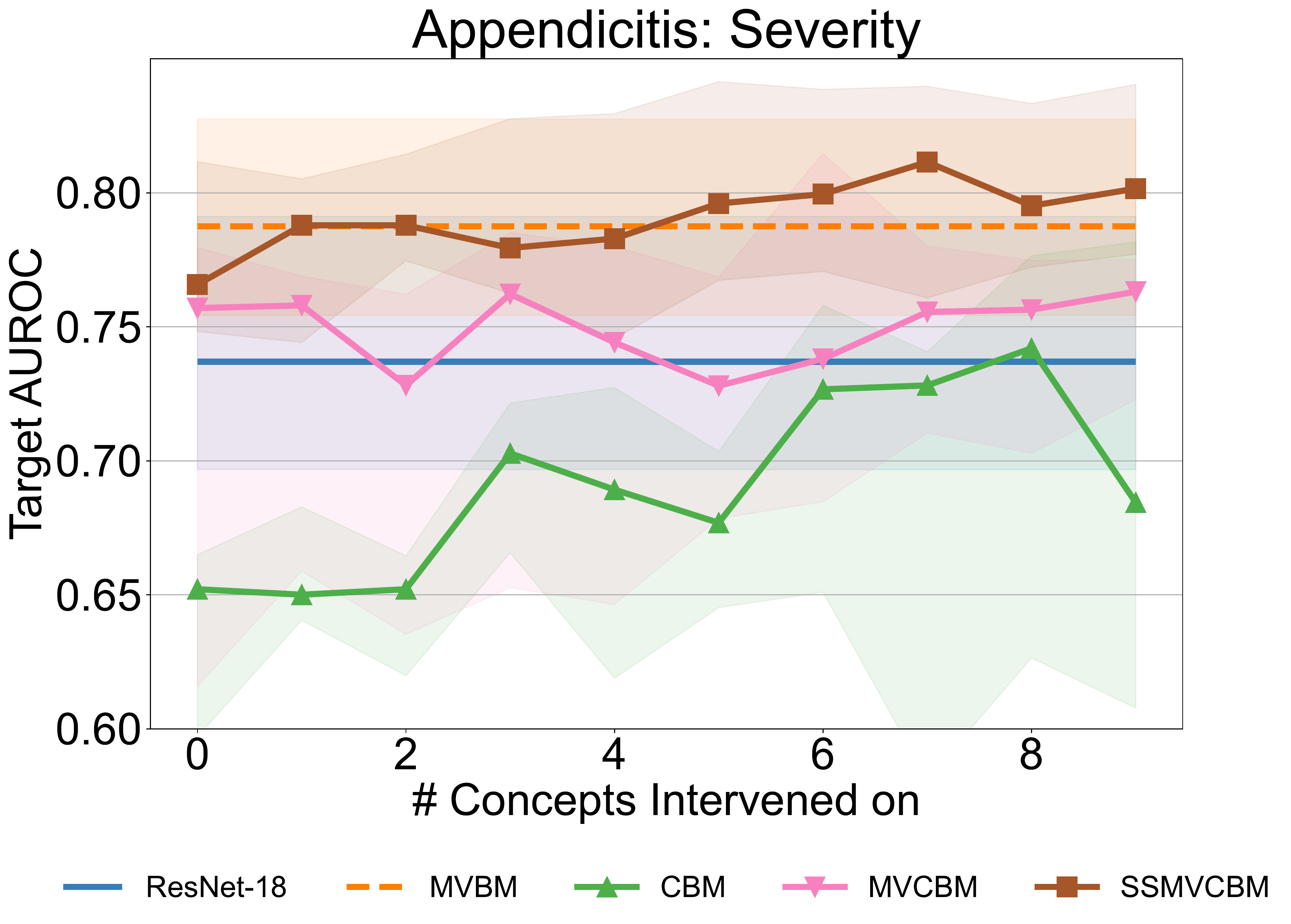}
}

\subfigure[]{
    \includegraphics[width=0.3\linewidth]{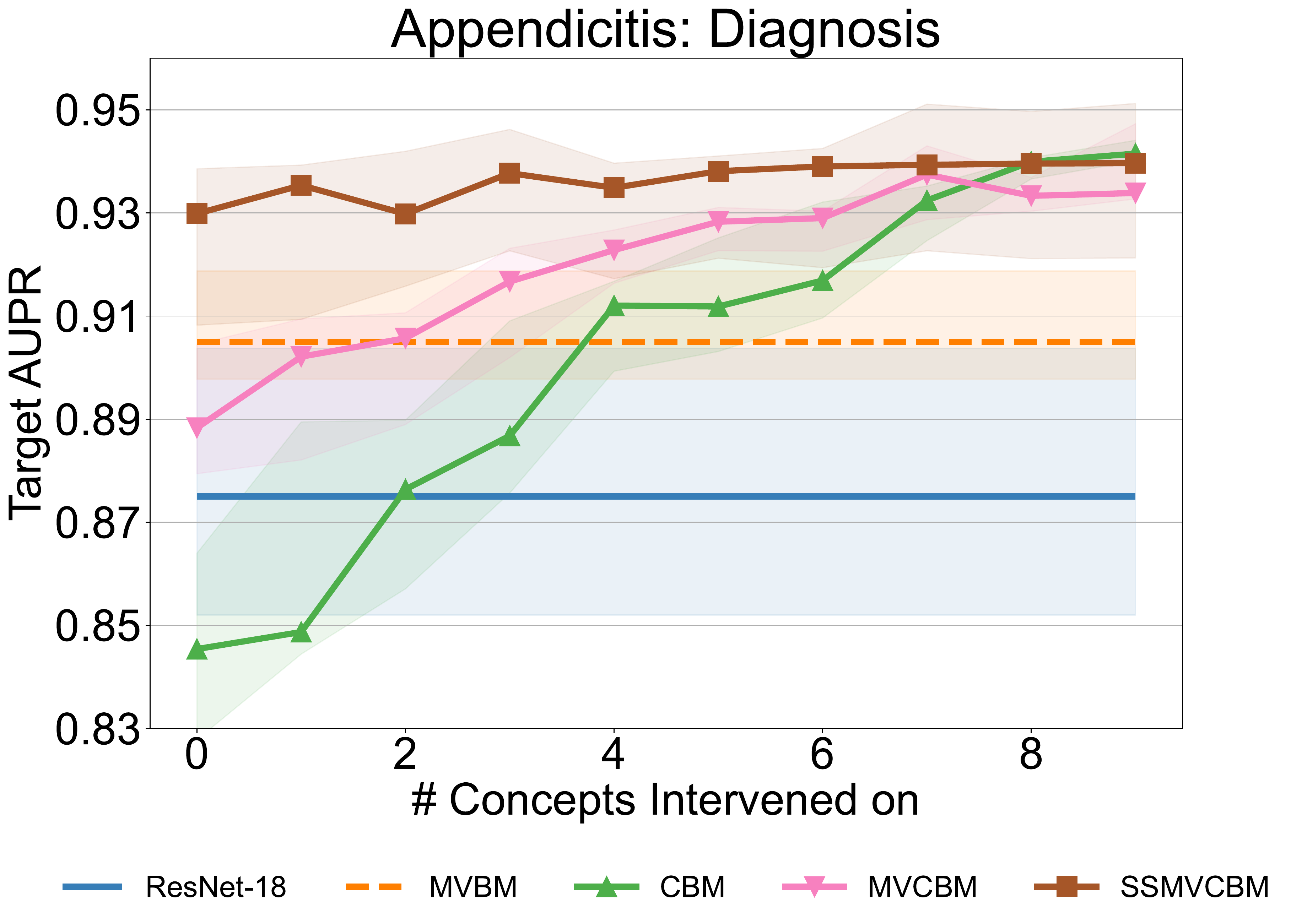}
}
\subfigure[]{
    \includegraphics[width=0.3\linewidth]{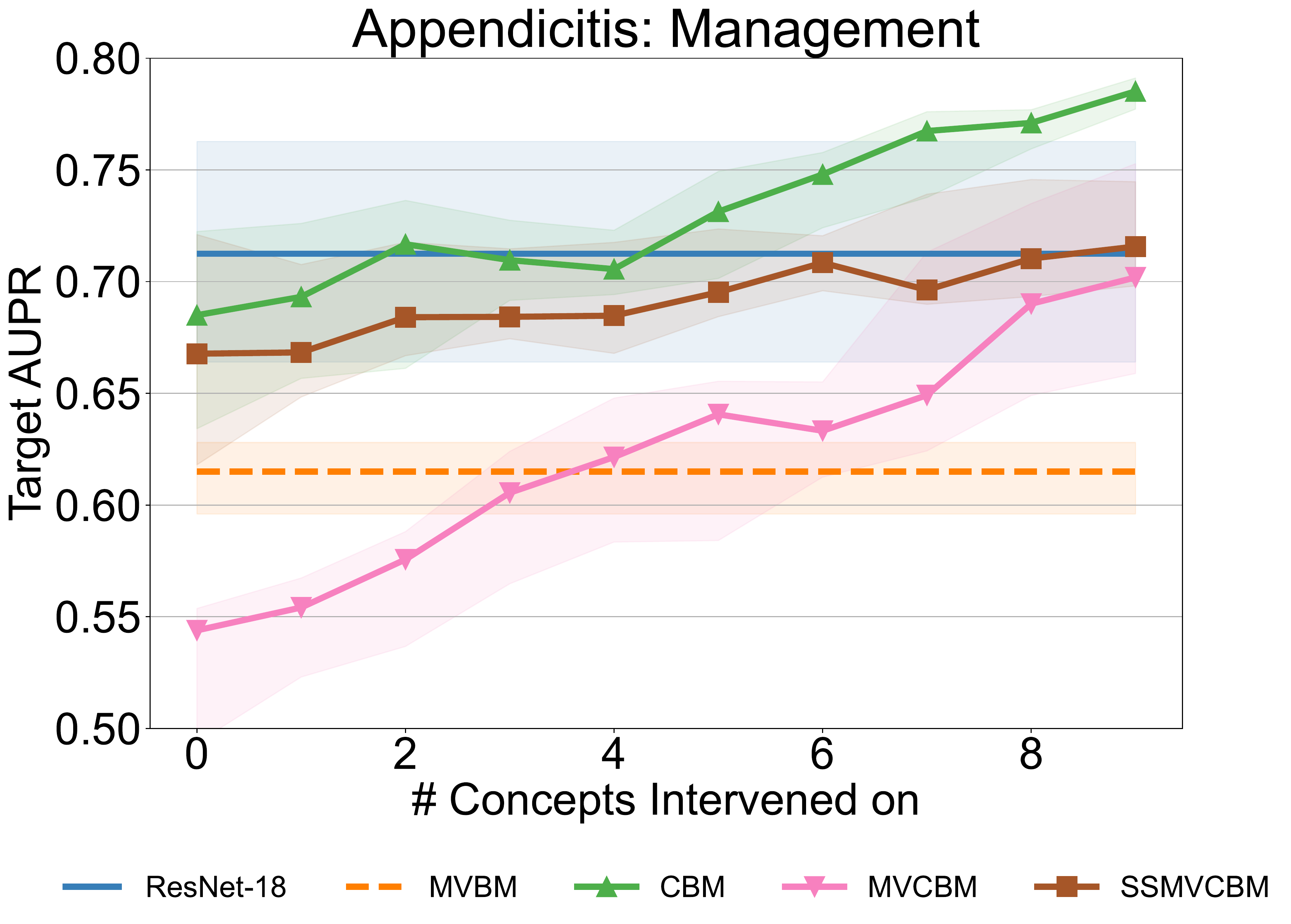}
}
\subfigure[]{
    \includegraphics[width=0.3\linewidth]{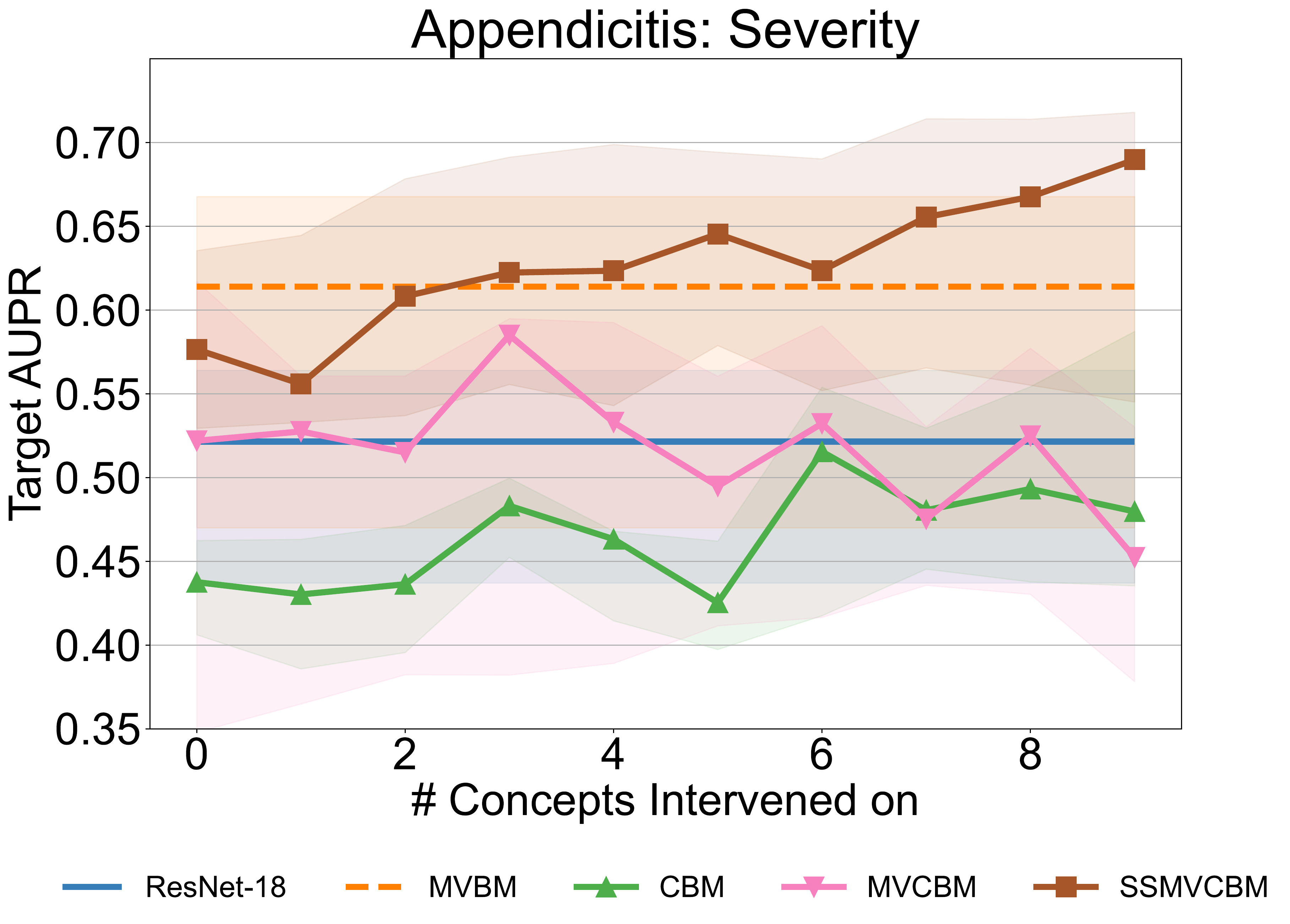}
}

\caption{Intervention experiment results for the pediatric appendicitis dataset. Interventions were performed by replacing the concept values predicted by the concept-based models with the ground truth for the \mbox{(a,d) diagnosis}, \mbox{(b,e) management,} and \mbox{(c,f) severity} as target variables. Lines correspond to median \mbox{(a--c) test-set} AUROCs and \mbox{(d--f) AUPRs} attained by intervened models across ten initializations and three randomly sampled concept subsets. The performances of non-intervenable ResNet-18 and MVBM baselines are shown as horizontal lines.}
\label{fig:intervention_res}
\end{figure*}

The practical utility of CBMs lies in the ability of the human user, in the current use case, the physician, to intervene on the concepts predicted by the model, thus affecting the model's behavior at test time. 
Similarly to the proof-of-concept experiments, we intervened on the bottleneck layers of the  CBM, MVCBM, and SSMVCBM trained on the pediatric appendicitis data. 
Figure~\ref{fig:intervention_res} summarizes these results. 
Since LSTM-based and sequentially trained classifiers generally captured the concepts better (Table~\ref{tab:appendicitis_concept_res}), we only considered this specific configuration.
Figure~\ref{fig:intervention_res} shows the effect of interventions on the three models for the diagnosis (Figures~\ref{fig:intervention_res}(a) and \ref{fig:intervention_res}(d)), management (Figures~\ref{fig:intervention_res}(b) and \ref{fig:intervention_res}(e)), and severity (Figures~\ref{fig:intervention_res}(c) and \ref{fig:intervention_res}(f)). 
The lines show changes in AUROCs and AUPRs when intervening on randomly chosen concept subsets of varying sizes. 

For the diagnosis, the intervention effect is similar to the behavior of the models on the synthetic data shown in Figure~\ref{fig:synthetic_res}(c). 
Namely, AUROC and AUPR increase steadily with the number of concepts intervened on: for all models, the maximum median AUROC and AUPR achieved are approx. 0.85 and 0.94, respectively. 
Being the best-performing model (Table~\ref{tab:appendicitis_target_res}), SSMVCBM demonstrates only a slight increase in median predictive performance after intervening on the full concept set. 

Similarly, for management, we observed an increase in AUROC and AUPR. 
However, for predicting this target, a single-view CBM performed surprisingly well and overtook multiview models after interventions. 
Last but not least, interventions yielded no visible performance improvement for severity, possibly, due to considerable variance across initializations and randomly sampled concept subsets. 

\subsection{Online Prediction Tool}

\begin{sidewaysfigure*}[ph!]%
\centering
\includegraphics[width=1.0\textwidth]{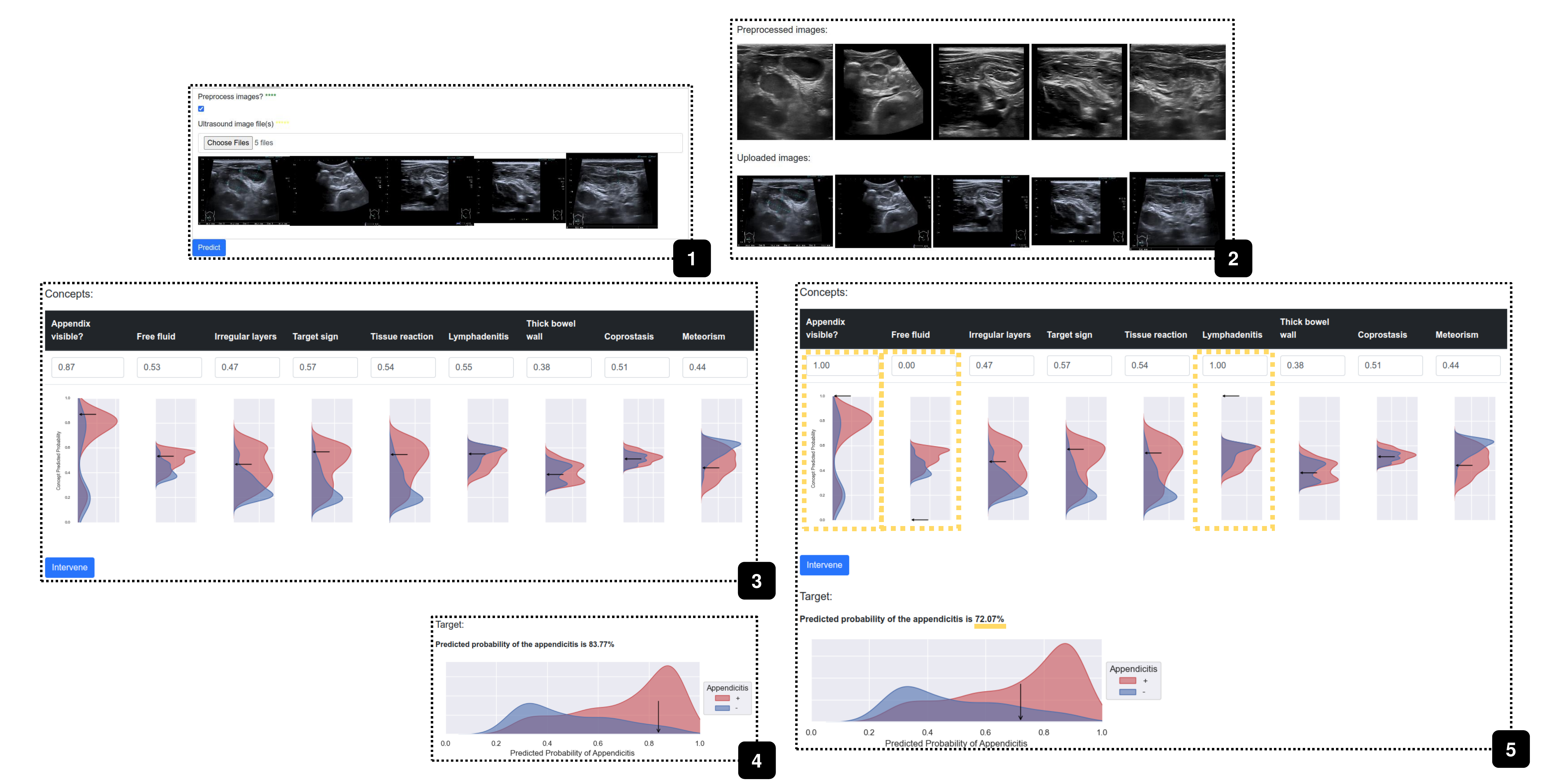}
\caption{An illustrated use case for the pediatric appendicitis online prediction tool. \mbox{(1) The} user uploads input ultrasound images corresponding to a single patient. \mbox{(2) Optionally,} preprocessing is performed, and the tool displays original and preprocessed US images. \mbox{(3) The} tool displays predicted concept values, given by sigmoid activations from the corresponding units in the concept bottleneck layer, alongside predicted value histograms obtained from the \emph{training} data (plotted separately for appendicitis and non-appendicitis cases in red and blue, respectively). The user can compare current concept predictions to those made for labeled \emph{training} data points. \mbox{(4) The} tool shows the prediction for the target variable, i.e. for the diagnosis, given by the sigmoid activation. \mbox{(5) The} user may choose to intervene on the concept predictions and, thus, affect the target prediction. For example, if the user was trained in interpreting ultrasound images, they may correct wrongly predicted concepts by setting corresponding variables to 0 (negative) or 1 (positive). In this example, concept predictions that were intervened on (chosen arbitrarily for demonstration) are indicated by yellow dotted lines.}\label{fig:tool_gui}
\end{sidewaysfigure*}

As a first step towards enabling clinicians and other interested parties to benefit from ML-based decision support, we developed and published an online decision support tool based on the abovementioned methods, available at \url{https://papt.inf.ethz.ch/mvcbm}. 
The use case is illustrated in Figure~\ref{fig:tool_gui}. 
The tool utilizes the multiview CBM model (Figure~\ref{fig:workflow}) for predicting the diagnosis in suspected appendicitis patients. 
The user may upload several ultrasonography images, each representing a different view of the same patient. 
Image preprocessing, described in the \emph{Methods} section and demonstrated in  Figure~\ref{fig:us_example}, may be optionally executed. 
In addition to predicting the diagnosis, the tool allows the user to intervene on the concept predictions (Table~\ref{tab:concept_distribution}) by setting corresponding sigmoid activations to 0 (negative) or 1 (positive). 
Uploaded images are protected using server-side sessions, which are only temporarily stored on the server and are purged after 30 minutes. 
See Appendix~\ref{app:tool} for more information.

\section{Discussion}\label{sec:discussion}
Most of the prior work on using ML for appendicitis has focused on tabular datasets with handcrafted features \citep{Hsieh2011,Deleger2013,Reismann2019,Aydin2020,Akmese2020,Stiel2020,Marcinkevics2021,RoigAparicio2021,Xia2022} or more invasive imaging modalities, such as computed tomography \citep{Rajpurkar2020}. 
This work takes the first step towards the computer-aided diagnosis of appendicitis based on abdominal ultrasound, a noninvasive, accessible, and cheap technique. 
Moreover, to facilitate the replication of our results and allow for comparison with new methods, we made our anonymized dataset publicly accessible. 
It includes laboratory, physical exam, clinical, and US data from 579 patients. 
In addition, for demonstratory purposes, we deployed the MVCBM model for the diagnosis as an easy- and free-to-use web tool. 

Although appendicitis is a common condition in the pediatric population, diagnosing it and choosing the best therapeutic option is challenging. 
Early differentiation between simple and complicated, necrotizing appendicitis is crucial for effective management and prognosis \citep{Reddan2016,Reismann2019,Kiss2021}. 
The advances in US resolution, especially with the high-frequency sonography, support the detection of a normal appendix and the identification of indirect appendicitis signs, such as surrounding tissue inflammation and the reaction of the intestinal bowel wall \citep{Park2011}. 
ML-based decision support tools may further increase diagnostic accuracy and prove pivotal in improving treatment outcomes. 
The results of the current study are promising, as they suggest that direct interpretation of US images by ML models is a feasible goal. 
Predictive models, such as the ones developed in this study, may assist physicians in interpreting acquired US images and may even enable comparison of the results with the newly conducted US exams to characterize the progress or resolution of the inflammation.

Moreover, this work presents an improvement upon traditional concept bottleneck models \citep{Koh2020}, making them more readily applicable to medical imaging datasets where multiple images or modalities may be observable for each subject. 
In order to accomplish this, we proposed a practical architecture based on the hybrid fusion approach \citep{Baltrusaitis2019}, which can effectively handle varying numbers of views per data point, partial observability of the concepts from individual images, and the incorporation of spatial or temporal ordering. 
While prior research has explored the use of averaging and LSTM techniques for aggregating representations \citep{Havaei2016,Ma2019}, our focus is specifically on interpretable models, particularly those involving concept-based classification. 
To the best of our knowledge, this problem setting has not been previously discussed in the literature despite its relevance to biomedical applications \citep{Wang2020,Qian2021}. 

Another scenario that we studied, similarly pertinent to applications, is when the concept set given to a CBM is insufficient \citep{Yeh2020}, i.e. does not entirely capture the predictive relationship between the covariates and the target. 
To address this issue and improve the CBM's predictive performance, our model learns additional representations complementary to the concepts, \mbox{i.e.} de-correlated from the concepts yet helpful in the downstream prediction problem. 
To achieve this objective, we modified the model's architecture, incorporated an adversarial regularization term into the loss function, and adapted the training procedure accordingly.

A few previous works have investigated related limitations of the CBMs when the concept set provided to the CBM proves insufficient, and have explored alternative model designs. 
For instance, \citet{Sawada2022} combined CBMs with self-explaining neural networks to learn additional unsupervised concepts; however, they did not investigate the disentanglement of the given and learned concepts or the intervenability of their extended bottleneck layer. 
\citet{Yuksekgonul2022} proposed fitting a concept bottleneck \emph{post hoc} for a pretrained backbone and utilized residual fitting to compensate for an incomplete concept set. 
Moreover, they investigated the global model edition, e.g. to mitigate the classifier's reliance on spurious correlation. 
In contrast, our work assumes an \emph{ante hoc} modeling scenario and focuses on the local, i.e. single-data-point, interventions. 
Another related line of research also studied the problem of unobserved concepts and concept leakage \citep{Marconato2022}, employing generative representation learning, which may be challenging to apply to smaller datasets in practice. 
The most closely related is the concurrent work by \citet{Havasi2022}, who extended the standard CBM architecture with a side channel to learn latent concepts and compensate for insufficiency. 
While their method is similar to ours, it does not address multiview learning or consider medical imaging data.

In our experiments, we have demonstrated the feasibility of the proposed models and the benefits of the multiview and semi-supervised concept-based approach on synthetic and medical image data. 
Our findings have shown that the MVCBM and SSMVCBM models have generally outperformed vanilla CBM in terms of both concept and target prediction. 
Moreover, based on the US data, we have developed predictive models for appendicitis, its severity and the management of pediatric patients with abdominal pain  (Tables~\ref{tab:appendicitis_concept_res}--\ref{tab:appendicitis_target_res}). 
Our results suggest that, for the diagnosis, multiview concept bottlenecks can achieve comparable performance to black-box models while allowing medical practitioners to interpret and intervene on the predictions. 
For management and severity, we observed somewhat inconclusive results with little difference across the single- and multiview classifiers. 
We attribute the latter to the limited predictive power of the ultrasonographic findings for these targets \citep{Marcinkevics2021}, the diagnostic nature of the chosen concepts and the overall moderate size of the training set. For instance, it had been previously shown that the most important predictor of the treatment assignment is peritonitis/abdominal guarding \citep{Marcinkevics2021} assessed during a clinical examination. Among the US findings, most other predictively useful attributes can be identified based on the RLQ image alone. Therefore, we hypothesize that the additional views, e.g. depicting pathological lymph nodes or meteorism, are not as helpful for the management classification. This observation might explain the relatively worse performance of the multiview approaches for this target variable.

Nevertheless, the current study exhibits certain limitations with regard to its design, experimental setup, and proposed methods.
The appendicitis dataset represents a moderately-sized and relatively homogeneous patient cohort recruited from a single clinical center over a short time (between 2016 and 2021). 
Hence, in order to further validate predictive models, an external validation is necessary using data from diverse US devices, clinical centers, and countries. 
Another limitation is the lack of histologically confirmed diagnoses among the conservatively treated patients. 
This implies that the model validation and comparison results presented above must be interpreted cautiously since we do not have access to the true disease status for all subjects. The image preprocessing pipeline could be improved further: currently, we discard scale information in the US images, making it impossible to detect the appendix diameter, a relevant sonographic sign of appendicitis \citep{Reddan2016}.
Lastly, concepts could be modeled in a more fine-grained manner to incorporate physicians' uncertainty. Instead of just differentiating between the lack or presence of a finding, intermediate concept categories could be included by, for example, collecting data from multiple raters and considering discrepancies among them.

From the methodological perspective, we currently have a limited theoretical understanding of the (SS)MVCBMs. 
In particular, it would be desirable to explore the representations learned by SSMVCBMs and the identifiability of the ground-truth generative factors. 
Moreover, in the current implementation, it is not trivial to interpret the representations; thus, additional regularization may be necessary, such as rendering these representations disentangled.

Another potential improvement would be adopting a probabilistic approach to the concept and target variable prediction, facilitating more principled uncertainty estimation. As evidenced by the experiments, our predictive models could benefit from calibration. Explicit uncertainty modeling would allow for better-calibrated and more interpretable probabilistic predictions that could be utilized downstream to perform selective classification \citep{Geifman2017} and uncertainty-based concept interventions \citep{Shin2023}. In practice, uncertainty in concept predictions could be modeled by adapting the proposed architecture with the modules from the stochastic segmentation networks \citep{Monteiro2020} or probabilistic concept bottlenecks \citep{Kim2023}.

\section{Conclusion and Outlook}\label{sec:conclusion}

Motivated by the demand for model interpretability in biomedical applications, we investigated the use of concept bottleneck models for predicting the diagnosis, management and severity among pediatric patients with suspected appendicitis, leveraging abdominal ultrasound images.
The densely annotated dataset used to develop the predictive models was made publicly available, and one of the models was deployed as a freely available demo web tool (\url{https://papt.inf.ethz.ch/mvcbm}).
Methodologically, we introduced several enhancements to the conventional concept-based classification approach. 
Our proposed models can handle multiple views of the object of interest and insufficient concept sets. 
Overall, our experimental results suggest that the proposed methods can deliver competitive performance, while offering an alternative to black-box deep learning models and allowing for real-time interaction with the end user.

In future work, we aim to address several limitations outlined above. 
We plan to validate the predictive models externally on the data from a hospital located in another country. 
Various model design alterations, such as other choices of learnable fusion, further regularization of the learned representations, and uncertainty quantification, are also to be considered. 
Moreover, we recognize the significance of extending our investigation beyond the retrospective study.
For instance, it would be interesting to explore the use of active learning to decide on the acquisition of US images and concept labels for each subject. 
From the clinical perspective, developed models should be extended to incorporate clinical and laboratory parameters and consider other conditions, such as COVID-19, during appendicitis. 
Additionally, we anticipate that using more refined definitions of the target variables could provide more insightful results, e.g. differentiating between subacute and acute appendicitis for the diagnosis and predicting the risk of secondary appendectomy for the management. 
Adjustments in the model architecture and the acquisition of a larger training dataset will facilitate the incorporation of the color Doppler images in the analysis, potentially making the prediction of the disease severity progression more accurate.
\section*{Data availability}

The anonymized data are available on Zenodo at \url{https://doi.org/10.5281/zenodo.7711412}, and the code can be found in a GitHub repository at \url{https://github.com/i6092467/semi-supervised-multiview-cbm}.

\printbibliography
\section*{Acknowledgments}

We thank the members of the Medical Data Science group at ETH Zurich for stimulating discussions and feedback. Our gratitude goes to Dr. Johanna Joe and Dr. Markus Ebert from the Ultrasonography Center, KUNO, St. Hedwig Clinic Regensburg, for their help with the data acquisition. We are also sincerely thankful to Marcel Buehring and the IT Service Group of the Department of Computer Science, ETH Zurich, for their assistance with the deployment of the online prediction tool. RM was supported by the SNSF grant \#320038189096, EO was supported by the SNSF grant P500PT-206746. This preprint was created using the LaPreprint template\footnote{\url{https://github.com/roaldarbol/lapreprint}} by \mbox{Mikkel Roald-Arb\o l \textsuperscript{\orcidlink{0000-0002-9998-0058}}.}

\section*{Competing interests}

The authors declare no competing interests.

\if@endfloat\clearpage\processdelayedfloats\clearpage\fi


\newpage
\begin{appendices}

\begin{appendix}

\makeatletter
\makeatother

\counterwithin{figure}{section}
\counterwithin{table}{section}
\counterwithin{equation}{section}
\counterwithin{algorithm}{section}

{\Huge \color{darkColour}\noindent\textbf{Supplementary Material}}

\bigskip

\section{Pediatric Appendicitis Dataset\label{app:appendicitis_data}}


The study was approved by the Ethics Committee of the University of Regensburg (no. 18-1063-101, 18-1063\_1-101, and 18-1063\_2-101) and was performed following applicable guidelines and regulations. 
The ethics committee confirmed that there was no need for written informed consent for the retrospective analysis and publication of anonymized routine data according to Art. 27 para. 4 of the Bavarian Hospital Law. 
For patients followed up after discharge, written informed consent was obtained from parents or legal representatives.

\begin{figure}[H]
    \centering
    \begin{center}
        \includegraphics[width=0.725\textwidth]{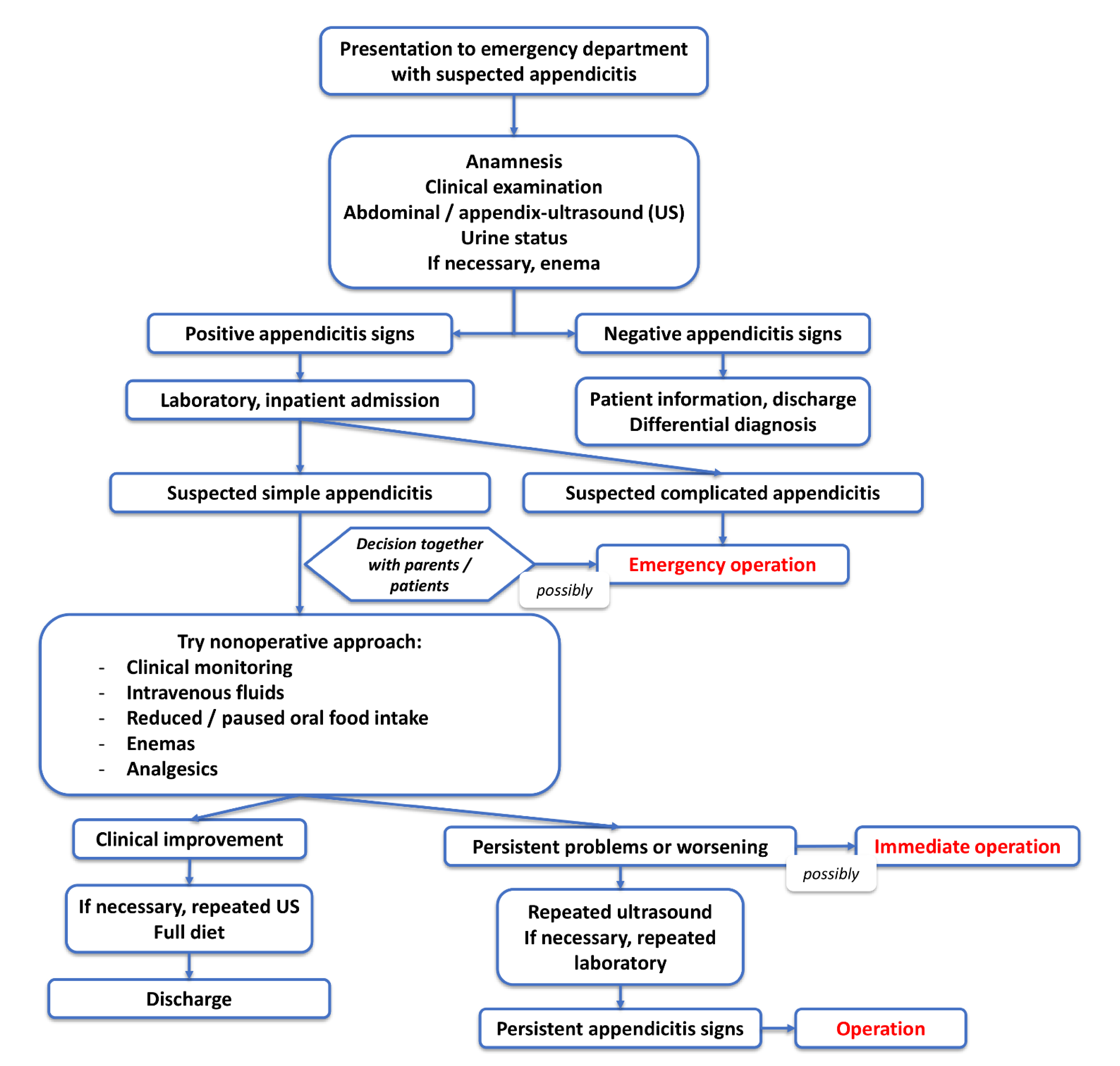}
    \end{center}
    \caption{A schematic of the routine management procedure for patients with suspected appendicitis. The decision on conservative management and operation was made by a senior pediatric surgeon based on clinical presentation combined with laboratory and ultrasound findings. This procedure is exercised at the Department of Pediatric Surgery and Pediatric Orthopedics, Hospital St. Hedwig of the Order of St. John of God, University Children's Hospital Regensburg (KUNO), Regensburg, Germany, where the dataset was acquired. The figure was adapted from the Supplementary Material of the article by \citet{Marcinkevics2021}. \label{fig:app_management_algo}}
\end{figure}

As mentioned, this study presents a retrospective analysis. 
The patients included in the cohort were managed according to the procedure 
summarized in Figure~\ref{fig:app_management_algo}. 
For the concept variables in Table~\ref{tab:concept_distribution}, missing entries were imputed with the negative findings. 
A comprehensive dataset summary with detailed variable explanations is available at \url{http://bit.ly/3SoA5E5}.

\begin{figure}[H]%
\centering

\subfigure[]{
    \includegraphics[width=\linewidth]{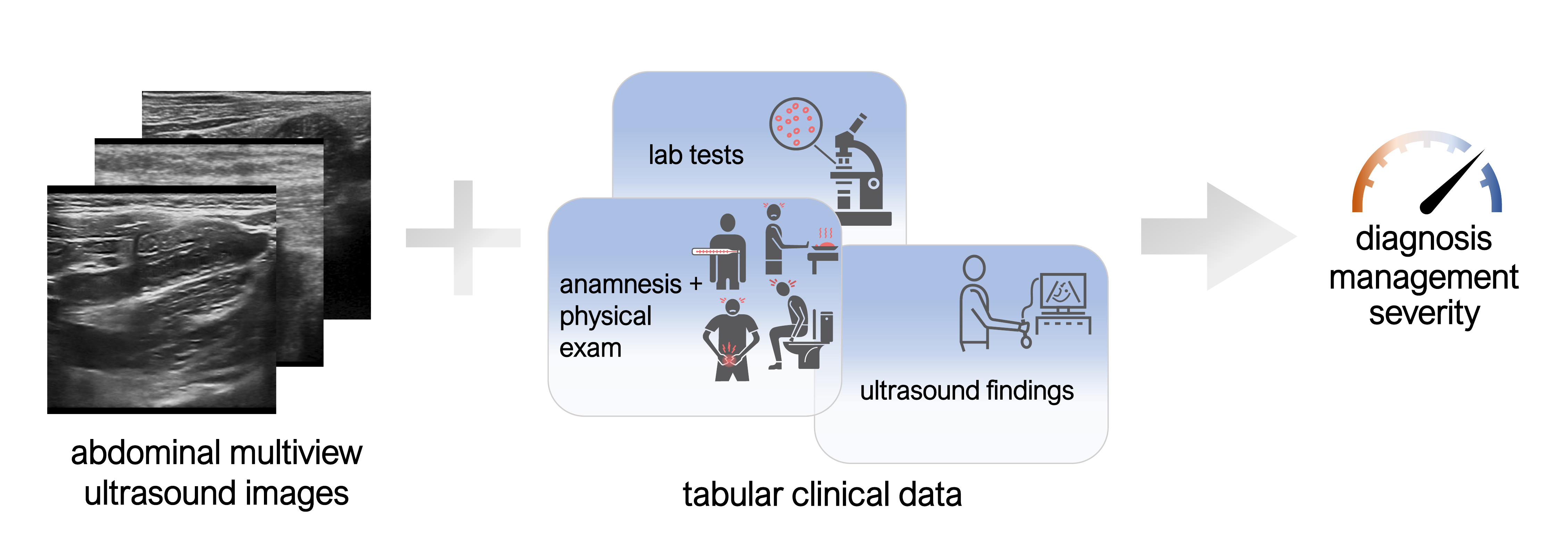}
}

\subfigure[]{
    \includegraphics[width=0.45\linewidth]{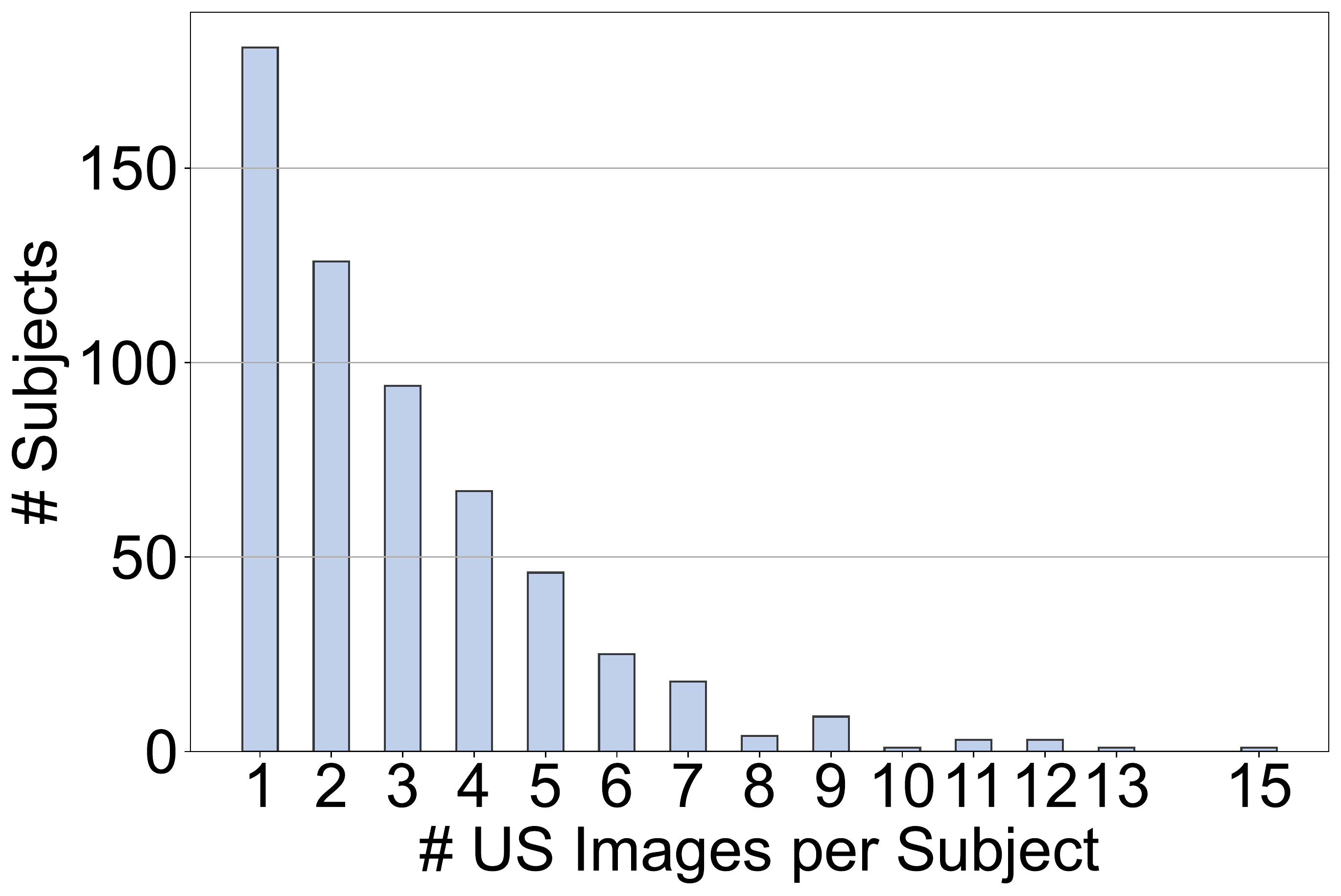}
}
\subfigure[]{
    \includegraphics[width=0.45\linewidth]{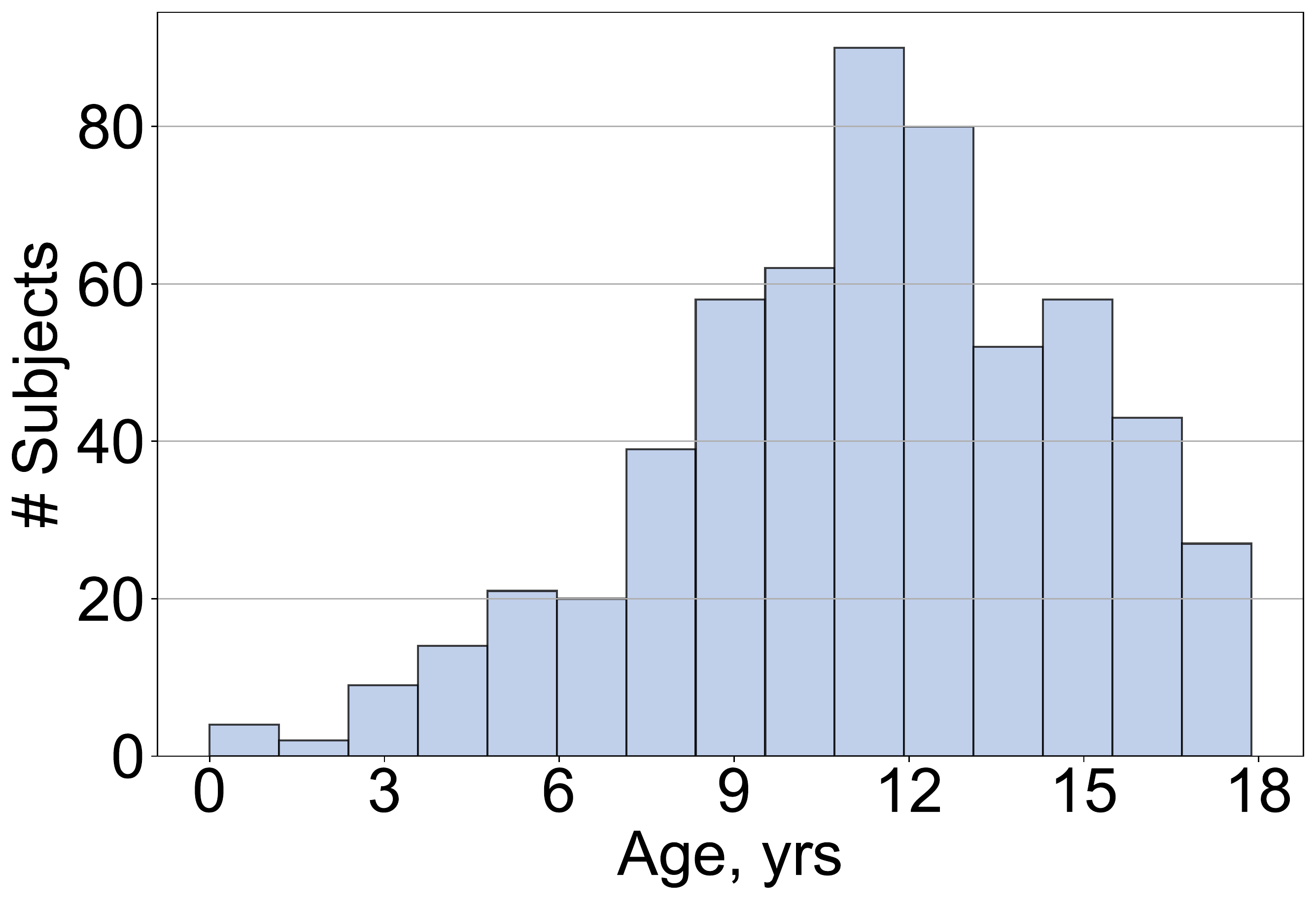}
}

\caption{Overview of the pediatric appendicitis dataset. \mbox{(a) The} dataset structure: for each pediatric patient from the cohort, we acquired multiple ultrasound images, aka views, tabular data comprising laboratory, physical examination, scoring results, and ultrasonographic findings extracted manually by the experts, and three target variables, namely, diagnosis, management and severity. (b--c) Distributions of the number of US images acquired per patient and subjects' age.}
\label{fig:data_overview}
\end{figure}

\section{Synthetic Tabular Nonlinear Dataset\label{app:synthetic}}

To compare different variants of the proposed model and baselines in a controlled manner, we designed a simple synthetic multiview dataset with nonlinear relationships between the covariates, concepts and labels. Let $N$, $p$, $V$, and $K$ denote the number of data points, covariates per view, views, and concepts respectively. Below we outline the generative process behind the data:
\begin{enumerate}
    \item Let $\boldsymbol{\mu}\in\mathbb{R}^{pV}$ be a randomly drawn vector where each component $\mu_j\sim\mathrm{Uniform}\left(-5,\,5\right)$ for $1\leq j\leq pV$.
    \item Let $\boldsymbol{\Sigma}\in\mathbb{R}^{pV\times pV}$ be a randomly generated symmetric, positive-definite matrix.
    \item Let $\boldsymbol{X}\in\mathbb{R}^{N\times pV}$ be a randomly generated feature matrix where $\boldsymbol{X}_{i,:}\sim\mathcal{N}_{pV}\left(\boldsymbol{\mu},\,\boldsymbol{\Sigma}\right)$.
    \item Let $\boldsymbol{x}_i^v=\boldsymbol{X}_{i,\left(1+p(v-1)\right):pv}$ for $1\leq i\leq N$ and $1\leq v\leq V$.
    \item Let $\boldsymbol{g}:\:\mathbb{R}^{pV}\rightarrow\mathbb{R}^K$ and $f:\:\mathbb{R}^K\rightarrow\mathbb{R}$ be randomly initialized MLPs with ReLU nonlinearities.
    \item Let $c_{i,k}=\boldsymbol{1}_{\left\{\left[\boldsymbol{g}\left(\boldsymbol{X}_{i,:}\right)\right]_k\geq m_k\right\}}$, where $m_k=\mathrm{median}\left(\left\{\left[\boldsymbol{g}\left(\boldsymbol{X}_{l,:}\right)\right]_k\right\}_{l=1}^N\right)$, for $1\leq i\leq N$ and $1\leq k\leq K$.
    \item Let $y_i=\boldsymbol{1}_{\left\{f\left(\boldsymbol{c}_i\right)\geq m_y\right\}}$, where $m_y=\textrm{median}\left(\left\{f\left(\boldsymbol{c}_i\right\}_{l=1}^N\right)\right)$, for $1\leq i\leq N$.
\end{enumerate}
Observe that the procedure above results in $N$ triples $\left(\left\{\boldsymbol{x}_i^{v}\right\}_{v=1}^{V},\, \boldsymbol{c}_i,\,y_i\right)$, for $1\leq i\leq N$. By contrast with the appendicitis ultrasonography dataset (Appendix~\ref{app:appendicitis_data}), herein, all data points have the same number of views. In our experiments, we set $N=\mbox{8000}$, $p=500$, $V=3$, and $K=30$. \mbox{2000} data points were held out as a test set. The simulation was repeated across ten independent replications.

\section{Multiview Animals with Attributes\label{app:mvawa}}

\begin{figure}[h!]
    \centering
    \begin{center}
        \includegraphics[width=0.75\columnwidth]{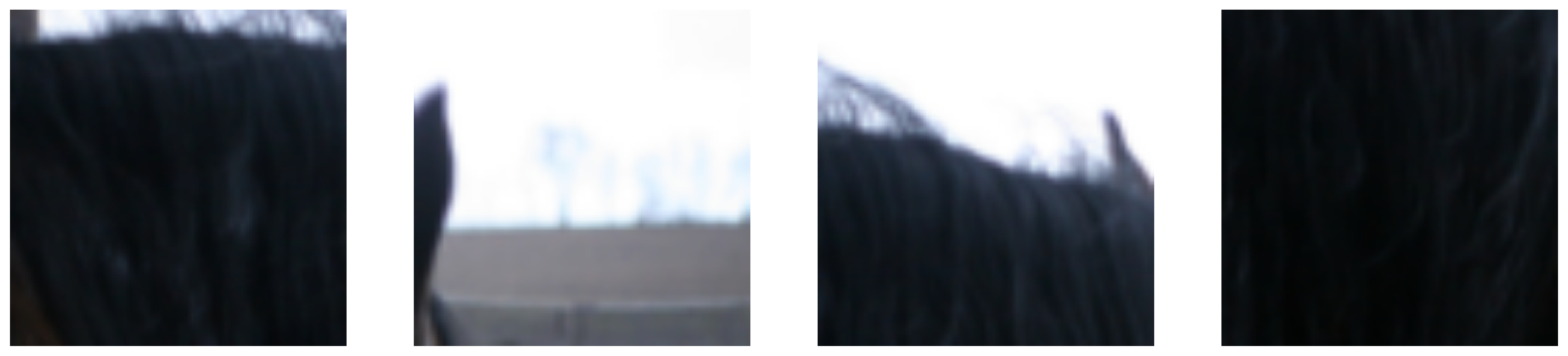}
        
        \includegraphics[width=0.75\columnwidth]{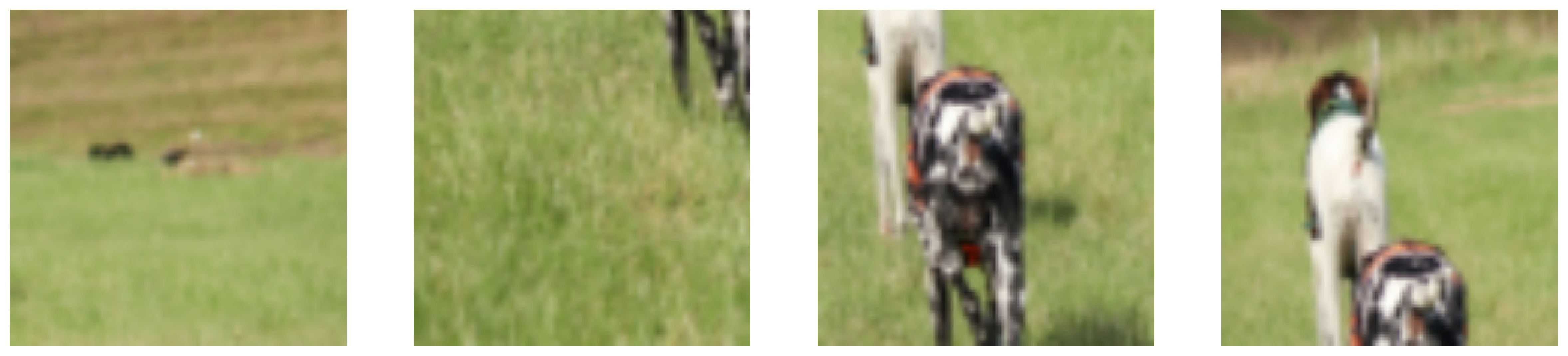}
        
        \includegraphics[width=0.75\columnwidth]{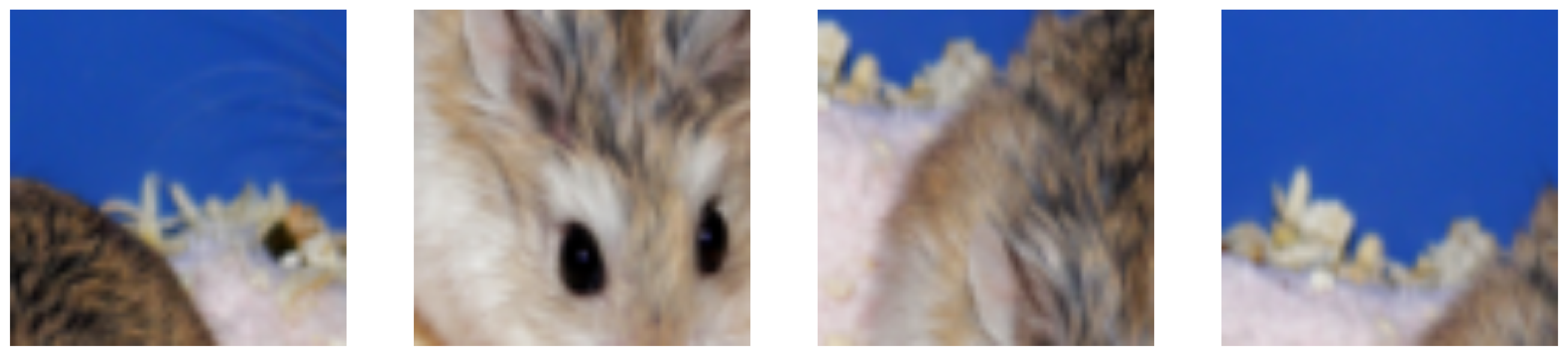}
    \end{center}
    \caption{Three examples of the four-view data points from the multiview AwA dataset. Each row corresponds to a single data point. Every view (column) constitutes a random patch of the original AwA image. Therefore, in this dataset, the views are exchangeable. Moreover, some concepts can be identified only from certain views, e.g. in the bottom row, attributes referring to the background cannot be detected from the second (counting from the left) view.\label{fig:mvawa_ex}}
\end{figure}

In addition to the purely synthetic classification task described above, we adapted a popular attribute-based classification dataset \emph{Animals with Attributes 2} (AwA) \citep{Lampert2009,Xian2019} to the multiview scenario. 
The original AwA consists of 37322 images of 50 animal classes with $K=85$ binary-valued concepts, i.e. attributes. 
Similar to the UCSD Birds experiment for vanilla CBMs \citep{Koh2020}, the concepts are labeled per class and \emph{not} per data point, e.g. all polar bears are assumed to have white fur. 
We extended AwA by randomly cropping $V=4$ patches, 60$\times$60 px\textsuperscript{2} big, from each original image $i$ to produce multiple ``views'', as shown in Figure~\ref{fig:mvawa_ex}. 
Note that, while the concepts are only partially observable from individual images, there is no ordering among the patches, and, for simplicity, we generate the same number of views for each data point. 
Nevertheless, compared with the original AwA, classification based on a single view becomes markedly more challenging. 
During the experiments, we divided the dataset according to the 60\%-20\%-20\% train-validation-test split. 
Simulations were repeated ten times independently.

\newpage

\section{Optimization Procedure for the SSMVCBM\label{app:ssmvcbm}}

The training procedure is summarized in Algorithm~\ref{alg:opt_ssmvcbm}.

\begin{algorithm*}[h!]
\small
\caption{Mini-batch stochastic gradient descent for training the SSMVCBM model. $E_{\boldsymbol{c}}$, $E_{\boldsymbol{z}}$, $E_{a}$, $E_{y}$ denote the numbers of epochs, while $\eta_{\boldsymbol{c}}$, $\eta_{\boldsymbol{z}}$, $\eta_{a}$, $\eta_{y}$ are the respective learning rates for the concept model, representation learning, adversary, and the target model optimization. $C$ is the total number of iterations for the adversarial training, and $\lambda>0$ determines the weight of the adversarial regularizer.}\label{alg:opt_ssmvcbm}
\begin{algorithmic}[1]
\State Initialize $\boldsymbol{\hat{\phi}}^{\boldsymbol{c}}=\left\{\boldsymbol{\hat{\psi}}^{\boldsymbol{c}},\boldsymbol{\hat{\xi}}^{\boldsymbol{c}},\boldsymbol{\hat{\zeta}}^{\boldsymbol{c}}\right\}$
\For{$e \gets 1$ to $E_{\boldsymbol{c}}$} \Comment{{\color{gray}training the concept model}}
    \For{mini-batch $\mathcal{B} \subseteq \{1,...,N\}$}
        \State $\boldsymbol{\hat{\phi}}^{\boldsymbol{c}} \gets \boldsymbol{\hat{\phi}}^{\boldsymbol{c}} - \eta_{\boldsymbol{c}} \nabla_{\boldsymbol{\hat{\phi}}^{\boldsymbol{c}}} \sum\limits_{\substack{\scriptscriptstyle i \in \mathcal{B}\\}}\sum_{k=1}^K w^t_i w_i^{c_k} \mathcal{L}^{c_k} \left(\hat{c}_{i,k}, c_{i,k}\right)$
    \EndFor
\EndFor
\medskip
\State Initialize $\boldsymbol{\hat{\phi}}^{\boldsymbol{z}}=\left\{\boldsymbol{\hat{\psi}}^{\boldsymbol{z}},\boldsymbol{\hat{\xi}}^{\boldsymbol{z}},\boldsymbol{\hat{\zeta}}^{\boldsymbol{z}}\right\}$, $\boldsymbol{\hat{\theta}}$ and $\boldsymbol{\hat{\tau}}$
\For{$j \gets 1$ to $C$} \Comment{{\color{gray}adversarial training}}
    \smallskip
    \For{$e \gets 1$ to $E_{\boldsymbol{z}}$} \Comment{{\color{gray}representation learning}}
        \For{mini-batch $\mathcal{B}\subseteq\{1,...,N\}$}
            \State $
            \left\{\boldsymbol{\hat{\phi}}^{\boldsymbol{z}},\boldsymbol{\hat{\theta}}\right\} \gets \left\{\boldsymbol{\hat{\phi}}^{\boldsymbol{z}},\boldsymbol{\hat{\theta}}\right\} - \eta_{\boldsymbol{z}} \nabla_{\left\{\boldsymbol{\hat{\phi}}^{\boldsymbol{z}},\boldsymbol{\hat{\theta}}\right\}} \Bigg[ \sum\limits_{i \in \mathcal{B}} w^t_i \mathcal{L}^t\left(\hat{y}_i, y_i\right)
            -\lambda \sum\limits_{\substack{\scriptscriptstyle i \in \mathcal{B}}}\sum_{i=1}^K \mathcal{L}^{c_k} \left(\left[a_{\boldsymbol{\hat{\tau}}}\left(\boldsymbol{\hat{z}}_i\right)\right]_k, \hat{c}_{i,k}\right) \Bigg]
            $
        \EndFor
    \EndFor
    \smallskip
    \For{$e \gets 1$ to $E_{a}$} \Comment{{\color{gray}training the adversary}} 
        \For{mini-batch $\mathcal{B} \subseteq \{1, \ldots, N\}$}
            \State $\boldsymbol{\hat{\tau}} \gets \boldsymbol{\hat{\tau}} - \eta_{a} \nabla_{\boldsymbol{\hat{\tau}}} \sum\limits_{\substack{\scriptscriptstyle i \in \mathcal{B}}}\sum_{k=1}^K w_i^{c_k} \mathcal{L}^{c_k} \left(\left[a_{\boldsymbol{\hat{\tau}}}\left(\boldsymbol{\hat{z}}_i\right)\right]_k, \hat{c}_{i,k}\right)$
        \EndFor
    \EndFor
\EndFor
\medskip
\State Re-initialize $\boldsymbol{\hat{\theta}}$
\For{$e \gets 1$ to $E_{y}$} \Comment{{\color{gray}training the target model}} 
    \For{batch $\mathcal{B} \subseteq \{1,...,N\}$}
        \State $\boldsymbol{\hat{\theta}} \gets \boldsymbol{\hat{\theta}} - \eta_{y} \nabla_{\boldsymbol{\hat{\theta}}} \sum\limits_{i \in \mathcal{B}} w^t_i \mathcal{L}^t \left(\hat{y}_i, y_i\right)$
    \EndFor
\EndFor
\medskip
\State\Return $\left\{\boldsymbol{\hat{\phi}}^{\boldsymbol{c}},\boldsymbol{\hat{\phi}}^{\boldsymbol{z}},\boldsymbol{\hat{\theta}}\right\}$
\end{algorithmic}
\end{algorithm*}

\newpage

\section{Further Implementation Details\label{app:imp}}
\subsection{Architectures}
Table~\ref{tab:architectures} provides a detailed description of the MVCBM architectures implemented in our experiments. 
Herein, $B$ denotes the batch size, $V$ is the maximum number of views, $K$ is the number of concepts, $H$ is the number of units in the hidden layer of $f_{\boldsymbol{\theta}}(\cdot)$, and $N_o$ is the number of output units. 
Table~\ref{tab:architectures}(a) shows the architectures for the tabular synthetic data, and Table~\ref{tab:architectures}(b) shows the architectures utilized for the image datasets. 
As can be seen, the encoder network $\boldsymbol{h}_{\boldsymbol{\psi}}(\cdot)$ is different: in  the first case, it is fully connected, whereas, in the latter, it is comprised of the ResNet-18 without the penultimate fully connected layer. Notably, the number of output units $N_o$ and the activation function depend on the number of classes of the target variable.

\begin{table}[h!]
    \scriptsize
    \centering
    \caption{Summary of the MVCBM architectures used for the \mbox{(a) synthetic} and \mbox{(b) MVAwA} and pediatric appendicitis datasets. Here, $B$ denotes the batch size, $V$ the maximum number of views, $w$ and $h$ the width and height of the input image, $K$ the number of concepts, $H$ the number of units in the hidden layer of $f_{\boldsymbol{\theta}}(\cdot)$, and $N_o$ the number of output units. \label{tab:architectures}}

    \subfigure[]{
        \centering
        \begin{tabular}{clll}
            \textbf{Module} & \textbf{Layers} & \textbf{Input Dimensions} & \textbf{Output Dimensions}\\
            \toprule
            \multirow{10}{.5cm}{$\boldsymbol{h}_{\boldsymbol{\psi}}(\cdot)$} & \texttt{Linear} & ($B$, $V$, 500) & ($B$, $V$, 256) \\
            & \texttt{Dropout(0.05)} & ($B$, $V$, 256) & ($B$, $V$, 256) \\
            & \texttt{BatchNorm1d} & ($B$, $V$, 256) & ($B$, $V$, 256) \\
            & \texttt{Linear} & ($B$, $V$, 256) & ($B$, $V$, 256) \\
            & \texttt{Dropout(0.05)} & ($B$, $V$, 256) & ($B$, $V$, 256) \\
            & \texttt{BatchNorm1d} & ($B$, $V$, 256) & ($B$, $V$, 256) \\
            & \texttt{Linear} & ($B$, $V$, 256) & ($B$, $V$, 256) \\
            & \texttt{Dropout(0.05)} & ($B$, $V$, 256) & ($B$, $V$, 256) \\
            & \texttt{BatchNorm1d} & ($B$, $V$, 256) & ($B$, $V$, 256) \\
            & \texttt{Linear} & ($B$, $V$, 256) & ($B$, $V$, 128) \\
            \midrule
            $\boldsymbol{r}_{\boldsymbol{\xi}}(\cdot)$ & \texttt{LSTM}/\texttt{mean} & ($B$, $V$, 128) & ($B$, 128)\\
            \midrule
            \multirow{6}{.5cm}{$\boldsymbol{s}_{\boldsymbol{\zeta}}(\cdot)$} & \texttt{Linear} & ($B$, 128) & ($B$, 256)\\
            & \texttt{ReLu} & ($B$, 256) & ($B$, 256)\\
            & \texttt{Linear} & ($B$, 256) & ($B$, 64)\\
            & \texttt{ReLu} & ($B$, 64) & ($B$, 64)\\
            & \texttt{Linear} & ($B$, 64) & ($B$, $K$)\\
            & \texttt{Sigmoid} & ($B$, $K$) & ($B$, $K$)\\
            \midrule
            \multirow{4}{.5cm}{$f_{\boldsymbol{\theta}}(\cdot)$}
            & \texttt{Linear} & ($B$, $K$) & ($B$, $H$)\\
            & \texttt{ReLu} & ($B$, $H$) & ($B$, $H$)\\
            & \texttt{Linear} & ($B$, $H$) & ($B$, 1)\\
            & \texttt{Sigmoid} & ($B$, 1) & ($B$, 1)\\
            \bottomrule
        \end{tabular}
    }
    
    \subfigure[]{
        \centering
        \begin{tabular}{clll}
            \textbf{Module} & \textbf{Layers} & \textbf{Input Dimensions} & \textbf{Output Dimensions}\\
            \toprule
            $\boldsymbol{h}_{\boldsymbol{\psi}}(\cdot)$ & \texttt{ResNet-18} & ($B$, $V$, 3, $w$, $h$) & ($B$, $V$, 512) \\
            \midrule
            $\boldsymbol{r}_{\boldsymbol{\xi}}(\cdot)$ & \texttt{LSTM}/\texttt{mean} & ($B$, $V$, 512) & ($B$, 512)\\
            \midrule
            \multirow{6}{.5cm}{$\boldsymbol{s}_{\boldsymbol{\zeta}}(\cdot)$} & \texttt{Linear} & ($B$, 512) & ($B$, 256)\\
            & \texttt{ReLu} & ($B$, 256) & ($B$, 256)\\
            & \texttt{Linear} & ($B$, 256) & ($B$, 64)\\
            & \texttt{ReLu} & ($B$, 64) & ($B$, 64)\\
            & \texttt{Linear} & ($B$, 64) & ($B$, $K$)\\
            & \texttt{Sigmoid} & ($B$, $K$) & ($B$, $K$)\\
            \midrule
            \multirow{4}{.5cm}{$f_{\boldsymbol{\theta}}(\cdot)$}
            & \texttt{Linear} & ($B$, $K$) & ($B$, $H$)\\
            & \texttt{ReLu} & ($B$, $H$) & ($B$, $H$)\\
            & \texttt{Linear} & ($B$, $H$) & ($B$, $N_o$)\\
            & \texttt{Sigmoid}/\texttt{Softmax} & ($B$, $N_o$) & ($B$, $N_o$)\\
            \bottomrule
        \end{tabular}
    }
\end{table}

Note that, in the appendicitis dataset, all US image sequences were padded to the length of $V=20$. 
However, as intended, fusion layers discard the padding and can be applied to variable-length sequences. 
As mentioned in Appendices~\ref{app:synthetic} and \ref{app:mvawa}, we considered $V=3$ and $4$ views for the synthetic and MVAwA datasets, respectively. 
The number of concepts was $K=30$, $85$, and $9$ for the synthetic, MVAwA, and appendicitis datasets, respectively. 
Notably, in MVAwA, the input images were 224$\times$224 px, while US images were 400$\times$400 px big. 
For the synthetic dataset and MVAwA, we fixed $H=100$, and for the appendicitis data, it was set to $5$. 
For MVAwA, the output layer was $N_o=50$ units wide and had \texttt{Softmax} activation. 
Since all labels in the pediatric appendicitis dataset were binary, we set $N_o=1$ and used \texttt{Sigmoid} activation.

For the SSMVCBM, we had to choose architectures for the concept prediction and representation learning ``branches'' of the model, given by Eq.~(\ref{eqn:ssmvcbm}). 
For both, we utilized architectures similar to those from Table~\ref{tab:architectures}. 
For representation learning, instead of the sigmoid, we applied the hyperbolic tangent activation function at the output of $\boldsymbol{s}_{\boldsymbol{\zeta}}^{\boldsymbol{z}}(\cdot)$. 
Another architectural hyperparameter of the SSMVCBM is the number of dimensions of the vector $\boldsymbol{\hat{z}}_i$, denoted by $J$. 
In the experiments reported in Figure~\ref{fig:synthetic_res}, $J$ was set to the difference between the number of the ground-truth concepts and the number of the concepts given to the model during training. 
For the experiments from Table~\ref{tab:mvawa_full_res}, we set $J=24$. 
Finally, for pediatric appendicitis, we fixed $J=5$ across all target variables. 
Lastly, another architectural difference from the MVCBM was that the number of inputs in the target model $f_{\boldsymbol{\theta}}(\cdot)$ had to be $K+J$. 
For more detailed architecture specifications not covered above, see our code at \url{https://github.com/i6092467/semi-supervised-multiview-cbm}.

\subsection{Hyperparameters}

In all experiments, deep learning models were trained using the Adam optimizer. 
To avoid potential overfitting on the moderately-sized appendicitis dataset, throughout training, we applied on-the-fly data augmentation with Gaussian noise addition, random black rectangle insertion, and one additional randomly chosen transformation: brightness adjustment, rotation, shearing, resizing, change of image sharpness, or gamma correction. 

Applicable model hyperparameter values used for the synthetic, MVAwA, and appendicitis datasets are provided in Tables~\ref{tab:parameters_synthetic}--\ref{tab:parameters_app_severity}. 
The numbers of training epochs and learning rates were selectively tuned on the training set using five-fold cross-validation. 
In the tables below, by $E_y$ and $\eta_y$, we denote the number of epochs used to train a model and the initial learning rate, respectively. 
Note that sequentially trained MVCBMs allow for a separate hyperparameter configuration for the concept model $\boldsymbol{g}_{\boldsymbol{\phi}}(\cdot)$. 
We exploit this possibility for the number of epochs ($E_{\boldsymbol{c}}$) and the initial learning rate ($\eta_{\boldsymbol{c}}$). 
Due to the lack of this freedom, we have found that jointly trained MVCBMs sometimes require tuning $E_y$ and $\eta_y$ for the model weights to converge. 
Recall that parameter $\alpha$ controls the trade-off between the target and concept loss terms in the jointly trained concept bottleneck models. 
We did not explore the influence of this hyperparameter, fixing it to $\alpha=1.0$. 
The remaining parameters belong to the semi-supervised variant of the MVCBM (see the procedure in Algorithm~\ref{alg:opt_ssmvcbm}): $C$ denotes the number of iterations for the adversarial training; $E_{\boldsymbol{z}}$ and $\eta_{\boldsymbol{z}}$ are the number of training epochs and learning rate, respectively, for the representation learning module; $E_a$ and $\eta_a$ are the number of epochs and learning rate for training the adversary network; and, finally, $\lambda$ is the parameter controlling the weight of the adversarial penalty in the loss function for optimizing the representation learning module and target model parameters.

\begin{table}[H]
    \scriptsize
    \centering
    \caption{Final hyperparameter values used to train models on the \textbf{synthetic nonlinear data}. The meaning of the hyperparameters: $E_{\boldsymbol{c}}$, the number of training epochs for the concept model; $C$, the number of iterations in the adversarial training procedure for the SSMVCBM; $E_{\boldsymbol{z}}$, the number of training epochs for the representation learning module; $E_{a}$, the number of training epochs for the adversary; $E_y$, the number of training epochs for the target model or the full model; $\eta_{\boldsymbol{c}}$, the learning rate (LR) for the concept model; $\eta_{\boldsymbol{z}}$, the LR for the representation learning module; $\eta_a$, the LR for the adversary; $\eta_y$, the LR for the target or the full model; $B$, the mini-batch size; $\alpha$, a parameter controlling the trade-off between target and concept prediction in the joint optimization; $\lambda$, the weight of the adversarial regularizer in the loss function of the SSMVCBM.}
    \label{tab:parameters_synthetic} 
    \centering
    \begin{tabular}{l c c c c c c c c c c c c} 
        \toprule
        \multirow{2}{*}{\raisebox{-\heavyrulewidth}{\textbf{Model}}} & \multicolumn{11}{c}{\textbf{Hyperparameter}}\\
        \cmidrule{2-13} & $E_{\boldsymbol{c}}$ & $C$ & $E_{\boldsymbol{z}}$ & $E_a$ & $E_y$ & $\eta_{\boldsymbol{c}}$ & $\eta_{\boldsymbol{z}}$ & $\eta_a$ & $\eta_y$ & $B$ & $\alpha$ & $\lambda$ \\
        \midrule
        MLP & --- & --- & --- & --- & 150 & --- & --- & --- & 1.0e-3 & 64 & --- & --- \\
        \midrule
        CBM-seq & 100 & --- & --- & --- & 50 & 1.0e-3 & --- & --- & 1.0e-3 & 64 & --- & --- \\
        CBM-joint & --- & --- & --- & --- & 120 & --- & ---- & --- & 1.0e-4 & 64 & 1.0 & --- \\
        \midrule
        MVBM-avg & --- & --- & --- & --- & 150 & --- & --- & --- & 1.0e-3 & 64 & --- & --- \\
        MVBM-LSTM & --- & --- & --- & --- & 150 & --- & ---- & --- & 1.0e-3 & 64 & --- & --- \\
        \midrule
        MVCBM-seq-avg & 100 & --- & --- & --- & 50 & 1.0e-3 & --- & --- & 1.0e-3 & 64 & --- & ---\\
        MVCBM-seq-LSTM & 100 & --- & --- & --- & 50 & 1.0e-3 & --- & --- & 1.0e-3 & 64 & --- & --- \\
        MVCBM-joint-avg & --- & --- & --- & --- & 120 & --- & --- & --- & 1.0e-4 & 64 & 1.0 & --- \\
        MVCBM-joint-LSTM & --- & --- & --- & --- & 120 & --- & --- & --- & 1.0e-4 & 64 & 1.0 & --- \\
        \midrule
        SSMVCBM-avg & 100 & 7 & 30 & 30 & 50 & 1.0e-3 & 1.0e-3 & 1.0e-3  & 1.0e-3 & 64 & --- & 1.0e-2 \\
        SSMVCBM-LSTM & 100 & 7 & 30 & 30 & 50 & 1.0e-3 & 1.0e-3 & 1.0e-3  & 1.0e-3 & 64 & --- & 1.0e-2 \\
        \bottomrule
    \end{tabular}
\end{table}

\begin{table}[H]
    \scriptsize
    \centering
    \caption{Final hyperparameter values used to train models on the \textbf{multiview animals with attributes}.}
    \label{tab:parameters_mvawa}
    \begin{tabular}{l c c c c c c c c c c c c} 
        \toprule
        \multirow{2}{*}{\raisebox{-\heavyrulewidth}{\textbf{Model}}} & \multicolumn{11}{c}{\textbf{Hyperparameter}}\\
        \cmidrule{2-13} & $E_{\boldsymbol{c}}$ & $C$ & $E_{\boldsymbol{z}}$ & $E_a$ & $E_y$ & $\eta_{\boldsymbol{c}}$ & $\eta_{\boldsymbol{z}}$ & $\eta_a$ & $\eta_y$ & $B$ & $\alpha$ & $\lambda$ \\
        \midrule
        ResNet-18 & --- & --- & --- & --- & 120 & --- & --- & --- & 1.0e-4 & 64 & --- & --- \\
        \midrule
        CBM-seq & 25 & --- & --- & --- & 20 & 1.0e-4 & --- & --- & 1.0e-2 & 64 & --- \\
        CBM-joint & --- & --- & --- & --- & 120 & --- & --- & --- & 1.0e-4 & 64 & 1.0 & --- \\
        \midrule
        MVBM-avg & --- & --- & --- & --- & 120 & --- & --- & --- & 1.0e-4 & 64 & --- & --- \\
        MVBM-LSTM & --- & --- & --- & --- & 120 & --- & --- & --- & 1.0e-4 & 64 & --- & --- \\
        \midrule
        MVCBM-seq-avg & 25 & --- & --- & --- & 20 & 1.0e-4 & --- & --- & 1.0e-2 & 64 & --- & --- \\
        MVCBM-seq-LSTM & 25 & --- & --- & --- & 20 & 1.0e-4 & --- & --- & 1.0e-2 & 64 & --- & --- \\
        MVCBM-joint-avg & --- & --- & --- & --- & 120 & --- & --- & --- & 1.0e-4 & 64 & 1.0 & --- \\
        MVCBM-joint-LSTM & --- & --- & --- & --- & 120 & --- & --- & --- & 1.0e-4 & 64 & 1.0 & --- \\
        \midrule
        SSMVCBM-avg & 25 & 7 & 15 & 10 & 20 & 1.0e-4 & 1.0e-4 & 1.0e-2 & 1.0e-2 & 64 & --- & 1.0e-2 \\
        SSMVCBM-LSTM & 25 & 7 & 15 & 10 & 20 & 1.0e-4 & 1.0e-4 & 1.0e-2 & 1.0e-2 & 64 & --- & 1.0e-2 \\
        \bottomrule
    \end{tabular}
    \footnotetext{$E_{\boldsymbol{c}}$: the number of training epochs for the concept model; $C$: the number of iterations in the adversarial training procedure for the SSMVCBM; $E_{\boldsymbol{z}}$: the number of training epochs for the representation learning module; $E_{a}$: the number of training epochs for the adversary; $E_y$: the number of training epochs for the target model or the full model; $\eta_{\boldsymbol{c}}$: the learning rate (LR) for the concept model; $\eta_{\boldsymbol{z}}$: the LR for the representation learning module; $\eta_a$: the LR for the adversary; $\eta_y$: the LR for the target or the full model; $B$: the mini-batch size; $\alpha$: a parameter controlling the trade-off between target and concept prediction in the joint optimization; $\lambda$: the weight of the adversarial regularizer in the loss function of the SSMVCBM.}
\end{table}

\begin{table}[H]
    \scriptsize
    \centering
    \caption{Final hyperparameter values used to train models on the \textbf{appendicitis} data with the \textbf{diagnosis} as the target.}
    \label{tab:parameters_app_diagnosis}
    \begin{tabular}{l c c c c c c c c c c c c} 
        \toprule
        \multirow{2}{*}{\raisebox{-\heavyrulewidth}{\textbf{Model}}} & \multicolumn{11}{c}{\textbf{Hyperparameter}}\\
        \cmidrule{2-13} & $E_{\boldsymbol{c}}$ & $C$ & $E_{\boldsymbol{z}}$ & $E_a$ & $E_y$ & $\eta_{\boldsymbol{c}}$ & $\eta_{\boldsymbol{z}}$ & $\eta_a$ & $\eta_y$ & $B$ & $\alpha$ & $\lambda$ \\
        \midrule
        ResNet-18 & --- & --- & --- & --- & 120 & --- & --- & --- & 1.0e-4 & 4 & --- & --- \\
        \midrule
        CBM-seq & 25 & --- & --- & --- & 20 & 1.0e-4 & --- & --- & 1.0e-2 & 4 & --- & --- \\
        CBM-joint & --- & --- & --- & --- & 120 & --- & --- & --- & 1.0e-4 & 4 & 1.0 & --- \\
        \midrule
        MVBM-avg & --- & --- & --- & --- & 120 & --- & --- & --- & 1.0e-4 & 4 & --- & --- \\
        MVBM-LSTM & --- & --- & --- & --- & 50 & --- & --- & --- & 1.0e-4 & 4 & --- & --- \\
        \midrule
        MVCBM-seq-avg & 20 & --- & --- & --- & 20 & 1.0e-4 & --- & --- & 1.0e-2 & 4 & --- & --- \\
        MVCBM-seq-LSTM & 20 & --- & --- & --- & 20 & 1.0e-4 & --- & --- & 1.0e-2 & 4 & --- & --- \\
        MVCBM-joint-avg & --- & --- & --- & --- & 120 & --- & --- & --- & 1.0e-4 & 4 & 1.0 & --- \\
        MVCBM-joint-LSTM & --- & --- & --- & --- & 40 & --- & --- & --- & 1.0e-3 & 4 & 1.0 & --- \\
        \midrule
        SSMVCBM-avg & 20 & 7 & 15 & 10 & 20 & 1.0e-4 & 1.0e-4 & 1.0e-2 & 1.0e-2 & 8 & --- & 1.0e-2 \\
        SSMVCBM-LSTM & 20 & 7 & 15 & 10 & 20 & 1.0e-4 & 1.0e-4 & 1.0e-2 & 1.0e-2 & 8 & --- & 1.0e-2 \\
        \bottomrule
    \end{tabular}
    \footnotetext{$E_{\boldsymbol{c}}$: the number of training epochs for the concept model; $C$: the number of iterations in the adversarial training procedure for the SSMVCBM; $E_{\boldsymbol{z}}$: the number of training epochs for the representation learning module; $E_{a}$: the number of training epochs for the adversary; $E_y$: the number of training epochs for the target model or the full model; $\eta_{\boldsymbol{c}}$: the learning rate (LR) for the concept model; $\eta_{\boldsymbol{z}}$: the LR for the representation learning module; $\eta_a$: the LR for the adversary; $\eta_y$: the LR for the target or the full model; $B$: the mini-batch size; $\alpha$: a parameter controlling the trade-off between target and concept prediction in the joint optimization; $\lambda$: the weight of the adversarial regularizer in the loss function of the SSMVCBM.}
\end{table}

\begin{table}[H]
    \scriptsize
    \centering
    \caption{Final hyperparameter values used to train models on the \textbf{appendicitis} data with the \textbf{management} as the target.}
    \label{tab:parameters_app_management}
    \begin{tabular}{l c c c c c c c c c c c c} 
        \toprule
        \multirow{2}{*}{\raisebox{-\heavyrulewidth}{\textbf{Model}}} & \multicolumn{11}{c}{\textbf{Hyperparameter}}\\
        \cmidrule{2-13} & $E_{\boldsymbol{c}}$ & $C$ & $E_{\boldsymbol{z}}$ & $E_a$ & $E_y$ & $\eta_{\boldsymbol{c}}$ & $\eta_{\boldsymbol{z}}$ & $\eta_a$ & $\eta_y$ & $B$ & $\alpha$ & $\lambda$ \\
        \midrule
        ResNet-18 & --- & --- & --- & --- & 120 & --- & --- & --- & 1.0e-4 & 4 & --- & --- \\
        \midrule
        CBM-seq & 25 & --- & --- & --- & 20 & 1.0e-4 & --- & --- & 1.0e-2 & 4 & --- & --- \\
        CBM-joint & --- & --- & --- & --- & 120 & --- & --- & --- & 1.0e-4 & 4 & 1.0 & --- \\
        \midrule
        MVBM-avg & --- & --- & --- & --- & 120 & --- & --- & --- & 1.0e-4 & 4 & --- & --- \\
        MVBM-LSTM & --- & --- & --- & --- & 50 & --- & --- & --- & 1.0e-4 & 4 & --- & --- \\
        \midrule
        MVCBM-seq-avg & 20 & --- & --- & --- & 20 & 1.0e-4 & --- & --- & 1.0e-2 & 4 & --- & --- \\
        MVCBM-seq-LSTM & 20 & --- & --- & --- & 20 & 1.0e-4 & --- & --- & 1.0e-2 & 4 & --- & --- \\
        MVCBM-joint-avg & --- & --- & --- & --- & 120 & --- & --- & --- & 1.0e-4 & 4 & 1.0 & --- \\
        MVCBM-joint-LSTM & --- & --- & --- & --- & 120 & --- & --- & --- & 1.0e-4 & 4 & 1.0 & --- \\
        \midrule
        SSMVCBM-avg & 20 & 7 & 15 & 10 & 20 & 1.0e-4 & 1.0e-4 & 1.0e-2 & 1.0e-2 & 8 & --- & 1.0e-2 \\
        SSMVCBM-LSTM & 20 & 7 & 15 & 10 & 20 & 1.0e-4 & 1.0e-4 & 1.0e-2 & 1.0e-2 & 8 & --- & 1.0e-2 \\
        \bottomrule
    \end{tabular}
    \footnotetext{$E_{\boldsymbol{c}}$: the number of training epochs for the concept model; $C$: the number of iterations in the adversarial training procedure for the SSMVCBM; $E_{\boldsymbol{z}}$: the number of training epochs for the representation learning module; $E_{a}$: the number of training epochs for the adversary; $E_y$: the number of training epochs for the target model or the full model; $\eta_{\boldsymbol{c}}$: the learning rate (LR) for the concept model; $\eta_{\boldsymbol{z}}$: the LR for the representation learning module; $\eta_a$: the LR for the adversary; $\eta_y$: the LR for the target or the full model; $B$: the mini-batch size; $\alpha$: a parameter controlling the trade-off between target and concept prediction in the joint optimization; $\lambda$: the weight of the adversarial regularizer in the loss function of the SSMVCBM.}
\end{table}

\begin{table}[H]
    \footnotesize
    \centering
    \caption{Final hyperparameter values used to train models on the \textbf{appendicitis} data with the \textbf{severity} as the target.}
    \label{tab:parameters_app_severity}
    \begin{tabular}{l c c c c c c c c c c c c} 
        \toprule
        \multirow{2}{*}{\raisebox{-\heavyrulewidth}{\textbf{Model}}} & \multicolumn{11}{c}{\textbf{Hyperparameter}}\\
        \cmidrule{2-13} & $E_{\boldsymbol{c}}$ & $C$ & $E_{\boldsymbol{z}}$ & $E_a$ & $E_y$ & $\eta_{\boldsymbol{c}}$ & $\eta_{\boldsymbol{z}}$ & $\eta_a$ & $\eta_y$ & $B$ & $\alpha$ & $\lambda$ \\
        \midrule
        ResNet-18 & --- & --- & --- & --- & 120 & --- & --- & --- & 1.0e-4 & 4 & --- & --- \\
        \midrule
        CBM-seq & 25 & --- & --- & --- & 20 & 1.0e-4 & --- & --- & 1.0e-2 & 4 & --- & --- \\
        CBM-joint & --- & --- & --- & --- & 120 & --- & --- & --- & 1.0e-4 & 4 & 1.0 & --- \\
        \midrule
        MVBM-avg & --- & --- & --- & --- & 120 & --- & --- & --- & 1.0e-4 & 4 & --- & --- \\
        MVBM-LSTM & --- & --- & --- & --- & 70 & --- & --- & --- & 1.0e-4 & 4 & --- & --- \\
        \midrule
        MVCBM-seq-avg & 30 & --- & --- & --- & 40 & 1.0e-4 & --- & --- & 1.0e-3 & 4 & --- & --- \\
        MVCBM-seq-LSTM & 30 & --- & --- & --- & 40 & 1.0e-4 & --- & --- & 1.0e-3 & 4 & --- & --- \\
        MVCBM-joint-avg & --- & --- & --- & --- & 100 & --- & --- & --- & 1.0e-4 & 4 & 1.0 & --- \\
        MVCBM-joint-LSTM & --- & --- & --- & --- & 100 & --- & --- & --- & 1.0e-4 & 4 & 1.0 & --- \\
        \midrule
        SSMVCBM-avg & 20 & 7 & 15 & 10 & 20 & 1.0e-4 & 1.0e-4 & 1.0e-2 & 1.0e-2 & 8 & --- & 1.0e-2 \\
        SSMVCBM-LSTM & 20 & 7 & 15 & 10 & 20 & 1.0e-4 & 1.0e-4 & 1.0e-2 & 1.0e-2 & 8 & --- & 1.0e-2 \\
        \bottomrule
    \end{tabular}
    \footnotetext{$E_{\boldsymbol{c}}$: the number of training epochs for the concept model; $C$: the number of iterations in the adversarial training procedure for the SSMVCBM; $E_{\boldsymbol{z}}$: the number of training epochs for the representation learning module; $E_{a}$: the number of training epochs for the adversary; $E_y$: the number of training epochs for the target model or the full model; $\eta_{\boldsymbol{c}}$: the learning rate (LR) for the concept model; $\eta_{\boldsymbol{z}}$: the LR for the representation learning module; $\eta_a$: the LR for the adversary; $\eta_y$: the LR for the target or the full model; $B$: the mini-batch size; $\alpha$: a parameter controlling the trade-off between target and concept prediction in the joint optimization; $\lambda$: the weight of the adversarial regularizer in the loss function of the SSMVCBM.}
\end{table}

\section{Further Results\label{app:res}}
\subsection{Multiview Animals with Attributes\label{app:mvawa_results}}

As mentioned, we also adapted a popular natural image attribute-based  \emph{Animals with Attributes 2} dataset \citep{Lampert2009,Xian2019} to the multiview classification (Appendix~\ref{app:mvawa}). 
The main challenge of this dataset is that only some concepts may be identifiable from every view because cropping may remove an image region with the input relevant to a specific concept. 
During model comparison, we trained and evaluated classifiers by performing a train-test split on several independent simulations, i.e. replicates. 

\begin{figure}[H]
\centering

\subfigure[]{
    \includegraphics[width=0.3\linewidth]{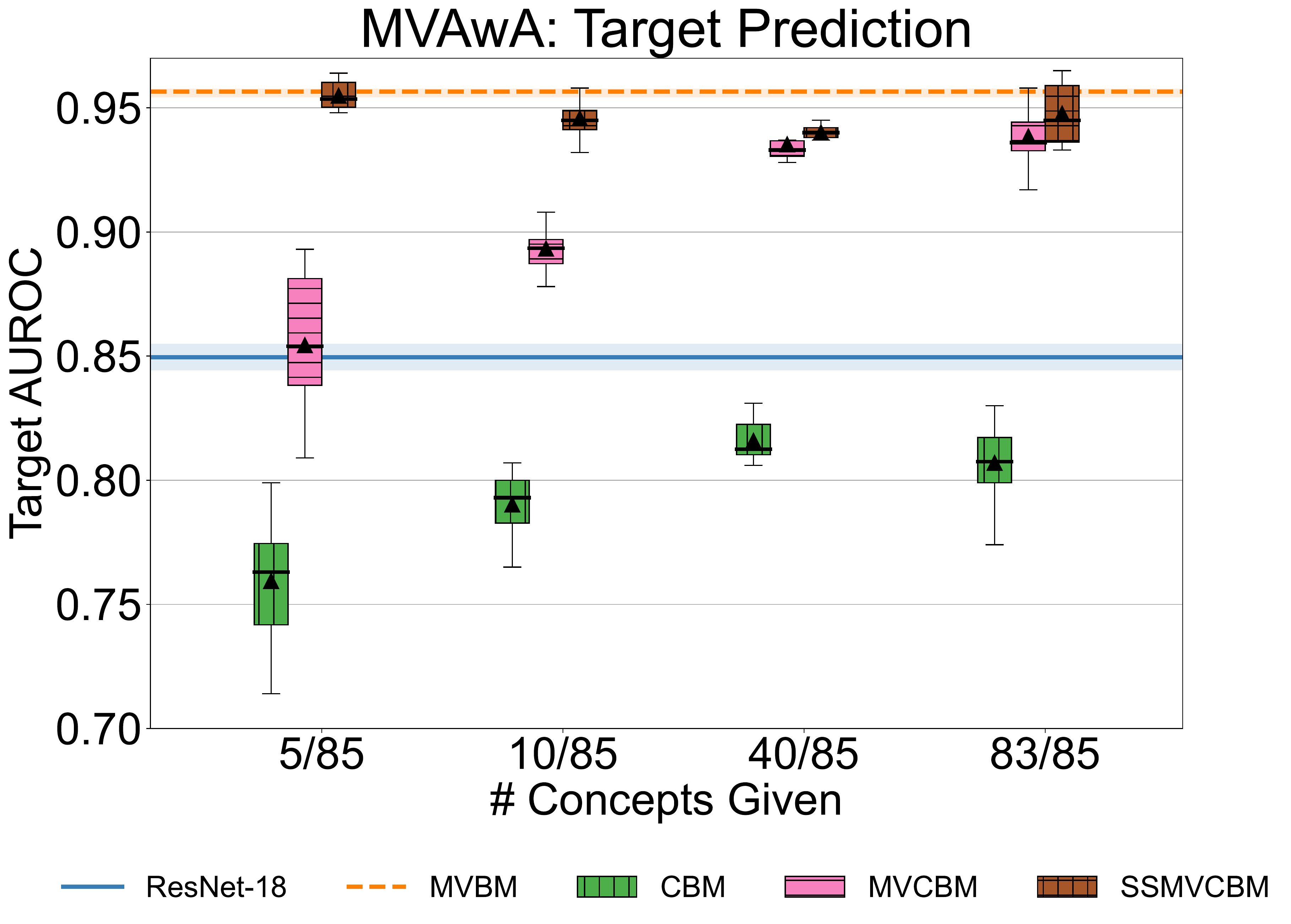}
}
\subfigure[]{
    \includegraphics[width=0.29\linewidth]{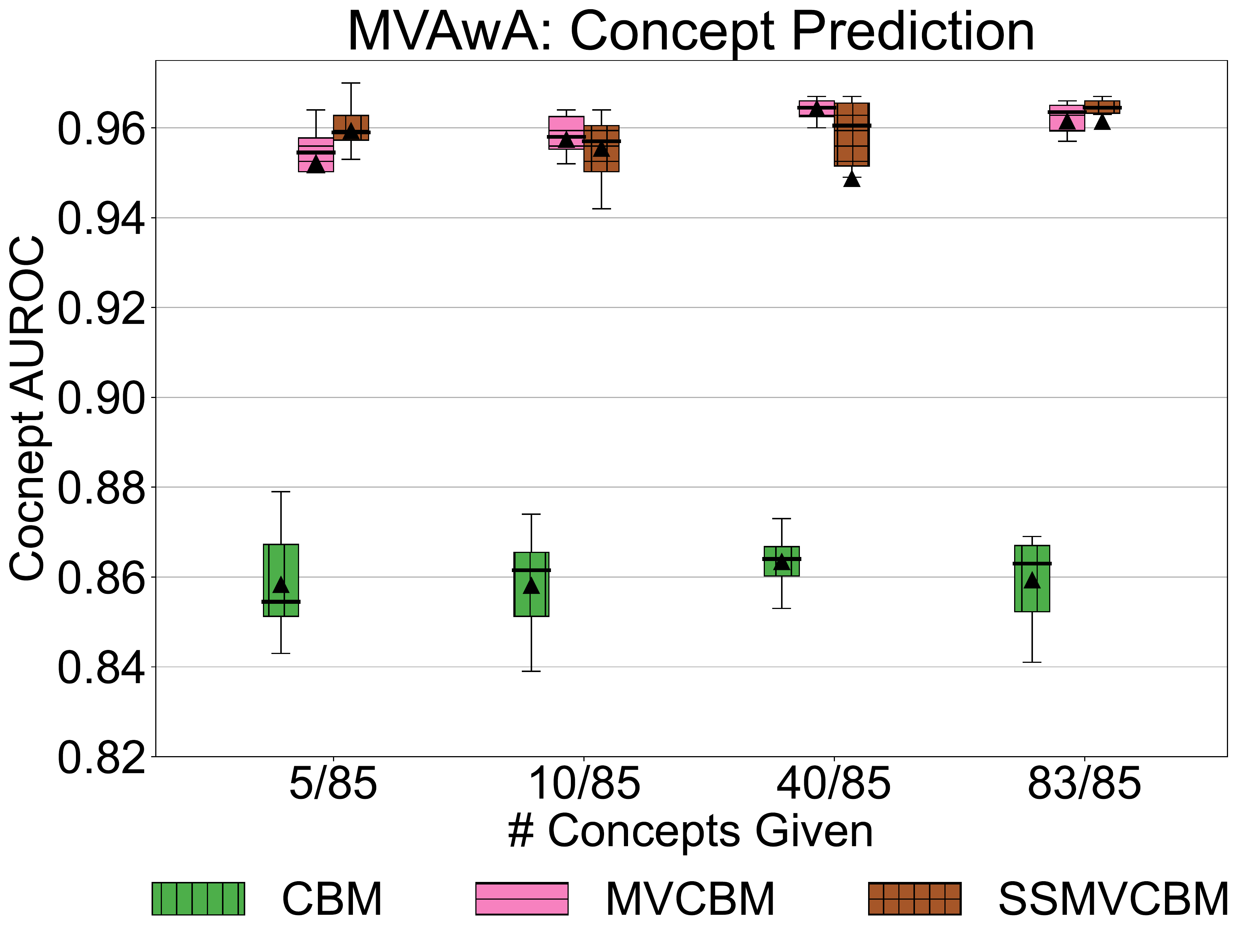}
}
\subfigure[]{
    \includegraphics[width=0.3\linewidth]{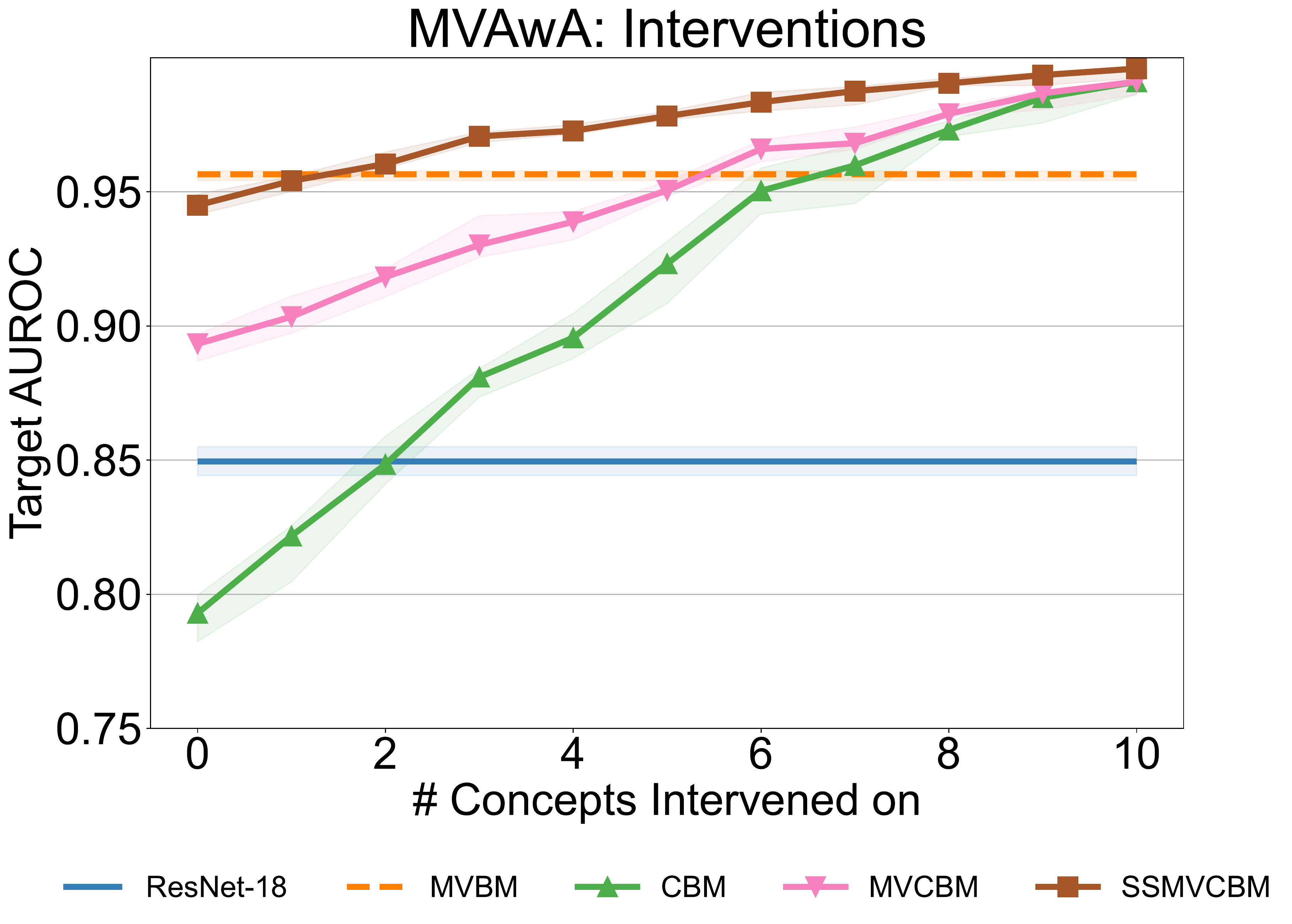}
}

\caption{Target and concept prediction results on the semi-synthetic multiview animals with attributes (MVAwA) for the proposed multiview concept bottleneck (MVCBM) and semi-supervised multiview concept bottleneck (SSMVCBM) models alongside several baselines. All plots were produced across ten independent simulations. \mbox{(a) One-}vs-all AUROCs for predicting the target on the test data under the varying number of observed concepts. ResNet-18 and MVBM do not rely on concepts; their AUROCs are shown as horizontal lines for reference. \mbox{(b) AUROCs} for predicting concepts on the test data under the varying number of observed concepts. AUROCs were averaged across the observed concepts. \mbox{(c) AUROCs} for predicting the target on the test data after intervening on the varying number of concepts. The intervention experiment was performed for 10/85 observed concepts, i.e. under an incomplete concept set. Confidence bands correspond to interquartile ranges across independent simulations and several randomly sampled concept subsets.}
\label{fig:mvawa}
\end{figure}

For this dataset,  the experiment results are very similar to the ones on the synthetic data: \mbox{(i) multiview} techniques perform superior to single-view techniques, as shown in Figures~\mbox{\ref{fig:mvawa}}(a)--(b); \mbox{(ii) when} given the complete concept set, MVCBM is comparable to an end-to-end black-box, as shown in Figure~\mbox{\ref{fig:mvawa}}(a); and \mbox{(iii) the} proposed multiview and semi-supervised extensions of the CBM are intervenable, as shown in Figure~\ref{fig:mvawa}(c). 
Herein, for MVCBMs, we focused on a simple approach to aggregating multiple views and a single optimization procedure; however, other design choices are plausible. 
Table~\ref{tab:mvawa_full_res} reports additional results with alternative fusion functions and optimization schemes for the MVAwA experiment under the complete concept set.

\begin{table}[H]
\centering
\caption{Target and concept prediction results for MVCBM and SSMVCBM models and several baselines under different optimization procedures and fusion functions on the MVAwA data with the full concept set. The performance is reported as averages and standard deviations of the AUROC across ten independent simulations. Herein, ``seq'' and ``joint'' denote sequential and joint optimization, respectively; whereas ``avg'' and ``LSTM'' stand for the averaging- and LSTM-based fusion. Bold indicates the best result; italics indicates the second best.}
\label{tab:mvawa_full_res}
\scriptsize
\begin{tabular}{lm{1.25cm}cm{1.25cm}}
    \toprule
    \multirow{2}{*}{\textbf{Model}} & \multicolumn{3}{c}{\textbf{AUROC}}\\
    \cmidrule{2-4} &
    \multicolumn{1}{c}{\textbf{Target}} & & \multicolumn{1}{c}{\textbf{Concepts}} \\
    \toprule
    Random & 0.50 &  & 0.50 \\
    \midrule
    ResNet-18 & \tentry{0.85}{0.01} & & --- \\
    \midrule
    CBM-seq & \tentry{0.81}{0.01} & & \tentry{0.86}{0.01} \\
    CBM-joint & \tentry{0.84}{0.01} & & \tentry{0.85}{0.01} \\
    \midrule
    MVBM-avg & \textbf{\tentry{0.96}{0.00}} & & --- \\
    MVBM-LSTM & \textit{\tentry{0.95}{0.00}} & & --- \\
    \midrule
    MVCBM-seq-avg & \textit{\tentry{0.95}{0.01}} & & \textbf{\tentry{0.97}{0.00}} \\
    MVCBM-seq-LSTM & \tentry{0.92}{0.01} & & \tentry{0.95}{0.00} \\
    MVCBM-joint-avg & \tentry{0.94}{0.01} & & \textit{\tentry{0.96}{0.00}} \\
    MVCBM-joint-LSTM & \textit{\tentry{0.95}{0.00}} & & \textit{\tentry{0.96}{0.01}} \\
    \midrule
    SSMVCBM-avg & \tentry{0.94}{0.00} & & \textbf{\tentry{0.97}{0.00}} \\
    SSMVCBM-LSTM & \tentry{0.92}{0.01} & & \tentry{0.95}{0.00} \\
    \bottomrule
\end{tabular}
\end{table}

\subsection{SSMVCBM Ablation\label{app:ssmvcbm_abl}}

As mentioned before, the semi-supervised variant of the proposed multiview concept bottleneck model includes an adversarial regularizer to de-correlate learned representations $\boldsymbol{\hat{z}}\in\mathbb{R}^J$ and concept predictions $\boldsymbol{\hat{c}}\in\mathbb{R}^K$ (Eq.~(\ref{eqn:opt_ssmvcbm_minimax})). 
To better understand the impact of this regularization on the model's predictive performance and intervenability, we performed an ablation study on the synthetic tabular nonlinear and MVAwA datasets by training semi-supervised concept bottlenecks under varying values of the regularization parameter $\lambda\in\{0,\, 0.01,\, 0.1\}$. 

\begin{figure}[H]
    \centering
    \subfigure[]{
        \begin{minipage}{\textwidth}
            \scriptsize
            \begin{center}
                \begin{tabular}{llm{1.25cm}cm{1.25cm}c}
                    \toprule
                    \multirow{2}{*}{\textbf{Dataset}} & \multirow{2}{*}{\textbf{Model}} & \multicolumn{3}{c}{\textbf{AUROC}} & \multirow{2}{*}{$\mathrm{median}\: \left\vert\widehat{\mathrm{corr}}\left(\hat{c}_i,\hat{z}_j\,\vert\,y=k\right)\right\vert$} \\
                    \cmidrule{3-5} &
                    & \multicolumn{1}{c}{\textbf{Target}} & & \multicolumn{1}{c}{\textbf{Concepts}} \\
                    \toprule
                    \multirow{4}{*}{Synthetic} & MVCBM & \tentry{0.605}{0.029} &  & \tentry{0.756}{0.020} & --- \\
                    & SSMVCBM, $\lambda=0$ & \tentry{0.640}{0.020} &  & \tentry{0.756}{0.020} &  0.059; [0.045, 0.075] \\
                    & SSMVCBM, $\lambda=0.01$ & \tentry{0.638}{0.020} &  & \tentry{0.756}{0.020} & 0.059; [0.043, 0.082] \\
                    & SSMVCBM, $\lambda=0.1$ & \tentry{0.629}{0.014} &  & \tentry{0.756}{0.020} & 0.052; [0.040, 0.063] \\
                    \midrule
                    \multirow{4}{*}{MVAwA} & MVCBM & \tentry{0.893}{0.009} &  &\tentry{0.957}{0.006} & --- \\
                    & SSMVCBM, $\lambda=0$ & \tentry{0.963}{0.011} &  & \tentry{0.957}{0.006} & 0.114; [0.109, 0.120]\\
                    & SSMVCBM, $\lambda=0.01$ & \tentry{0.946}{0.008} &  & \tentry{0.957}{0.006} & 0.105; [0.097, 0.116] \\
                    & SSMVCBM, $\lambda=0.1$ & \tentry{0.927}{0.011} &  & \tentry{0.957}{0.006} & 0.071; [0.056, 0.081] \\
                    \bottomrule
                \end{tabular}
            \end{center}
            \vspace{0.5cm}
        \end{minipage}
    }
    \subfigure[]{
        \includegraphics[width=0.65\textwidth]{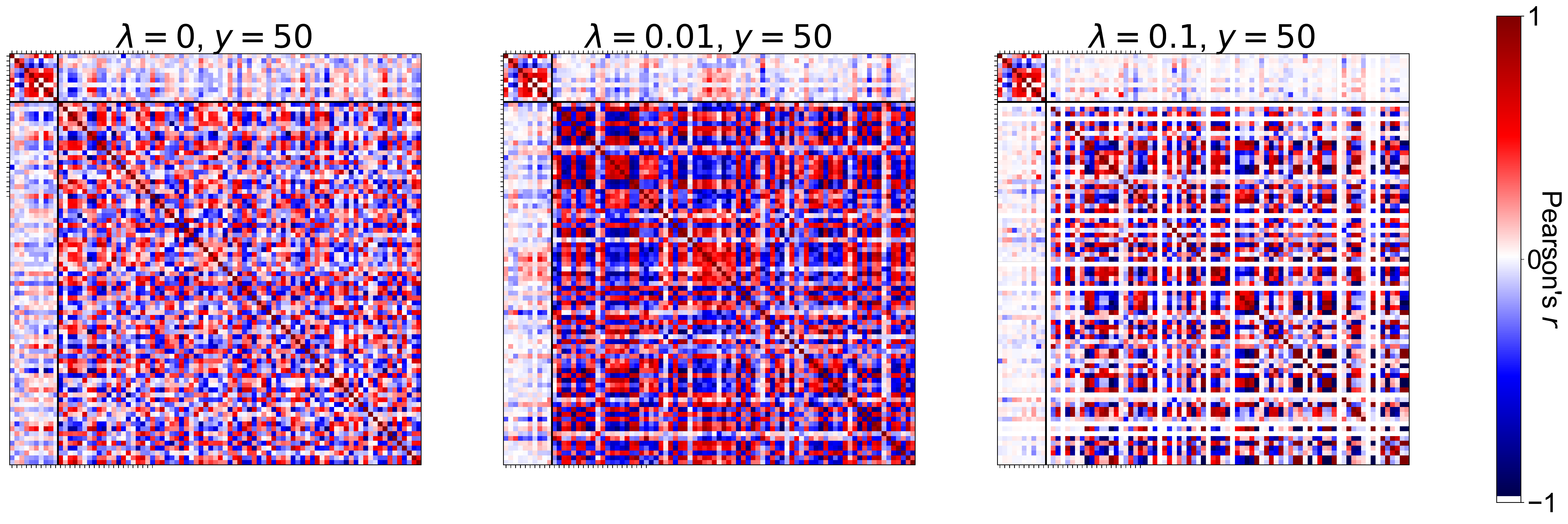}
    }
    \subfigure[]{
        \includegraphics[width=0.25\textwidth]{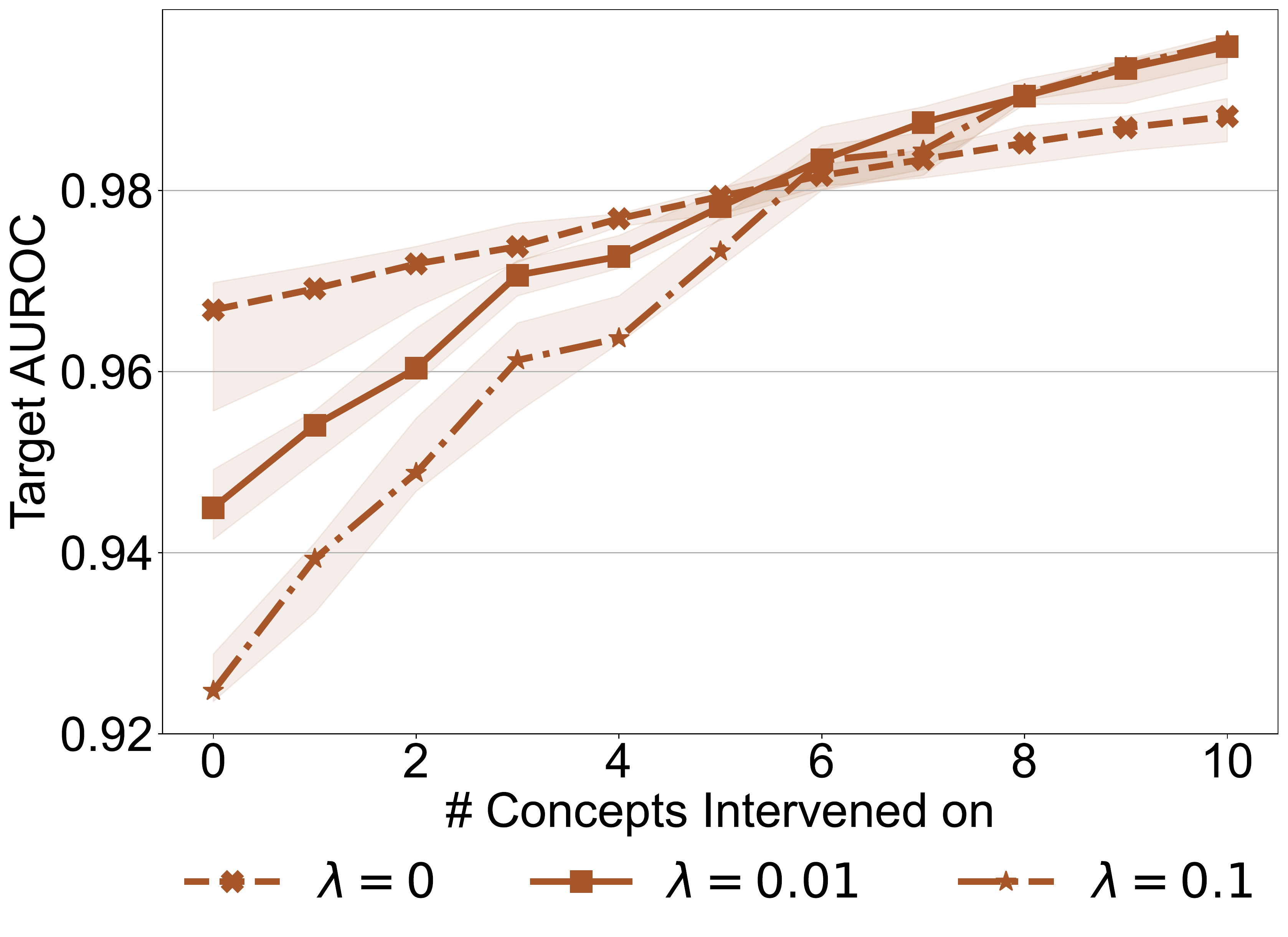}
    }
    \caption{Results of the ablation study on the effect of the adversarial regularization in the semi-supervised multiview concept bottleneck mo\mbox{dels (SSMVCBM).} \mbox{(a) Target} and concept prediction results alongside correlation between the predicted concepts and representations conditioned on the target variable. SSMVCBM models were trained on the synthetic and MVAwA datasets under varying  regularization parameter $\lambda=0,\,0.01,\,0.1$. AUROCs are reported as averages and standard deviations; for the correlation, we report median values and interquartile ranges. Statistics were computed across ten independent simulations. \mbox{(b) Conditional} correlation among the predicted concepts and representations for class $y=50$ in the MVAwA dataset. \mbox{(c) Intervention} experiment results on the MVAwA dataset for varying regularization strength. Predicted concepts were iteratively replaced with ground-truth values while measuring changes in AUROC for predicting the target variable. \label{fig:ssmvcbm_ablation}}
\end{figure}

In this experiment, we assessed the predictive performance and intervenability of the resulting models and the correlation among the individual dimensions of the concept predictions and representations. 
For the latter, we have utilized Pearson's correlation coefficient conditional on the target variable; in particular, we have looked at the median absolute value of the pairwise correlation coefficient given by $\mathrm{median}_{i,j,k}\: \left\vert\widehat{\mathrm{corr}}\left(\hat{c}_i,\hat{z}_j\,\vert\,y=k\right)\right\vert$, where $\hat{c}_i$ and $\hat{z}_j$ denote the $i$-th and $j$-th components of the concept and representation vectors, respectively. 
For both datasets, the experiment was run under the incomplete concept set: $K=5$ (out of 30) observed concepts and $J=25$ for synthetic data and $K=10$ (out of 85) and $J=75$ for MVAwA. All results reported below correspond to the multiview CBMs with the averaging-based fusion.
Figure~\ref{fig:ssmvcbm_ablation} summarizes the results of the ablation study. 

It appears that stronger regularization expectedly hurts the performance at predicting the target variable but allows learning representations de-correlated from the given concepts, as shown in Figures~\ref{fig:ssmvcbm_ablation}(a) and \ref{fig:ssmvcbm_ablation}(b).
However, even in the absence of the adversarial regularization ($\lambda=0$), $\boldsymbol{\hat{c}}$ and $\boldsymbol{\hat{z}}$ are already relatively weakly correlated. 
Importantly, regularized models demonstrate a steeper increase in predictive performance during interventions on predicted concepts (Figure~\ref{fig:ssmvcbm_ablation}(c)). 
Moreover, when most of the concepts have been intervened on, the unregularized model predicts the target variable more poorly than the regularized ones.

In summary, we observed that the adversarial regularizer in the SSMVCBM's loss function helps de-correlated representation learning and improves the model's intervenability, albeit it may reduce the non-intervened model's predictive performance. 
In future work, it would be interesting to seek alternative regularization techniques for disentangling concepts and representations, possibly focusing explicitly on intervenability.

\newpage

\subsection{LSTM-based Fusion and View Ordering\label{app:lstm}}

To explore the sensitivity of the LSTM-based multiview concept bottlenecks to the view order, we performed an additional experiment on the pediatric appendicitis dataset, where the views are ordered chronologically. After randomly shuffling the views within every subject, we applied all LSTM-based multiview CBMs to the test set. Thus, the models trained on ordered data were assessed on the set with the perturbed order.

Table~\ref{tab:appendicitis_lstm_ablation} reports the predictive performance obtained on the original test set (for reference) and after shuffling the views for the three target variables. Expectedly, all LSTM-based multiview CBMs are sensitive to the order of inputs, especially for predicting the management and severity. The relative performance decrease after shuffling is particularly remarkable for the SSMVCBM.
\begin{table}[H]
    \scriptsize
    \centering
    \caption{Test-set performance of the multiview concept bottleneck models using the LSTM-based fusion. Models were evaluated on the test set with the views ordered chronologically, as in the training set, and after shuffling the views. AUROCs and AUPRs are reported as averages and standard deviations across ten independent initialization. Bold indicates the best result; italics indicates the second best.}
    \label{tab:appendicitis_lstm_ablation} 
    \begin{tabular}{llm{1.2cm}m{1.2cm}cm{1.2cm}m{1.2cm}cm{1.2cm}m{1.2cm}}
        \toprule
        \multirow{2}{*}{\textbf{Model}} & \multirow{2}{*}{\textbf{Shuffled?}} & \multicolumn{2}{c}{\textbf{Diagnosis}} &  & \multicolumn{2}{c}{\textbf{Management}} &  & \multicolumn{2}{c}{\textbf{Severity}}\\
        \cmidrule{3-4}\cmidrule{6-7}\cmidrule{9-10} &  & \multicolumn{1}{c}{\textbf{AUROC}} & \multicolumn{1}{c}{\textbf{AUPR}} &  & \multicolumn{1}{c}{\textbf{AUROC}} & \multicolumn{1}{c}{\textbf{AUPR}} &  & \multicolumn{1}{c}{\textbf{AUROC}} & \multicolumn{1}{c}{\textbf{AUPR}} \\
        \toprule
        Random & --- & 0.50 & 0.75 &  & 0.50 & 0.47 &  & 0.50 & 0.23 \\
        \midrule
        MVCBM-seq-LSTM & no & \textit{\tentry{0.73}{0.03}} & \textit{\tentry{0.89}{0.01}} &  & \textit{\tentry{0.57}{0.03}} & \tentry{0.53}{0.04} &  & \textit{\tentry{0.70}{0.11}} & \textit{\tentry{0.48}{0.16}} \\
        MVCBM-seq-LSTM & yes & \tentry{0.69}{0.04} & \tentry{0.88}{0.02} &  & \tentry{0.56}{0.05} & \textit{\tentry{0.57}{0.08}} &  & \tentry{0.55}{0.14} & \tentry{0.29}{0.11} \\
         \cdashline{1-10}
        MVCBM-joint-LSTM & no & \tentry{0.72}{0.02} & \tentry{0.88}{0.02} &  & \textit{\tentry{0.57}{0.05}} & \tentry{0.50}{0.04} &  & \tentry{0.65}{0.07} & \tentry{0.37}{0.10} \\
        MVCBM-joint-LSTM & yes & \textit{\tentry{0.73}{0.01}} & \textit{\tentry{0.89}{0.01}} &  & \tentry{0.52}{0.04} & \tentry{0.48}{0.05} &  & \tentry{0.47}{0.07} & \tentry{0.22}{0.07} \\
        \midrule
        SSMVCBM-LSTM & no & \textbf{\tentry{0.80}{0.06}} & \textbf{\tentry{0.92}{0.04}} &  & \textbf{\tentry{0.70}{0.03}} & \textbf{\tentry{0.67}{0.06}} &  & \textbf{\tentry{0.78}{0.05}} & \textbf{\tentry{0.58}{0.10}} \\
        SSMVCBM-LSTM & yes & \tentry{0.72}{0.05} & \tentry{0.88}{0.04} &  & \tentry{0.55}{0.05} & \tentry{0.51}{0.07} &  & \tentry{0.62}{0.10} & \tentry{0.36}{0.11} \\
        \bottomrule
    \end{tabular}
\end{table}

Thus, the results suggest that the LSTM-based fusion, expectedly, is sensitive to the order of input views and, therefore, when deployed, LSTM-based (SS)MVCBMs should be applied with caution, preserving the ordering of images represented in the training data.

\newpage

\subsection{True and False Positive Rates for Predicting Appendicitis\label{app:fprs}}

In addition to the overall AUROCs and AUPRs for the target prediction reported in Table~\ref{tab:appendicitis_target_res}, for the diagnosis, we examined false positive rates attained by the predictive models at fixed percentages of the true positive rate. The results of this analysis are shown in Table~\ref{tab:fpr_at_k} for the TPRs of 75, 80, 90, 95, and 99\%.

All models have relatively high FPRs, $> 30\%$, at all considered TPRs. Thus, in general, the amount of false positive predictions necessary to achieve a satisfactory TPR is too high for the predictive models to be used without human expert attendance. For the TPRs of 95 and 99\%, most models have an FPR of at least 80\%. 

Generally, the ordering among the models w.r.t. the predictive performance and relative performance of the different model classes are similar to those based on \mbox{AUROCs} and \mbox{AUPRs} (Table~\ref{tab:appendicitis_target_res}). In a similar vein to the results reported in Section~\ref{sec:results_app}, this further exploration suggests that the models' performance has to be improved for them to be practical and autonomous.
\begin{table}[H]
    \scriptsize
    \centering
    \caption{Test-set false positive rates (FPR) at varying percentages of the true positive rate (TPR) on the pediatric appendicitis dataset with the diagnosis as the target variable. Results are reported as averages and standard deviations across ten independent initializations. Bold indicates the best result; italics indicates the second best. Lower FPRs are better.}
    \label{tab:fpr_at_k} 
    \begin{tabular}{llm{1.2cm}m{1.2cm}m{1.2cm}m{1.2cm}m{1.2cm}}
        \toprule
        \multirow{2}{*}{\textbf{Model}} & \multicolumn{6}{c}{\textbf{FPR at \% TPR}} \\
        & \textbf{\%} & \textbf{75} & \textbf{80} & \textbf{90} & \textbf{95} & \textbf{99} \\
        \toprule
        Random & & 0.75 & 0.80 & 0.90 & 0.95 & 0.99 \\
        \midrule
        Radiomics + RF & & \tentry{0.55}{0.05} & \tentry{0.63}{0.07} & \tentry{0.85}{0.05} & \tentry{0.94}{0.02} & \tentry{0.97}{0.03} \\
        \midrule
        ResNet-18 & & \tentry{0.55}{0.18} & \tentry{0.70}{0.11} & \tentry{0.85}{0.13} & \tentry{0.91}{0.11} & \tentry{0.95}{0.08} \\
        \midrule
        CBM-seq & & \tentry{0.61}{0.08} & \tentry{0.69}{0.11} & \tentry{0.80}{0.08} & \tentry{0.87}{0.10} & \tentry{0.97}{0.05} \\
        CBM-joint & & \tentry{0.67}{0.06} & \tentry{0.73}{0.08} & \tentry{0.84}{0.09} & \tentry{0.94}{0.06} & \tentry{0.95}{0.06} \\
        \midrule
        MVBM-avg & & \textit{\tentry{0.40}{0.14}} & \tentry{0.51}{0.16} & \tentry{0.74}{0.21} & \tentry{0.86}{0.12} & \tentry{0.95}{0.07} \\
        MVBM-LSTM & & \tentry{0.45}{0.09} & \tentry{0.53}{0.12} & \tentry{0.74}{0.14} & \tentry{0.85}{0.09} & \tentry{0.94}{0.09} \\
        \midrule
        MVCBM-seq-avg & & \tentry{0.58}{0.14} & \tentry{0.61}{0.10} & \tentry{0.77}{0.09} & \tentry{0.89}{0.09} & \textit{\tentry{0.93}{0.07}} \\
        MVCBM-seq-LSTM & & \tentry{0.52}{0.04} & \tentry{0.55}{0.06} & \textit{\tentry{0.67}{0.11}} & \textbf{\tentry{0.75}{0.12}} & \textit{\tentry{0.93}{0.09}} \\
        MVCBM-joint-avg & & \tentry{0.57}{0.15} & \tentry{0.64}{0.10} & \tentry{0.77}{0.11} & \tentry{0.93}{0.09} & \tentry{0.95}{0.08} \\
        MVCBM-joint-LSTM & & \tentry{0.59}{0.10} & \tentry{0.65}{0.09} & \tentry{0.75}{0.11} & \tentry{0.84}{0.08} & \textit{\tentry{0.93}{0.08}} \\
        \midrule
        SSMVCBM-avg & & \textbf{\tentry{0.31}{0.11}} & \textit{\tentry{0.40}{0.11}} & \textbf{\tentry{0.65}{0.14}} & \textit{\tentry{0.78}{0.15}} & \textbf{\tentry{0.88}{0.14}} \\
        SSMVCBM-LSTM & & \textbf{\tentry{0.31}{0.16}} & \textbf{\tentry{0.38}{0.15}} & \tentry{0.68}{0.11} & \tentry{0.84}{0.10} & \textit{\tentry{0.93}{0.07}} \\
        \bottomrule
    \end{tabular}
\end{table}

\newpage

\subsection{Brier Scores for Concept Prediction\label{app:brier_scores}}

To supplement AUROCs and AUPRs reported in Tables~\ref{tab:appendicitis_concept_res}-\ref{tab:appendicitis_concept_res_severity}, we evaluated concept predictions in terms of the Brier score, as shown in Table~\ref{tab:appendicitis_concept_res_brier_scores}. Note that the scores were not adjusted for class imbalance, and most concept variables had few positive observations (Table~\ref{tab:concept_distribution}). The findings from this analysis are described and discussed in Sections~\ref{sec:results} and \ref{sec:discussion} of the main text, respectively.

\begin{table}[H]
    \scriptsize
    \begin{center}
    \caption{Models' Brier scores for concept prediction on the appendicitis dataset across the three target variables. Test-set results are reported as averages and standard deviations across ten independent initializations. Herein, ``seq'' and ``joint'' denote sequential and joint optimization, respectively, whereas ``avg'' and ``LSTM'' stand for the averaging- and LSTM-based fusion. Results that are significantly lower than the score of the constant prediction of 0.5 (random) are marked by ``\textsuperscript{\textbf{*}}''. Bold indicates the best result; italics indicates the second best. The meaning of the concept variables: $c_1$, visibility of the appendix; $c_2$, free intraperitoneal fluid; $c_3$, appendix layer structure; $c_4$, target sign; $c_5$, surrounding tissue reaction; $c_6$, pathological lymph nodes; $c_7$, thickening of the bowel wall; $c_8$, coprostasis; $c_9$, meteorism.}
    \label{tab:appendicitis_concept_res_brier_scores} 
    \begin{tabular}{p{0.8cm}lp{0.8cm}p{0.8cm}p{0.8cm}p{0.8cm}p{0.8cm}p{0.8cm}p{0.8cm}p{0.8cm}p{0.8cm}}
        \toprule
        \multirow{2}{*}{\raisebox{-\heavyrulewidth}{\textbf{Target}}} & \multirow{2}{*}{\raisebox{-\heavyrulewidth}{\textbf{Model}}} & \multicolumn{9}{c}{\textbf{Concept}}\\
        \cmidrule{3-11} & & $c_1$ & $c_2$ & $c_3$ & $c_4$ & $c_5$ & $c_6$ & $c_7$ & $c_8$ & $c_9$ \\
        
        \toprule
        \multirow{11}{20mm}{\begin{sideways}\textbf{Diagnosis}\end{sideways}}
        & \multicolumn{1}{l}{Random} & 0.25 & \textit{0.25} & 0.25 & 0.25 & \textit{0.25} & 0.25 & 0.25 & 0.25 & 0.25 \\
        \cmidrule{2-11} & \multicolumn{1}{l}{CBM-seq} & \tentry{0.26}{0.02} & \tentry{0.31}{0.02} & \tentry{0.20}{0.02}\textsuperscript{*} & \tentry{0.21}{0.01}\textsuperscript{*} & \tentry{0.26}{0.02} & \textit{\tentry{0.23}{0.02}\textsuperscript{*}} & \textit{\tentry{0.17}{0.02}\textsuperscript{*}} & \textit{\tentry{0.14}{0.01}\textsuperscript{*}} & \textit{\tentry{0.18}{0.02}\textsuperscript{*}} \\
        & \multicolumn{1}{l}{CBM-joint} & \tentry{0.30}{0.03} & \tentry{0.42}{0.02} & \textbf{\tentry{0.18}{0.01}\textsuperscript{*}} & \tentry{0.21}{0.02}\textsuperscript{*} & \tentry{0.32}{0.04} & \tentry{0.24}{0.03} & \textbf{\tentry{0.16}{0.01}\textsuperscript{*}} & \textbf{\tentry{0.12}{0.01}\textsuperscript{*}} & \textbf{\tentry{0.15}{0.02}\textsuperscript{*}} \\
        \cmidrule{2-11} & \multicolumn{1}{l}{MVCBM-seq-avg} & \tentry{0.26}{0.04} & \tentry{0.26}{0.01} & \tentry{0.21}{0.03}\textsuperscript{*} & \tentry{0.22}{0.02}\textsuperscript{*} & \tentry{0.28}{0.02} & \tentry{0.24}{0.03} & \tentry{0.23}{0.03} & \tentry{0.28}{0.03} & \tentry{0.26}{0.03} \\
        & \multicolumn{1}{l}{MVCBM-seq-LSTM} & \tentry{0.17}{0.02}\textsuperscript{*} & \textit{\tentry{0.25}{0.01}} & \textit{\tentry{0.19}{0.02}\textsuperscript{*}} & \textbf{\tentry{0.19}{0.01}\textsuperscript{*}} & \textit{\tentry{0.25}{0.01}} & \tentry{0.26}{0.01} & \tentry{0.21}{0.02}\textsuperscript{*} & \tentry{0.27}{0.02} & \tentry{0.25}{0.02} \\
        & \multicolumn{1}{l}{MVCBM-joint-avg} & \tentry{0.29}{0.07} & \tentry{0.32}{0.04} & \tentry{0.22}{0.07} & \tentry{0.25}{0.06} & \tentry{0.32}{0.07} & \textbf{\tentry{0.22}{0.03}} & \tentry{0.18}{0.04}\textsuperscript{*} & \tentry{0.22}{0.16} & \tentry{0.23}{0.10} \\
        & \multicolumn{1}{l}{MVCBM-joint-LSTM} & \textbf{\tentry{0.15}{0.01}\textsuperscript{*}} & \textbf{\tentry{0.24}{0.00}\textsuperscript{*}} & \tentry{0.22}{0.01}\textsuperscript{*} & \tentry{0.21}{0.01}\textsuperscript{*} & \textbf{\tentry{0.22}{0.01}\textsuperscript{*}} & \tentry{0.26}{0.00} & \tentry{0.22}{0.02}\textsuperscript{*} & \tentry{0.26}{0.01} & \tentry{0.24}{0.02} \\
        \cmidrule{2-11} & \multicolumn{1}{l}{SSMVCBM-avg} & \tentry{0.24}{0.04} & \tentry{0.26}{0.01} & \tentry{0.24}{0.09} & \tentry{0.22}{0.05} & \tentry{0.29}{0.02} & \textit{\tentry{0.23}{0.05}} & \tentry{0.22}{0.06} & \tentry{0.34}{0.13} & \tentry{0.23}{0.04} \\
        & \multicolumn{1}{l}{SSMVCBM-LSTM} & \textit{\tentry{0.16}{0.03}\textsuperscript{*}} & \textit{\tentry{0.25}{0.01}} & \tentry{0.22}{0.07} & \textit{\tentry{0.20}{0.03}\textsuperscript{*}} & \textit{\tentry{0.25}{0.03}} & \textit{\tentry{0.23}{0.04}} & \tentry{0.22}{0.06} & \tentry{0.28}{0.06} & \tentry{0.22}{0.04} \\
        
        \midrule
        \multirow{11}{20mm}{\begin{sideways}\textbf{Management}\end{sideways}}
        & \multicolumn{1}{l}{Random} & 0.25 & \textbf{0.25} & 0.25 & 0.25 & \textit{0.25} & 0.25 & 0.25 & 0.25 & 0.25 \\
        \cmidrule{2-11} & \multicolumn{1}{l}{CBM-seq} & \tentry{0.28}{0.03} & \tentry{0.27}{0.02} & \textit{\tentry{0.20}{0.03}\textsuperscript{*}} & \tentry{0.26}{0.05} & \textit{\tentry{0.25}{0.03}} & \tentry{0.25}{0.03} & \textit{\tentry{0.21}{0.04}} & \tentry{0.18}{0.05}\textsuperscript{*} & \tentry{0.22}{0.02}\textsuperscript{*} \\
        & \multicolumn{1}{l}{CBM-joint} & \tentry{0.33}{0.05} & \tentry{0.39}{0.05} & \textbf{\tentry{0.17}{0.03}\textsuperscript{*}} & \tentry{0.26}{0.06} & \tentry{0.34}{0.04} & \tentry{0.28}{0.05} & \textbf{\tentry{0.18}{0.03}\textsuperscript{*}} & \textbf{\tentry{0.14}{0.03}\textsuperscript{*
}} & \textit{\tentry{0.20}{0.05}} \\
        \cmidrule{2-11} & \multicolumn{1}{l}{MVCBM-seq-avg} & \tentry{0.26}{0.04} & \textit{\tentry{0.26}{0.01}} & \tentry{0.29}{0.06} & \tentry{0.26}{0.03} & \tentry{0.27}{0.01} & \textit{\tentry{0.22}{0.03}} & \tentry{0.28}{0.04} & \tentry{0.27}{0.05} & \tentry{0.25}{0.03} \\
        & \multicolumn{1}{l}{MVCBM-seq-LSTM} & \textbf{\tentry{0.16}{0.02}\textsuperscript{*}} & \textbf{\tentry{0.25}{0.01}} & \tentry{0.26}{0.03} & \textit{\tentry{0.24}{0.01}} & \textbf{\tentry{0.23}{0.01}\textsuperscript{*}} & \tentry{0.23}{0.01}\textsuperscript{*} & \tentry{0.28}{0.02} & \tentry{0.25}{0.01} & \tentry{0.23}{0.02} \\
        & \multicolumn{1}{l}{MVCBM-joint-avg} & \tentry{0.33}{0.13} & \tentry{0.28}{0.03} & \tentry{0.21}{0.05} & \textit{\tentry{0.24}{0.04}} & \tentry{0.31}{0.05} & \tentry{0.24}{0.06} & \tentry{0.22}{0.06} & \tentry{0.19}{0.09} & \tentry{0.26}{0.14} \\
        & \multicolumn{1}{l}{MVCBM-joint-LSTM} & \tentry{0.19}{0.03}\textsuperscript{*} & \tentry{0.32}{0.04} & \tentry{0.22}{0.07} & \textbf{\tentry{0.22}{0.03}\textsuperscript{*}} & \tentry{0.28}{0.03} & \tentry{0.23}{0.03} & \textit{\tentry{0.21}{0.04}} & \textit{\tentry{0.16}{0.04}\textsuperscript{*}} & \tentry{0.23}{0.06} \\
        \cmidrule{2-11} & \multicolumn{1}{l}{SSMVCBM-avg} & \tentry{0.21}{0.01}\textsuperscript{*} & \tentry{0.27}{0.02} & \tentry{0.34}{0.10} & \tentry{0.31}{0.05} & \tentry{0.28}{0.02} & \textbf{\tentry{0.21}{0.02}\textsuperscript{*}} & \tentry{0.26}{0.05} & \tentry{0.29}{0.07} & \textbf{\tentry{0.19}{0.03}\textsuperscript{*}} \\
        & \multicolumn{1}{l}{SSMVCBM-LSTM} & \textit{\tentry{0.17}{0.02}\textsuperscript{*}} & \textit{\tentry{0.26}{0.01}} & \tentry{0.26}{0.05} & \textit{\tentry{0.24}{0.03}} & \textbf{\tentry{0.23}{0.02}} & \textit{\tentry{0.22}{0.01}\textsuperscript{*}} & \tentry{0.29}{0.04} & \tentry{0.24}{0.02} & \tentry{0.22}{0.03} \\
        
        \midrule
        \multirow{11}{20mm}{\begin{sideways}\textbf{Severity}\end{sideways}}
        & \multicolumn{1}{l}{Random} & 0.25 & \textbf{0.25} & 0.25 & 0.25 & \textit{0.25} & 0.25 & 0.25 & 0.25 & 0.25 \\
        \cmidrule{2-11} & \multicolumn{1}{l}{CBM-seq} & \tentry{0.30}{0.04} & \textit{\tentry{0.26}{0.02}} & \tentry{0.23}{0.02} & \tentry{0.26}{0.05} & \tentry{0.27}{0.03} & \textit{\tentry{0.23}{0.03}} & \tentry{0.23}{0.03} & \tentry{0.21}{0.02}\textsuperscript{*} & \tentry{0.23}{0.04} \\
        & \multicolumn{1}{l}{CBM-joint} & \tentry{0.29}{0.05} & \tentry{0.40}{0.03} & \textbf{\tentry{0.18}{0.04}\textsuperscript{*}} & \tentry{0.27}{0.08} & \tentry{0.33}{0.05} & \tentry{0.30}{0.05} & \textit{\tentry{0.20}{0.04}\textsuperscript{*}} & \textbf{\tentry{0.15}{0.04}\textsuperscript{*}} & \textbf{\tentry{0.18}{0.04}\textsuperscript{*}} \\
        \cmidrule{2-11} & \multicolumn{1}{l}{MVCBM-seq-avg} & \tentry{0.29}{0.04} & \textit{\tentry{0.26}{0.01}} & \tentry{0.27}{0.07} & \tentry{0.27}{0.04} & \tentry{0.28}{0.02} & \textbf{\tentry{0.22}{0.03}} & \tentry{0.30}{0.08} & \tentry{0.28}{0.03} & \tentry{0.25}{0.03} \\
        & \multicolumn{1}{l}{MVCBM-seq-LSTM} & \textit{\tentry{0.19}{0.02}\textsuperscript{*}} & \textit{\tentry{0.26}{0.01}} & \tentry{0.24}{0.04} & \tentry{0.24}{0.02} & \textit{\tentry{0.25}{0.01}} & \textit{\tentry{0.23}{0.02}} & \tentry{0.27}{0.03} & \tentry{0.26}{0.02} & \tentry{0.25}{0.02} \\
        & \multicolumn{1}{l}{MVCBM-joint-avg} & \tentry{0.29}{0.03} & \tentry{0.28}{0.02} & \textit{\tentry{0.19}{0.05}\textsuperscript{*}} & \textit{\tentry{0.23}{0.04}} & \tentry{0.29}{0.03} & \tentry{0.24}{0.03} & \textbf{\tentry{0.19}{0.05}\textsuperscript{*}} & \textit{\tentry{0.17}{0.04}\textsuperscript{*}} & \tentry{0.26}{0.05} \\
        & \multicolumn{1}{l}{MVCBM-joint-LSTM} & \textbf{\tentry{0.18}{0.02}\textsuperscript{*}} & \tentry{0.28}{0.02} & \textbf{\tentry{0.18}{0.03}\textsuperscript{*}} & \textbf{\tentry{0.21}{0.03}} & \tentry{0.26}{0.03} & \tentry{0.25}{0.04} & \tentry{0.22}{0.04} & \tentry{0.20}{0.04} & \textit{\tentry{0.22}{0.03}} \\
        \cmidrule{2-11} & \multicolumn{1}{l}{SSMVCBM-avg} & \tentry{0.29}{0.06} & \tentry{0.28}{0.03} & \tentry{0.34}{0.15} & \tentry{0.32}{0.10} & \tentry{0.31}{0.06} & \textbf{\tentry{0.22}{0.02}\textsuperscript{*}} & \tentry{0.33}{0.16} & \tentry{0.30}{0.12} & \textit{\tentry{0.22}{0.05}} \\
        & \multicolumn{1}{l}{SSMVCBM-LSTM} & \tentry{0.23}{0.09} & \textit{\tentry{0.26}{0.01}} & \tentry{0.25}{0.06} & \tentry{0.26}{0.04} & \textbf{\tentry{0.24}{0.02}} & \textbf{\tentry{0.22}{0.03}} & \tentry{0.33}{0.12} & \tentry{0.26}{0.03} & \tentry{0.26}{0.05} \\
        \bottomrule
    \end{tabular}
    \vspace{1cm}
    \end{center}
\end{table}

\section{Online Prediction Tool\label{app:tool}}

Below, we provide details on the implementation of our online pediatric appendicitis prediction tool. 
We must emphasize that the current version is a research prototype and should be utilized solely for non-commercial, educational purposes, and not for clinical decision-making.
The web tool deploys a multiview concept bottleneck model trained to predict the diagnosis using the sequential optimization procedure and LSTM to fuse the views (MVCBM-seq-LSTM in Table~\ref{tab:appendicitis_target_res}). 
We use a single set of parameters obtained after training from one of the initializations included in the experiments. 
Note that the model \emph{was not} re-trained on the complete dataset.

\paragraph{Workflow}

\begin{figure}[H]%
\centering
\includegraphics[width=1.0\textwidth]{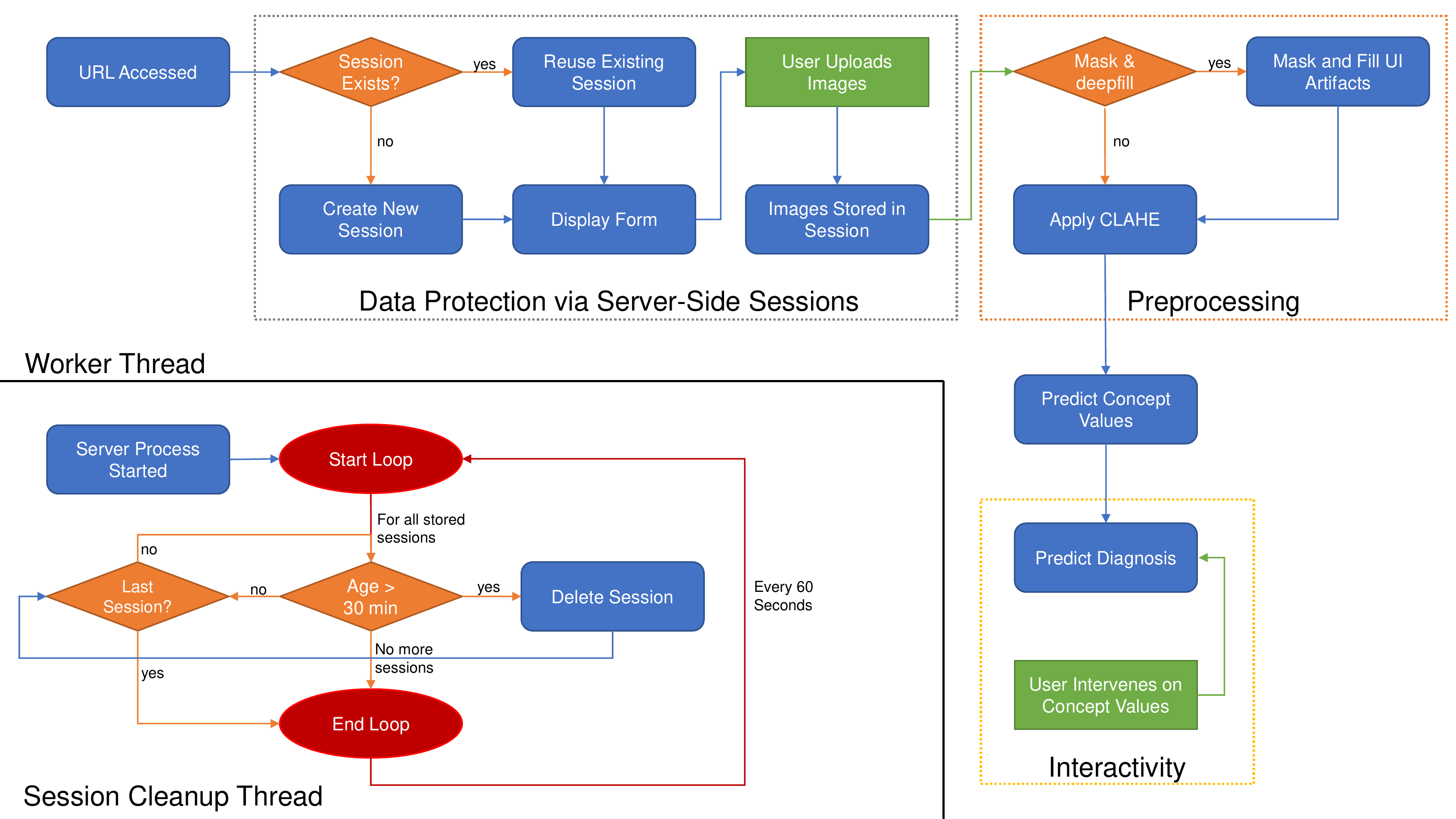}
\caption{Workflow diagram of the online appendicitis prediction tool.}\label{fig:webtool}
\end{figure}

Figure~\ref{fig:webtool} contains a workflow diagram for the website. 
Specifically, the \emph{worker thread} handles incoming requests and creates a new server-side session if it does not exist for the current user. 
The images uploaded by the user are saved in the session. 
If requested, UI element regions are masked and filled, and CLAHE is applied to the input images. 
Note that due to the characteristics of the images, the effectiveness of preprocessing may be limited. 
In particular, UI artifacts, such as text, logos and diagrams, which differ considerably from those in our collected dataset, may not be completely masked and filled. 
The processed images are then forwarded to the trained MVCBM network, which predicts the concept values and the diagnosis label and displays them. 
The user may intervene if they choose and re-calculate the final prediction using adjusted concept values. 
In the background, the \emph{session cleanup thread} is started along with the web application. 
It iterates every 60 seconds over all stored session objects. 
Sessions that have been inactive for over 30 minutes are eliminated, along with all related data.
After this, no data provided by the user or data resulting from processing the user's uploaded data are retained.

\end{appendix}

\end{appendices}


\end{document}